\documentclass[11pt]{article}
\pdfoutput=1 

\usepackage[a4paper,margin=1in]{geometry}
\usepackage{times}
\usepackage{latexsym}
\usepackage[T1]{fontenc}
\usepackage[utf8]{inputenc}
\usepackage{microtype}
\usepackage{inconsolata}

\usepackage{amsmath}
\usepackage{amssymb}
\usepackage{mathtools}

\usepackage{graphicx}
\usepackage{float}        
\usepackage{subcaption}
\usepackage{booktabs}
\usepackage{multirow}
\usepackage{array}
\usepackage{tabularx}
\usepackage{threeparttable}
\usepackage{wrapfig}
\usepackage{placeins}

\usepackage[table]{xcolor}
\definecolor{ThemeBlue}{RGB}{30,90,170}

\usepackage{url}
\usepackage[numbers,sort&compress]{natbib}
\usepackage[hidelinks]{hyperref}
\hypersetup{colorlinks=true,
  linkcolor=ThemeBlue!85!black,
  citecolor=ThemeBlue!85!black,
  urlcolor=ThemeBlue!85!black}

\usepackage[most]{tcolorbox}
\usepackage{titlesec}
\usepackage{enumitem}

\usepackage[capitalize,noabbrev]{cleveref}

\titleformat{\section}{\normalfont\Large\bfseries}{\thesection.}{0.5em}{}
\titleformat{\subsection}{\normalfont\large\bfseries}{\thesubsection}{0.5em}{}

\makeatletter
\renewcommand{\maketitle}{%
  \begingroup
    \begin{flushleft}
      {\huge\bfseries\@title\par}
      \vspace{1.0em}
      {\normalsize\@author\par}
    \end{flushleft}
  \endgroup
  \vspace{1.0em}
}
\makeatother

\renewenvironment{abstract}{%
  \begin{tcolorbox}[
    enhanced,
    colback=ThemeBlue!7,
    colframe=ThemeBlue!7,
    arc=2pt,
    boxrule=0pt,
    left=14pt, right=14pt, top=10pt, bottom=10pt,
  ]
  \small\noindent\ignorespaces
}{%
  \end{tcolorbox}
  \vspace{0.6em}
}

\newcommand{\model}{LVLMs }

\title{\textcolor{ThemeBlue}{Self-Improving Small Object Grounding in LVLMs}}

\author{%
  Tianze Yang \quad Yucheng Shi \quad Ruitong Sun \quad Ninghao Liu \quad Jin Sun\\[2pt]
  University of Georgia\\[3pt]
  {\normalsize\textbf{Project page:}~\url{https://groundvlm.github.io/}}
}

\begin{document}
\flushbottom 
\maketitle

\begin{abstract}
Can internal attention patterns in Large Vision Language Models (LVLMs) identify reliable small-object boxes without fine-tuning? In this work, we provide an affirmative answer. Attention structure in LVLMs encodes grounding quality---a lightweight IoU regressor trained solely on attention maps achieves strong IoU prediction (Pearson $r$> 0.67). This regressor powers the regressor-based variant of our \textbf{Attention-based Candidate Selection (ACS)} framework, called \textbf{ACS-Learned}, which selects the best box from multiple sampled candidates to improve object grounding. By analyzing what the regressor learns, we reveal which transformer layers and heads are most critical and derive \textbf{ACS-Free}: a training-free selector that ranks candidates by attention entropy on these discriminative heads, with no learned component at inference. Experiments on COCO and Objects365 demonstrate up to 19\% self-improvement on small object localization, with ACS-Free ranking best among all training-free methods, demonstrating that useful attention structure improves both localization reliability and interpretability in LVLMs.
\end{abstract}
\section{Introduction}
\label{sec:intro}

Large Vision Language Models (LVLMs)~\cite{bai2025qwen2,deitke2025molmo,wang2025internvl3,liu2023visual,chen2023shikra}
have shown remarkable capabilities in vision-language joint reasoning.
Recently, researchers explored using \model for direct object localization of bounding box coordinates~\cite{kang2025your,Yin_2025_CVPR}
without specialized detection heads (e.g., Qwen2.5-VL \cite{bai2025qwen2} and InternVL-3.5 \cite{wang2025internvl3}). This capability is crucial for autonomous navigation, robotics, AR, and visual assistance.

\begin{figure}[t]
    \centering
    \includegraphics[width=1.0\linewidth]{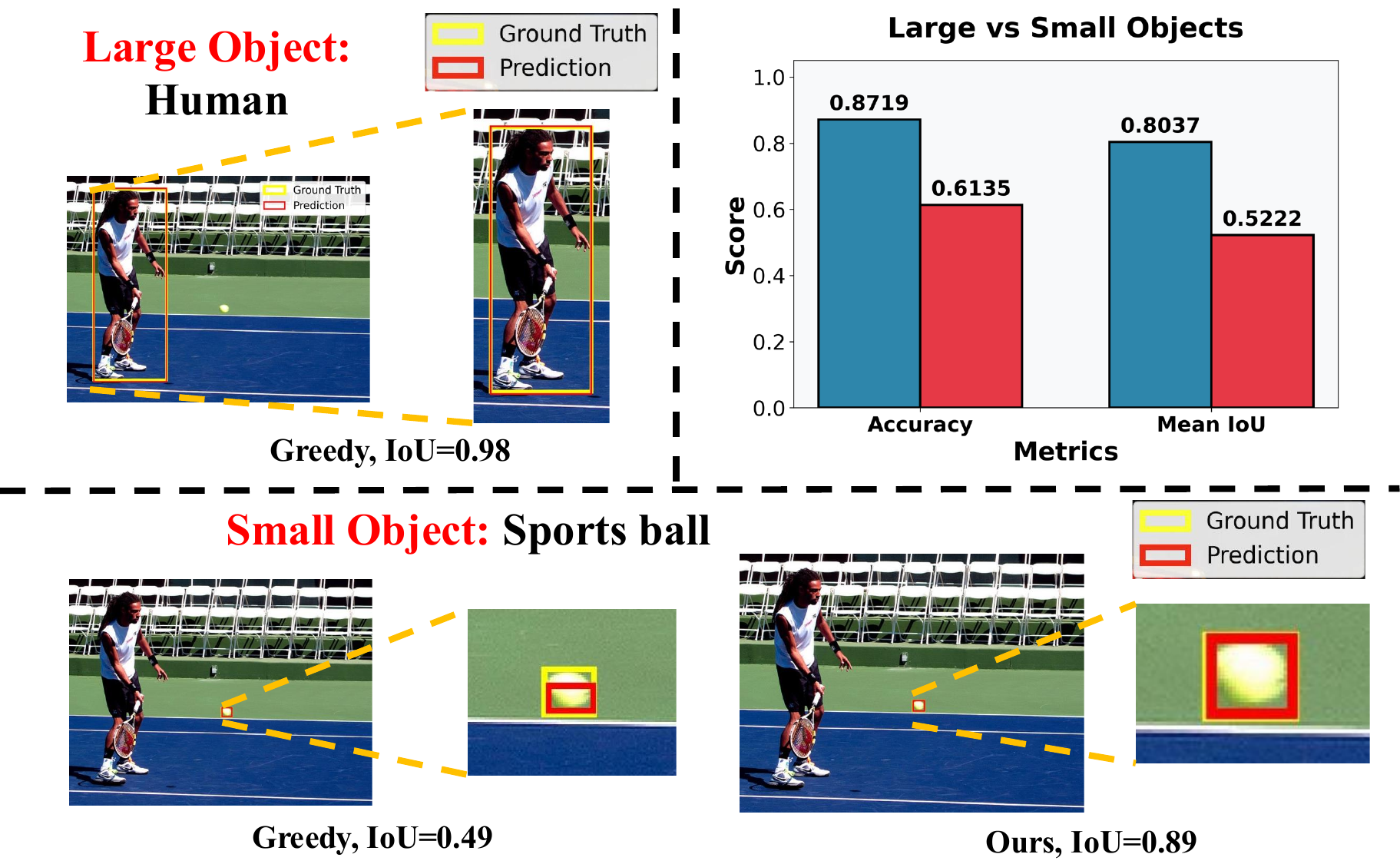}
    \caption{Performance comparison between large and small object grounding in Qwen2.5-VL. The significant performance gap highlights the challenge of small object detection.
    Our self-improvement framework helps LVLMs to find small targets.
    }
    \label{fig:large_small_comparison}
\end{figure}

Despite the notable progress,
we found that SOTA \model struggle with grounding \textit{small} objects:
as illustrated in Figure~\ref{fig:large_small_comparison},
they achieve \textbf{strong performance on large objects but not on small objects}.
This
is particularly concerning for safety-critical deployment, where small objects such as distant pedestrians are common.

To improve general localization performance, prior works~\cite{shen2025vlm, lai2024lisa, Yin_2025_CVPR, zhan2025griffon}
use
additional fine-tuning, external localization modules, or heuristic decoding strategies, yet they require substantial computational resources or architectural modifications.
Instead, we ask the following fundamental and intriguing question:

\vspace{-0.1in}
\begin{quote}
\emph{Can internal representations identify reliable small-object boxes without fine-tuning LVLMs?}
\end{quote}
\vspace{-0.1in}
In this work, we provide an affirmative answer. We propose a novel framework that leverages the intrinsic attention patterns of \model to estimate bounding box quality and select the best localization output from multiple sampled responses, with a focus on small objects. Our key insight is that the spatial attention structures across layers and heads correlate strongly with grounding quality.
To validate this idea, we introduce our \textbf{Attention-based Candidate Selection (ACS)} framework. First, we perform multi-response sampling to collect a set of candidate bounding boxes and their associated attention maps. Then, we train a lightweight IoU regressor to predict bounding box quality directly from attention patterns, demonstrating that attention alone carries sufficient signal for quality estimation. The regressor serves dual purposes: (1) as the regressor-based selector \textbf{ACS-Learned} that directly scores each candidate; and (2) as an analytical tool---through gradient attribution and entropy analysis of the trained regressor, we identify which layers and heads are most critical for localization and discover that specific attention heads exhibit strong entropy-IoU correlations. This finding enables \textbf{ACS-Free}, a training-free variant that selects boxes purely from entropy patterns in those discriminative heads, eliminating the need for any learned component at inference.

ACS
requires no modification, fine-tuning, or auxiliary supervision to LVLMs, apart from a small regression module used for analyzing attention patterns and selecting high-quality candidates.
This design achieves \textit{self-improvement} and allows our method to be applied to different LVLMs,
serving as a plug-and-play enhancement to those models.

Our main contributions are:



    \noindent \textbf{1) Discovery.} We identify a strong connection between internal attention patterns of \model and grounding quality: attention structure encodes whether a predicted small-object box is reliable. 

    \noindent \textbf{2) Validation and exploitation.} A lightweight IoU regressor trained on attention maps confirms this with high correlation, and ACS-Learned uses it directly for candidate selection, achieving consistent gains across different LVLMs and datasets.

    \noindent \textbf{3) Interpretability and distillation.} Gradient and entropy analysis of the regressor reveals which transformer heads matter for localization. ACS-Free operationalizes this as a parameter-free entropy rule, achieving the best performance among all training-free baselines.

    \noindent \textbf{4) Results and insights.} Our framework achieves up to 19\% self-improvement on small object localization without LVLM fine-tuning or external detectors. The localization-critical layers we identify provide interpretable insight into how LVLMs process spatial information in grounding objects.
\section{Related Works}
\label{sec:related_works}
\noindent\textbf{Object localization/grounding with VLMs.}
Vision Language Models 
have demonstrated strong ability in general visual understanding tasks.
In localization-related tasks, VLMs are commonly used for reference grounding~\cite{chen2023shikra,li2025rau,Yin_2025_CVPR,zhan2025griffon}.
Segmentation is also related to localization 
\cite{lai2024lisa,lan2024text4seg}.
Reasoning-based approach has been proposed to further enhance the visual understanding performance of VLMs~\cite{shen2025vlm,ren2024pixellm}.
To improve localization performance of VLMs, prior work~\cite{lin2024training,liu2025seg,kang2025your,lan2024text4seg,shen2024groundvlp} mainly uses external models such as SAM~\cite{kirillov2023segment} or object detectors~\cite{zhou2022detecting}.
In this work, we study LVLMs such as Qwen2.5-VL~\cite{bai2025qwen2} and InternVL-3.5~\cite{wang2025internvl3} that can directly output bounding box coordinates, and improve this ability without external models or finetuning.

\noindent\textbf{Attention in LLMs.}
Recent work~\cite{bi2025unveiling, zhang2025attention} shows that attention mechanisms
are key to understanding how LLMs process and ground information from image tokens~\cite{wu2024controlmllm,chen2025spatial}.
Building on this insight, many studies~\cite{kang2025your, lin2024training, wu2025f} further exploit the structure of attention patterns, using attention maps as effective localization priors and feeding them into lightweight localizers~\cite{kirillov2023segment} for downstream prediction.
Zhang et al.~\cite{zhang2023towards} use cross-modal attention to crop question-relevant regions, notably improving zero-shot VQA on small details.
In this work, we reveal new insights into attention patterns and their connection to localization quality in LVLMs.


\noindent\textbf{Inference-time sampling.}
Test-time scaling is a powerful training-free strategy for improving model performance. Early work shows that sampling multiple responses and aggregating them improves accuracy~\cite{wang2022self}, with self-consistency~\cite{wang2022self} using majority voting over diverse reasoning paths. Subsequent extensions explore weighted voting~\cite{li2023making}, confidence-based selection~\cite{xiong2023can}, and minimum Bayes risk decoding~\cite{suzgun2023follow}. Recent advances in compute allocation~\cite{snell2024scaling, AllenZhu-icml2024-tutorial} demonstrate that strategic sampling can rival larger models, with Best-of-N sampling~\cite{lightman2023let, cobbe2021training} and tree-search methods~\cite{yao2023tree, li2025mits, qi2024mutual} achieving strong gains in math reasoning. In vision-language settings, diverse sampling 
benefits 
visual chain-of-thought prompting~\cite{chen2024visual} and fine-grained reasoning~\cite{shi2025enhancing}. However, its use in object localization remains underexplored. Token-level vocabulary entropy has been explored as a proxy for model confidence~\cite{xiong2023can}, but it captures a fundamentally different signal from spatial attention patterns used in our work.
\section{Preliminary}

\noindent\textbf{Object grounding with LVLMs}.
Given an image $\mathbf{I} \in \mathbb{R}^{H \times W \times 3}$ and a text query $q$ for object localization, \model predict a bounding box of the queried object, represented as $b = [x_1, y_1, x_2, y_2]$ when the object is present in the image.
For challenging small objects, the boxes are inaccurate.

\noindent\textbf{Inference-time sampling for localization}.
One way to improve localization is to generate multiple responses
\cite{yao2023tree,qi2024mutual, cheng2025vision}.
As the number of responses increases, the chance that at least one box is good
also increases.
We can generate diverse boxes through temperature-controlled sampling:
$
    \mathcal{R} = \{r_1, r_2, \ldots, r_N\} \sim p_{\text{\model}}(\cdot | \mathbf{I}, q, \tau) ,
$
where there are $N$ responses and $\tau$ is the temperature. Each response $r_i$ generates $M_i$ bounding boxes, forming a \textit{candidate set} $\mathcal{B} = \{b_1, b_2, \ldots, b_T\}$ where $T = \sum_{i=1}^{N} M_i$.
The key challenge in inference-time sampling is to \textit{select the best bounding box from $\mathcal{B}$}. The bottleneck is not generating candidates but knowing which one is reliable---and we show that internal attention patterns are the key.

\section{Methodology}
\label{sec:method}
\begin{figure*}[t!]
    \centering
    \includegraphics[width=0.93\linewidth]{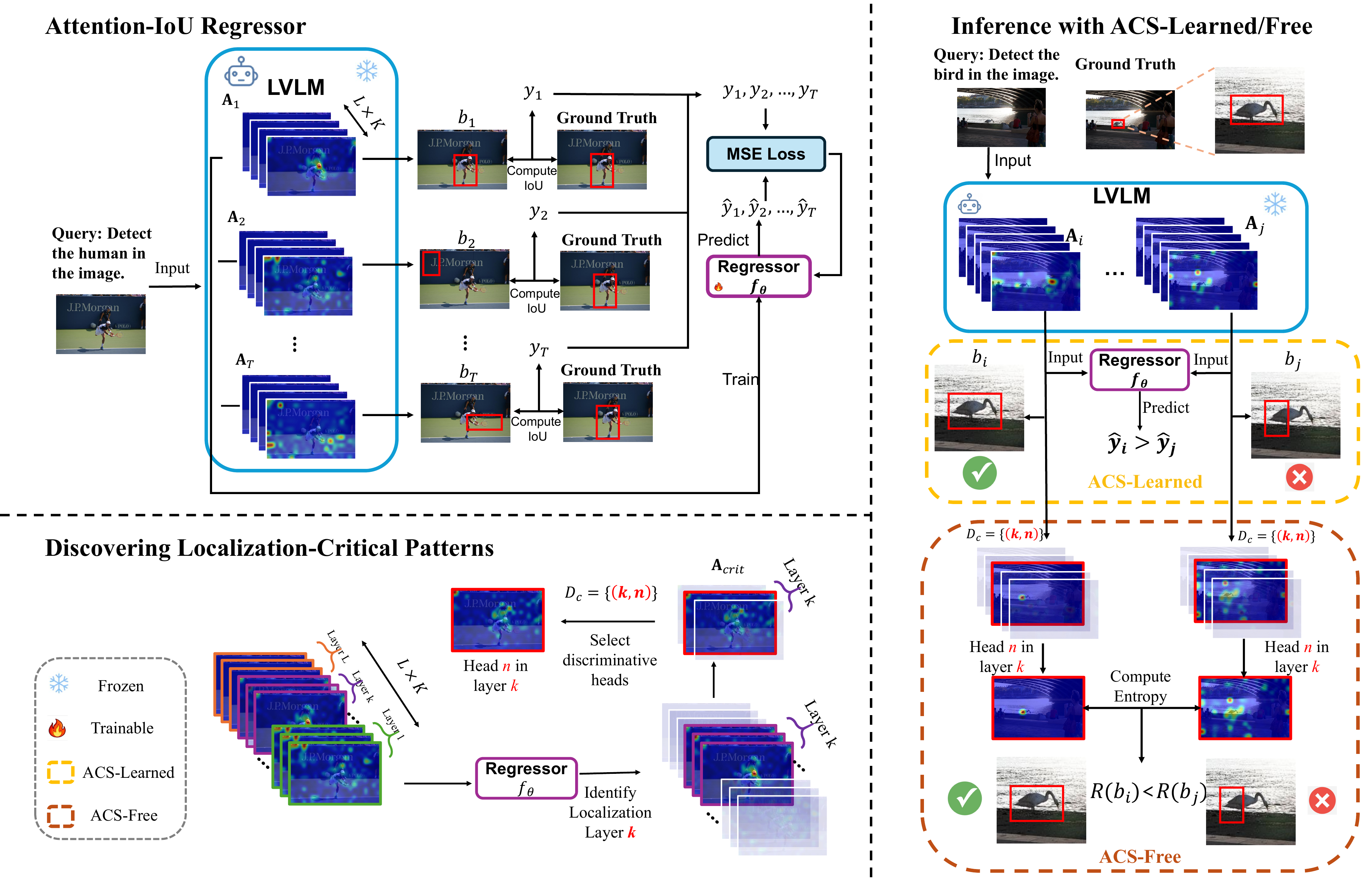}
    \caption{
    Our pipeline follows a discovery arc in three stages: (1) training a lightweight \textbf{IoU regressor} on attention maps to test the hypothesis that attention encodes localization quality
    and deploying as \textbf{ACS-Learned}; (2) analyzing the trained regressor via gradient attribution and entropy analysis to identify localization-critical layers and discriminative heads; and (3) distilling this understanding into \textbf{ACS-Free}, a training-free selector that ranks candidates by attention entropy on the identified heads, requiring no learned component at inference.
    }
    \label{fig:pipeline}
\end{figure*}

\subsection{Overview}

To select high-quality bounding boxes from the candidate set $\mathcal{B}$, we exploit the fact that coordinates are generated through autoregressive decoding where the LVLM attends to visual regions.
These attention patterns of visual regions should reflect the model's spatial understanding,
distinguishing good predictions from bad ones.
However, which specific attention characteristics are most informative for grounding quality, and whether they can be used without a learned model, remains unclear.


We address this through a three-stage investigation, formalized as the \textbf{Attention-based Candidate Selection (ACS)} framework (Figure~\ref{fig:pipeline}).
\underline{Stage 1}: we train a lightweight IoU regressor on attention maps to test the hypothesis that attention encodes localization quality (§\ref{sec:arm}); we deploy this regressor directly as \textbf{ACS-Learned}.
\underline{Stage 2}: we analyze the trained regressor via gradient attribution and entropy analysis to discover which layers and heads drive localization quality (§\ref{sec:attribution}).
\underline{Stage 3}: we distill this understanding into \textbf{ACS-Free}, a training-free variant that ranks candidates by entropy on the identified discriminative heads, requiring no learned component at inference (§\ref{sec:DES}).

\subsection{Attention-IoU Regressor \& ACS-Learned}
\label{sec:arm}
To identify which attention patterns correlate with localization quality, we directly train a lightweight \textit{Attention-IoU Regressor} $f_\theta: \mathbf{A} \rightarrow \hat{Y}$ that predicts IoU scores from attention patterns $\mathbf{A}$, where $Y \in [0,1]$ denotes the Intersection over Union (IoU) between prediction and ground truth.

\noindent\textbf{Attention extraction.}
For each candidate bounding box $b_j \in \mathcal{B}$, we extract their attention maps from all layers and heads. 
Specifically, we consider an LVLM with $L$ layers and $K$ attention heads per layer.
When generating each coordinate token $c \in \{x_1, y_1, x_2, y_2\}$, we extract the attention weights over vision tokens:
$
\mathcal{A}_{c,j}^{(l,h)}\in\mathbb{R}^{H_v\times W_v},
$
where $l\in[1, L]$ is the layer index, $h \in [1, K]$ is the head index, and $(H_v, W_v)$ are the spatial dimensions of the vision token grid.
We aggregate attention across all layers and heads for each coordinate:
$
\mathbf{A}_{c, j} = [\mathcal{A}_{c,j}^{(1,1)}, \mathcal{A}_{c,j}^{(1,2)},
\ldots, \mathcal{A}_{c,j}^{(L,K)}] \in \mathbb{R}^{L \times K \times H_v \times W_v}.
$
The representation for $b_j$ is:
$
\mathbf{A}_j = \{\mathbf{A}_{x_1,j}, \mathbf{A}_{y_1,j}, \mathbf{A}_{x_2,j}, \mathbf{A}_{y_2,j}\}.
$

\noindent\textbf{Regressor training.}
The IoU regressor $f_\theta$ predicts IoU scores through a neural network parameterized by $\theta$: $\hat{y}_j=f_\theta(\mathbf{A}_{j})\in[0,1]$. Given a training dataset $\mathcal{D}$ of (bounding box, IoU) pairs collected from LVLM inference, we train $f_\theta$ with the mean squared error:
\begin{equation}
\theta^* = \arg\min_{\theta}\frac{1}{|\mathcal{D}|}
\sum_{(b_j,y_j)\in\mathcal{D}}(y_j-\hat{y}_j)^2.
\label{eq:mse}
\end{equation}
Under the MSE loss, the optimal predictor learns the conditional expectation $f_\theta(\mathbf{A}) \approx \mathbb{E}[Y|\mathbf{A}]$. 

MSE training also has a principled information-theoretic justification~\cite{cover1999elements}.
The mutual information $I(f_\theta(\mathbf{A});Y) = H(Y) - H(Y|f_\theta(\mathbf{A}))$ quantifies how much uncertainty about IoU is reduced by observing the regressor's predictions. When we model the regression error as Gaussian noise, i.e., $Y = f_\theta(\mathbf{A}) + \varepsilon$ with $\varepsilon \sim \mathcal{N}(0, \sigma^2)$, the conditional entropy becomes $H(Y|f_\theta(\mathbf{A})) = \frac{1}{2}\log(2\pi e \sigma^2)$, where $\sigma^2=\mathbb{E}[(Y-\hat{Y})^2]$ is precisely the MSE (detailed derivation in Appendix~\ref{sec:differential-entropy}). Since $H(Y)$ is constant for a given dataset, minimizing MSE is equivalent to minimizing $H(Y|f_\theta(\mathbf{A}))$ as well as maximizing $I(f_\theta(\mathbf{A});Y)$. 
This result shows that the IoU regressor not only predicts IoU accurately, but also \textit{preserves localization-relevant information} contained in attention patterns.


\noindent\textbf{Inference with ACS-Learned.}
At inference time, the trained regressor is a natural solution for bounding box selection.
Given the candidate set $\mathcal{B}$
, we apply the IoU regressor to predict the IoU score for each candidate based on its attention patterns.
We refer to this regressor-based inference strategy as \textbf{ACS-Learned}:
$b^*=\arg\max_{b_j\in\mathcal{B}} f_\theta(\mathbf{A}_j).$
It ranks all candidates by their predicted IoUs and selects the one with the highest estimated quality.

\begin{figure*}[t]
    \centering
    \setlength{\tabcolsep}{1pt} 
    \renewcommand{\arraystretch}{0.8} 

    \begin{tabular}{@{}c@{\hspace{2pt}}c@{\hspace{2pt}}c@{\hspace{2pt}}c@{\hspace{2pt}}c@{\hspace{2pt}}c@{}}
        \includegraphics[width=0.16\textwidth]{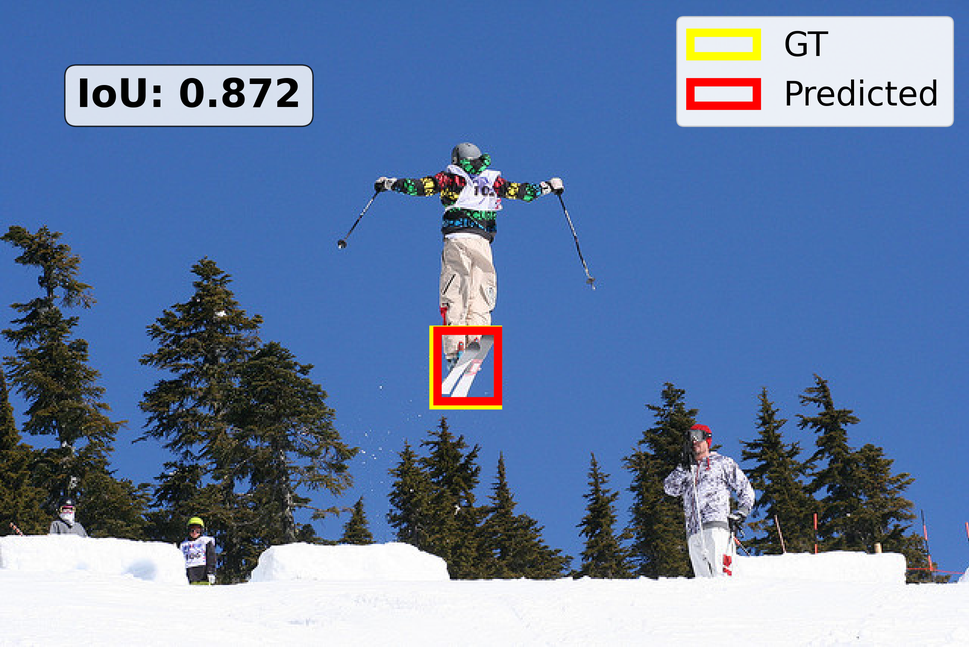} &
        \includegraphics[width=0.16\textwidth]{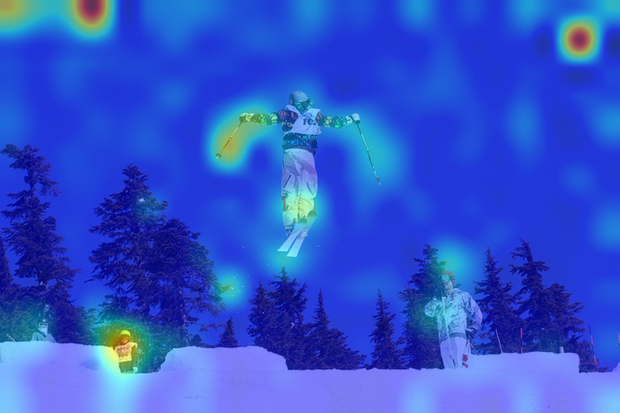} &
        \includegraphics[width=0.16\textwidth]{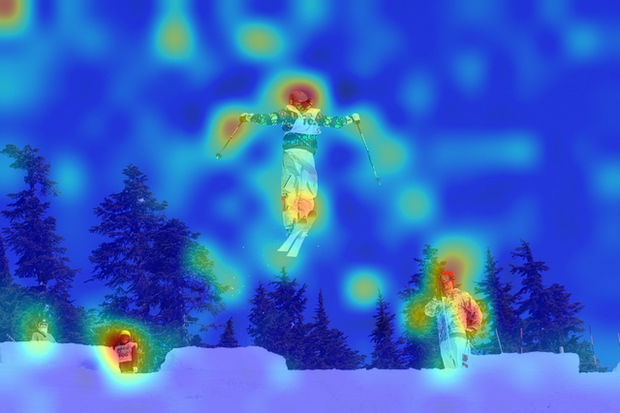} &
        \includegraphics[width=0.16\textwidth]{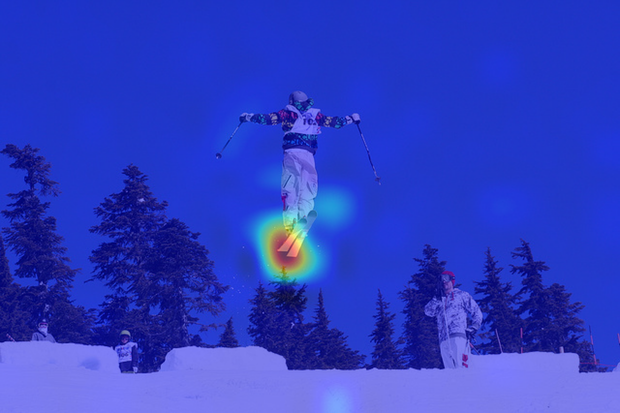} &
        \includegraphics[width=0.16\textwidth]{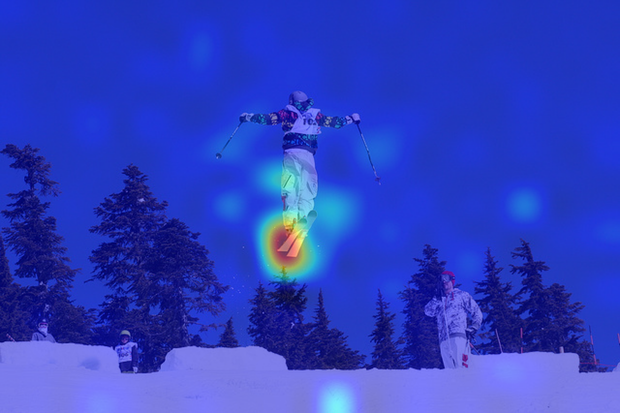} &
        \includegraphics[width=0.16\textwidth]{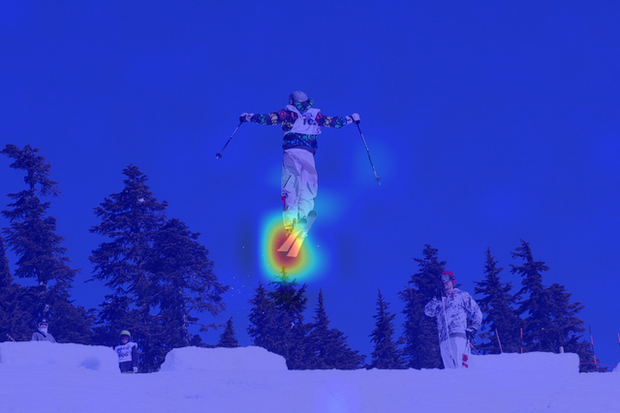} \\

        \small Original & \small Layer 5 & \small Layer 10 & \small Layer 16 & \small Layer 17 & \small Layer 18 \\[8pt]

        \includegraphics[width=0.16\textwidth]{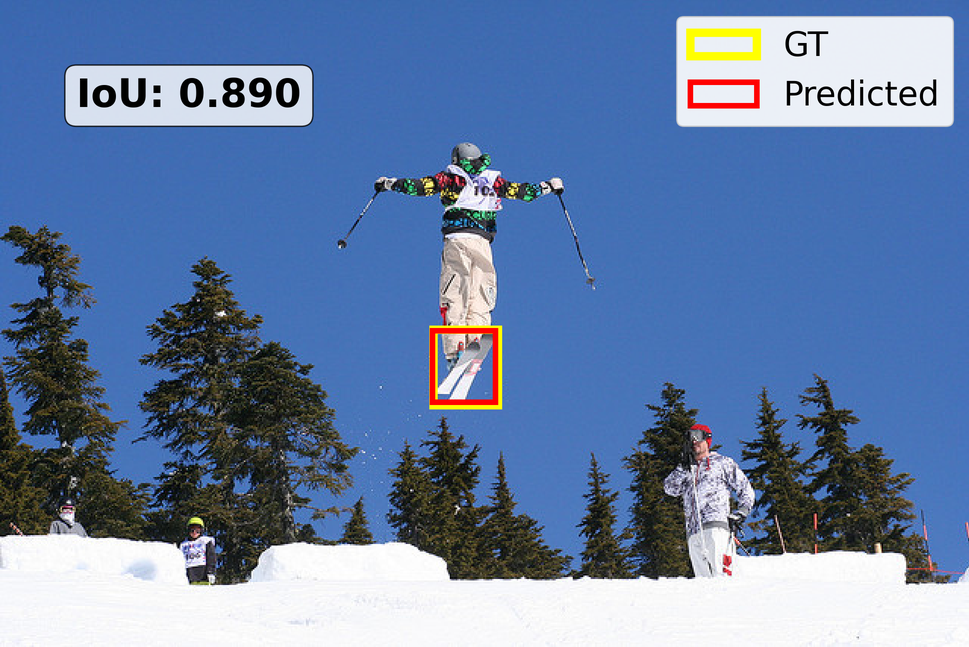} &
        \includegraphics[width=0.16\textwidth]{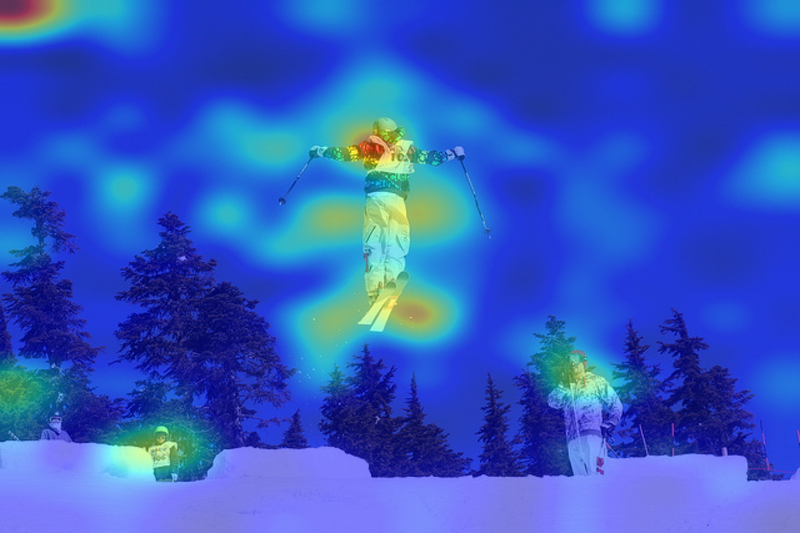} &
        \includegraphics[width=0.16\textwidth]{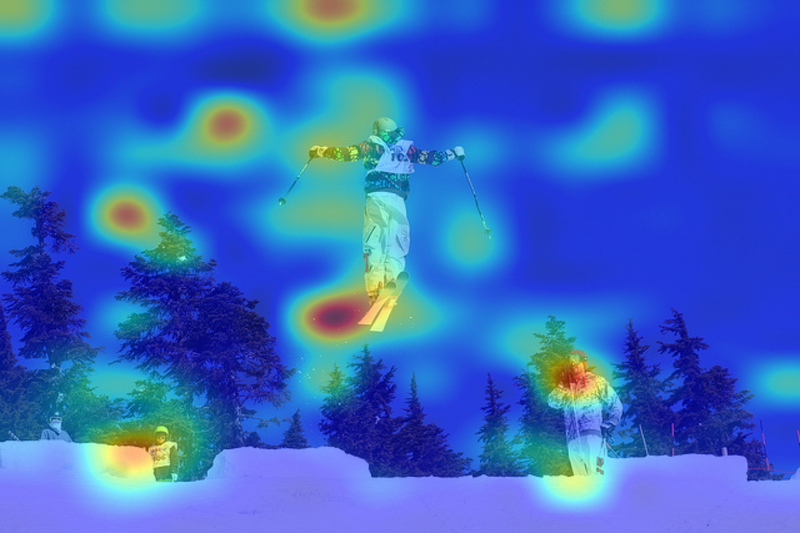} &
        \includegraphics[width=0.16\textwidth]{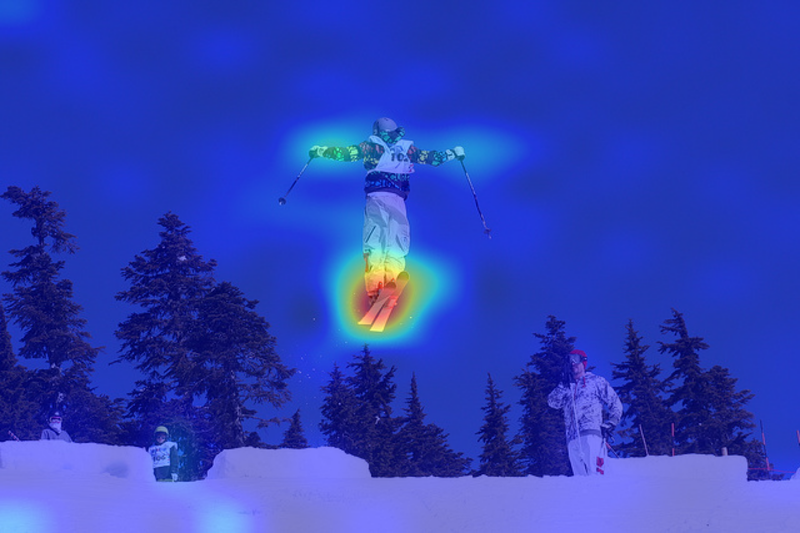} &
        \includegraphics[width=0.16\textwidth]{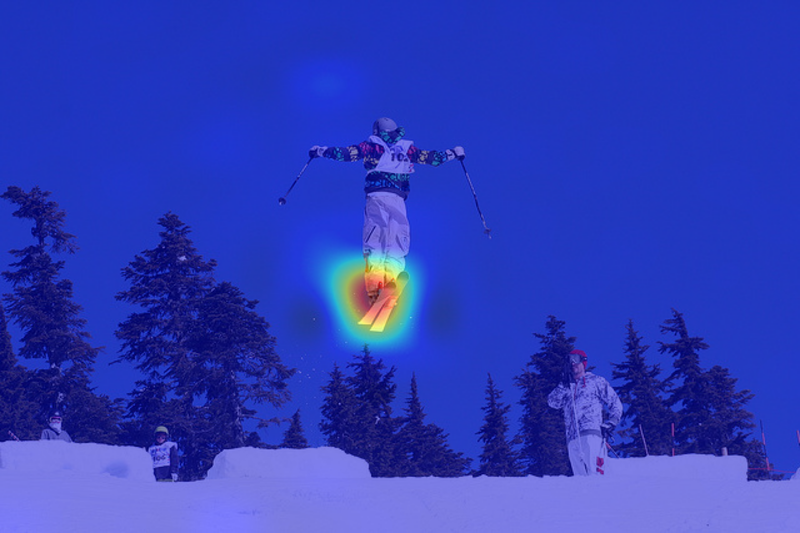} &
        \includegraphics[width=0.16\textwidth]{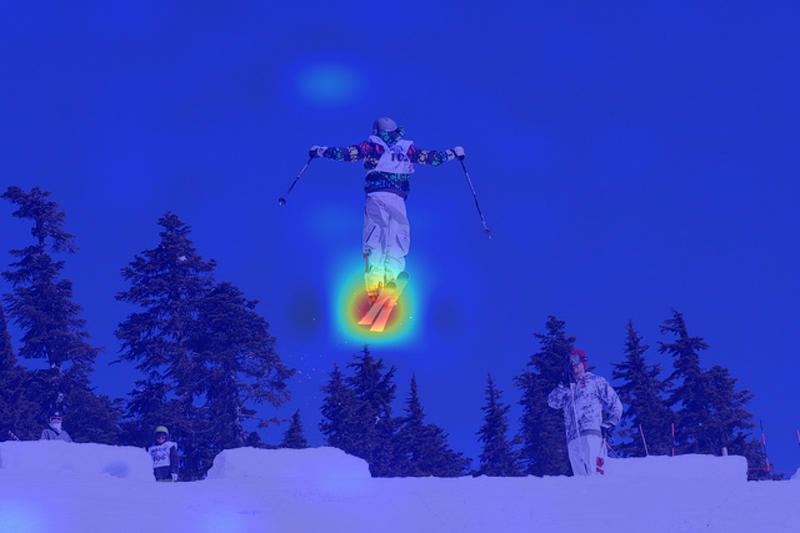} \\

        \small Original & \small Layer 8 & \small Layer 12 & \small Layer 22 & \small Layer 23 & \small Layer 24
    \end{tabular}

    \caption{Attention maps across important layers for Qwen2.5-VL (first row)
    and InternVL-3.5 (second row). The first column shows the original images with bounding boxes, followed by attention visualizations from five different layers.}
    \label{fig:important_layers_attention_maps}
\end{figure*}


\subsection{Discovering Localization-Critical Patterns}
\label{sec:attribution}
While ACS-Learned is effective for object grounding, it requires the learned model during inference.
To remove this dependency, we analyze what the trained IoU regressor has learned and translate those findings into training-free selection rules.
Our analysis proceeds in two steps. First, we identify the transformer layers that contribute most to localization prediction through gradient attribution. Second, we examine the attention characteristics within these layers that correlate with localization to obtain usable quantities for ACS-Free (§\ref{sec:DES}). 

\begin{table}[!ht]
\centering
\small
\setlength{\tabcolsep}{6pt}
\renewcommand{\arraystretch}{1.15}
\caption{Top-10 localization-critical layers.}
\label{tab:important_layers}
\begin{tabular}{l|l}
\toprule
\textbf{Model} & \textbf{Layer Index} \\
\midrule
Qwen2.5-VL-7B & 14, 15, 16, 17, 18, 19, 20, 21, 24, 25 \\
InternVL-3.5-8B  & 17, 18, 19, 20, 21, 22, 23, 24, 26, 35 \\
\bottomrule
\end{tabular}
\end{table}


\noindent\textbf{Localization-critical layers.} To assess the contribution of different layers to localization-related signals, we analyze gradients from the trained IoU regressor~\cite{simonyan2013deep,selvaraju2020grad,yang2025concept}. Specifically, for each layer $l$ and head $h$ in an LVLM, we compute the importance score for coordinate $c \in \{x_1, y_1, x_2, y_2\}$:
\begin{equation}
    I_c^{(l,h)} = \max_{i,j} \left| \frac{\partial \mathcal{L}_{\text{MSE}}(f_\theta(\mathbf{A}), y)}{\partial \mathcal{A}_{c,i,j}^{(l,h)}} \right|.
\end{equation}
The per-coordinate layer importance $I_c^{(l)} = \frac{1}{K}\sum_{h=1}^{K} I_c^{(l,h)}$ captures which layers most affect IoU predictions.

The top-10 important layers for both Qwen2.5-VL-7B and InternVL-3.5-8B are reported in Table~\ref{tab:important_layers}.
For Qwen2.5-VL, layers 14--21 contribute the most, which we identify as \textit{localization-critical layers} $\mathbf{A}_{\text{crit}}$.
This finding is confirmed by attention visualizations in Figure~\ref{fig:important_layers_attention_maps}: layers 16-18 focus precisely on target objects, while earlier layers ($\leq 12$) show dispersed patterns, suggesting a global-to-object transition
(more in Appendix~\ref{sec:complete_attention_maps}).

The concentration of gradient magnitudes in layers 14--21 indicates that mutual information $I(\mathbf{A}; Y)$ is predominantly captured by these layers. In other words, if $\mathbf{A}_{\text{crit}}$ denotes attention from these critical layers, then $I(\mathbf{A}_{\text{crit}}; Y) \approx I(\mathbf{A}; Y)$.
This naturally leads to a hypothesis: within localization-critical layers, accurate bounding boxes correspond to concentrated (low-entropy) attention, while poor predictions yield diffuse (high-entropy) attention.
\begin{wrapfigure}{r}{0.46\linewidth}
    \centering
    \vspace{-\baselineskip}
    \includegraphics[width=\linewidth]{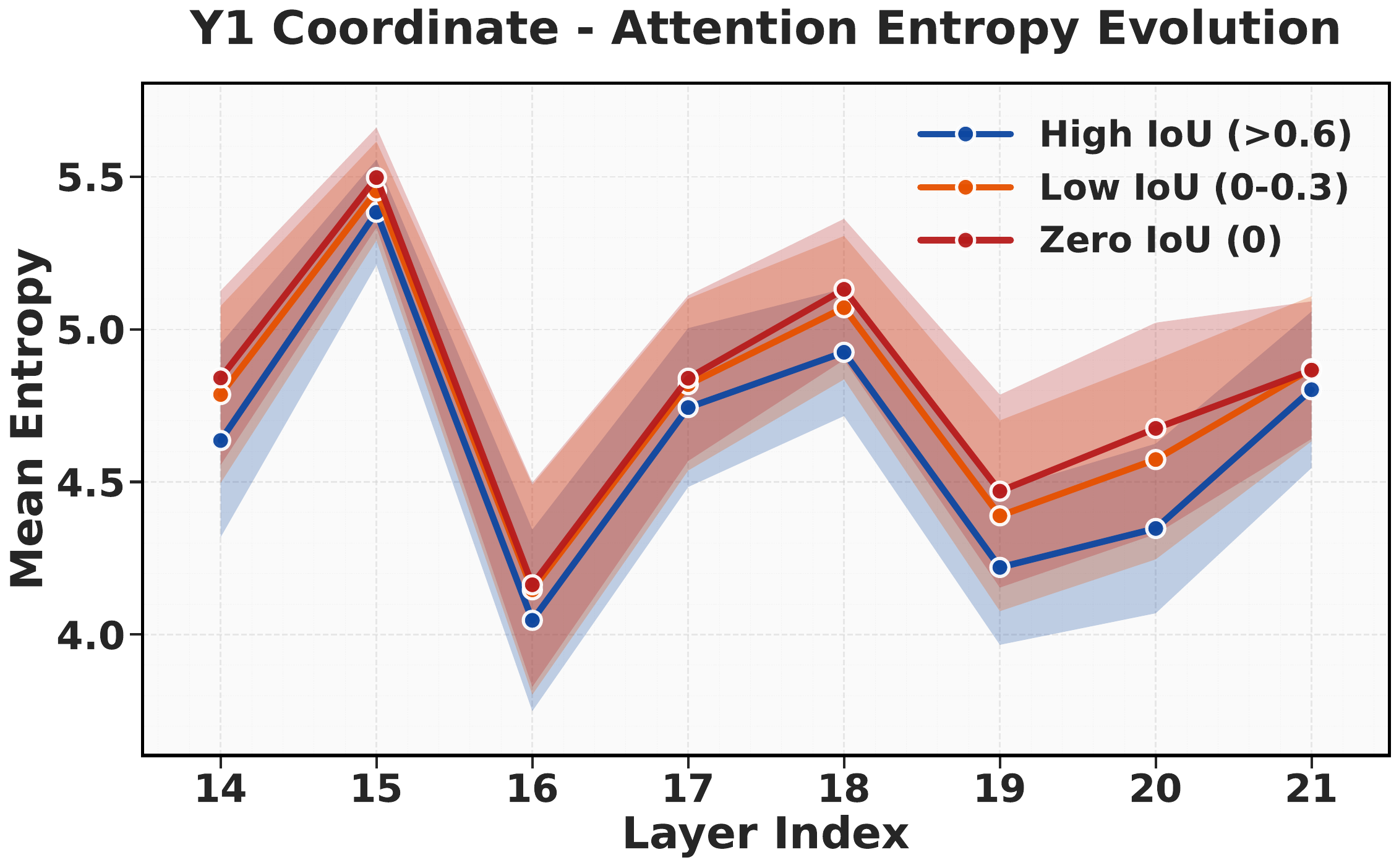}
    \caption{Attention entropy in localization-critical layers. High-IoU predictions have lower entropy than low/zero-IoU predictions.}
    \label{fig:y1_important_layer_analysis}
    \vspace{-\baselineskip}
\end{wrapfigure}

\noindent\textbf{Discovering the entropy-IoU correlation.}
To test the hypothesis that attention concentration reflects localization quality, we examine how attention spatial entropy within the localization-critical layers correlates with IoU scores. Specifically, we compute the entropy of attention maps for bounding boxes of varying quality and compare their distributions.
The entropy of an attention map measures its spatial concentration:
\begin{equation}
    H(\mathcal{A})=-\sum_{u=1}^{H_v} \sum_{v=1}^{W_v} p_{u,v} \log p_{u,v},
\end{equation}
where $p_{u,v} = \mathcal{A}_{u,v} / \sum_{k,l} \mathcal{A}_{k,l}$ normalizes the attention map into a probability distribution.
We partition training samples into three groups based on IoU:
$
    \mathcal{G}_{\text{high}}= \{b_j: \text{IoU}(b_j) > \tau_{\text{high}}\},
    \mathcal{G}_{\text{low}}=\{b_j: 0 < \text{IoU}(b_j) \leq \tau_{\text{low}}\},
    \mathcal{G}_{\text{zero}}=\{b_j: \text{IoU}(b_j)=0\},
$
where $\tau_{\text{low}} < \tau_{\text{high}}$ are predefined thresholds.
For each (layer $l$, head $h$) and each coordinate $c$, we compute the per-head mean entropy across samples in IoU group $g$:
$
    \bar{H}_{c,g}^{(l, h)}
    =
    \frac{1}{|\mathcal{G}_g|}
    \sum_{b_j \in \mathcal{G}_g}
    H\!\left(\mathcal{A}_{c, j}^{(l,h)}\right),
$
and aggregate to the layer level by averaging over the $K$ heads of layer $l$:
$
    \bar{H}_{c,g}^{(l)} = \tfrac{1}{K} \sum_{h=1}^{K} \bar{H}_{c,g}^{(l,h)},
$
which is what Figure~\ref{fig:y1_important_layer_analysis} visualizes across the three IoU groups:
In localization-critical layers, \textit{high-IoU samples consistently exhibit lower entropy than low/zero-IoU samples}, making entropy a strong discriminator of localization accuracy in these layers.
Not all heads within these layers contribute equally to this separation.
To identify which heads are most discriminative, we measure the \textit{entropy difference}
between high- and low-IoU groups:
$
    \Delta H_c^{(l,h)}
    =
    \bar{H}_{c,\text{low}}^{(l,h)}
    -
    \bar{H}_{c,\text{high}}^{(l,h)}.
$
Heads with large $\Delta H_c^{(l,h)}$ are selected as the key indicators.


\subsection{ACS-Free: Training-Free Distillation}
\label{sec:DES}
Given the findings in §\ref{sec:attribution}, ACS-Free distills the knowledge learned by the IoU regressor into a parameter-free rule based on spatial entropy ranking.
Note that ACS-Free is not an independent method. Its localization-critical attention head selection is derived from the regressor's gradient and entropy analysis and would not be discoverable without first training and analyzing the regressor. 

For each coordinate $c \in \{x_1, y_1, x_2, y_2\}$, we first identify the $n$ most discriminative heads:
\begin{equation*}
\mathcal{D}_c = \{(l,h) \in \mathcal{U} \;|\; (l,h) \text{ in Top-}n(\Delta H_c^{(l,h)})\},
\end{equation*}
where $\mathcal{U} = \{(l,h) : 1 \le l \le L, 1 \le h \le K\}$ denotes all layer-head pairs.
During inference, ACS-Free evaluates each candidate $b_j$ by computing its average entropy across $\mathcal{D}_c$:
\begin{equation}
\bar{H}_{c}(b_j) = \frac{1}{|\mathcal{D}_c|} \sum_{(l,h)\in\mathcal{D}_c} H\left(\mathcal{A}_{c, j}^{(l,h)}\right).
\end{equation}
To combine information from all four coordinate tokens, ACS-Free converts entropy values into ranks.
For each coordinate $c$, we define $r_c(b_j) = \mathrm{rank}(\bar{H}_c(b_j))$, where lower entropy receives better (smaller) rank.
The overall score is: $R(b_j) = r_{x_1}(b_j) + r_{y_1}(b_j) + r_{x_2}(b_j) + r_{y_2}(b_j)$.
Finally, ACS-Free selects the candidate with the lowest overall rank:
$
b^{*} = \arg\min_{b_j \in \mathcal{B}} R(b_j).
$
The detailed ranking procedure is provided in Appendix~\ref{sec:appendix-ranking}.


\section{Experiments}

\subsection{Setup}

\paragraph{Datasets and protocol.}

\begin{table*}[!ht]
\centering
\small
\setlength{\tabcolsep}{2pt}
\renewcommand{\arraystretch}{0.9}
\caption{Object detection performance on COCO and Objects365 (single-object). We compare two variants of our \textbf{Attention-based Candidate Selection (ACS)} framework---\textbf{ACS-Learned}, the regressor-based selector, and \textbf{ACS-Free}, the training-free variant derived from analyzing the regressor's discriminative heads---against greedy decoding and six sampling-based baselines that select from $N{=}10$ sampled responses.
Best method per column in \textbf{bold}; best training-free method per column (excluding ACS-Learned) is \underline{underlined}.}
\label{tab:main_results}
\begin{tabular}{l|c|cc|cc||cc|cc}
\toprule
& & \multicolumn{4}{c||}{\textbf{Qwen2.5-VL-7B}} & \multicolumn{4}{c}{\textbf{InternVL-3.5-8B}} \\
\cmidrule(lr){3-6}\cmidrule(lr){7-10}
\multirow{2}{*}{\textbf{Method}} & \multirow{2}{*}{\textbf{$\tau$}} & \multicolumn{2}{c|}{\textbf{COCO}} & \multicolumn{2}{c||}{\textbf{Objects365}} & \multicolumn{2}{c|}{\textbf{COCO}} & \multicolumn{2}{c}{\textbf{Objects365}} \\
& & \textbf{Acc@0.5} & \textbf{mIoU} & \textbf{Acc@0.5} & \textbf{mIoU} & \textbf{Acc@0.5} & \textbf{mIoU} & \textbf{Acc@0.5} & \textbf{mIoU} \\
\midrule
Greedy & -- & 61.4 & 52.2 & 43.0 & 36.9 & 49.1 & 42.9 & 21.9 & 21.9 \\
\midrule
\multirow{3}{*}{Pass@1}
& 1.0 & 52.3 & 45.3 & 30.3 & 26.8 & 38.8 & 36.1 & 13.3 & 15.1 \\
& 0.7 & 59.3 & 50.3 & 36.5 & 32.1 & 45.0 & 40.4 & 16.0 & 17.3 \\
& 0.5 & 59.6 & 50.9 & 39.9 & 34.6 & 47.1 & 41.9 & 19.8 & 19.8 \\
\midrule
\multirow{3}{*}{FirstValid}
& 1.0 & 54.2 & 47.0 & 32.4 & 28.8 & 42.2 & 39.8 & 15.6 & 18.0 \\
& 0.7 & 60.3 & 51.2 & 38.1 & 33.5 & 48.8 & 44.1 & 18.3 & 20.4 \\
& 0.5 & 60.9 & 52.0 & 41.7 & 36.1 & 50.8 & 45.5 & 21.9 & 22.3 \\
\midrule
\multirow{3}{*}{TokEntropy}
& 1.0 & 51.7 & 43.8 & 32.0 & 28.1 & 39.1 & 37.2 & 11.9 & 14.7 \\
& 0.7 & 55.9 & 47.2 & 35.3 & 30.5 & 47.1 & 43.1 & 15.3 & 17.8 \\
& 0.5 & 58.1 & 49.0 & 37.7 & 32.8 & 49.1 & 44.2 & 20.3 & 20.5 \\
\midrule
\multirow{3}{*}{MajVote}
& 1.0 & 56.0 & 48.5 & 32.4 & 30.0 & 50.2 & 45.0 & 18.6 & 20.5 \\
& 0.7 & 59.2 & 51.0 & 37.6 & 33.5 & 51.7 & 46.4 & 21.8 & 22.7 \\
& 0.5 & 60.4 & 51.6 & 39.4 & 34.6 & \underline{53.8} & 46.8 & 23.2 & 23.4 \\
\midrule
\multirow{3}{*}{MeanBBox}
& 1.0 & 32.7 & 32.1 & 14.1 & 15.2 & 36.7 & 35.3 & 10.2 & 14.1 \\
& 0.7 & 45.4 & 41.5 & 22.9 & 22.7 & 41.8 & 39.0 & 14.2 & 17.2 \\
& 0.5 & 50.1 & 44.8 & 29.8 & 27.7 & 45.4 & 41.1 & 16.6 & 19.1 \\
\midrule
\multirow{3}{*}{Smallest}
& 1.0 & 27.0 & 25.5 & 18.6 & 18.3 & 18.3 & 24.1 & 6.2 & 10.5 \\
& 0.7 & 42.7 & 37.7 & 26.2 & 24.7 & 28.0 & 30.2 & 10.8 & 13.9 \\
& 0.5 & 49.3 & 42.7 & 32.1 & 29.5 & 34.3 & 34.2 & 13.6 & 16.6 \\
\midrule
\multirow{3}{*}{ACS-Free}
& 1.0 & 59.4 & 50.2 & 39.4 & 34.6 & 47.9 & 44.5 & 20.6 & 21.7 \\
& 0.7 & 62.1 & 52.3 & 42.6 & 37.1 & 51.3 & 46.5 & 23.3 & 23.5 \\
& 0.5 & \underline{63.4} & \underline{53.5} & \underline{43.0} & \underline{37.9} & 53.1 & \underline{47.4} & \underline{24.8} & \underline{24.4} \\
\midrule
\multirow{3}{*}{\textbf{ACS-Learned}}
& 1.0 & 64.0 & 53.0 & 41.5 & 36.5 & 55.0 & 48.3 & 21.9 & 22.5 \\
& 0.7 & 64.9 & 54.7 & 45.0 & \textbf{39.2} & 58.4 & 50.4 & 23.5 & 23.6 \\
& 0.5 & \textbf{65.3} & \textbf{55.2} & \textbf{45.1} & 39.1 & \textbf{58.6} & \textbf{50.6} & \textbf{25.5} & \textbf{24.9} \\
\bottomrule
\end{tabular}
\end{table*}

Our experiments focus on \emph{small objects} (0.1\%--1\% of image area), evaluating on 2,225 instances each from MS COCO~\cite{lin2014microsoft} (all qualifying cases in the validation set) and Objects365~\cite{shao2019objects365}.
For the multi-object setting, we evaluate 759 COCO cases with multiple target small objects.
Cases with no valid prediction are assigned IoU = 0.

\begin{figure}[t]
\centering
\setlength{\tabcolsep}{3pt}
\renewcommand{\arraystretch}{1.0}





\begin{tabular}{ccc}
\includegraphics[width=0.31\textwidth]{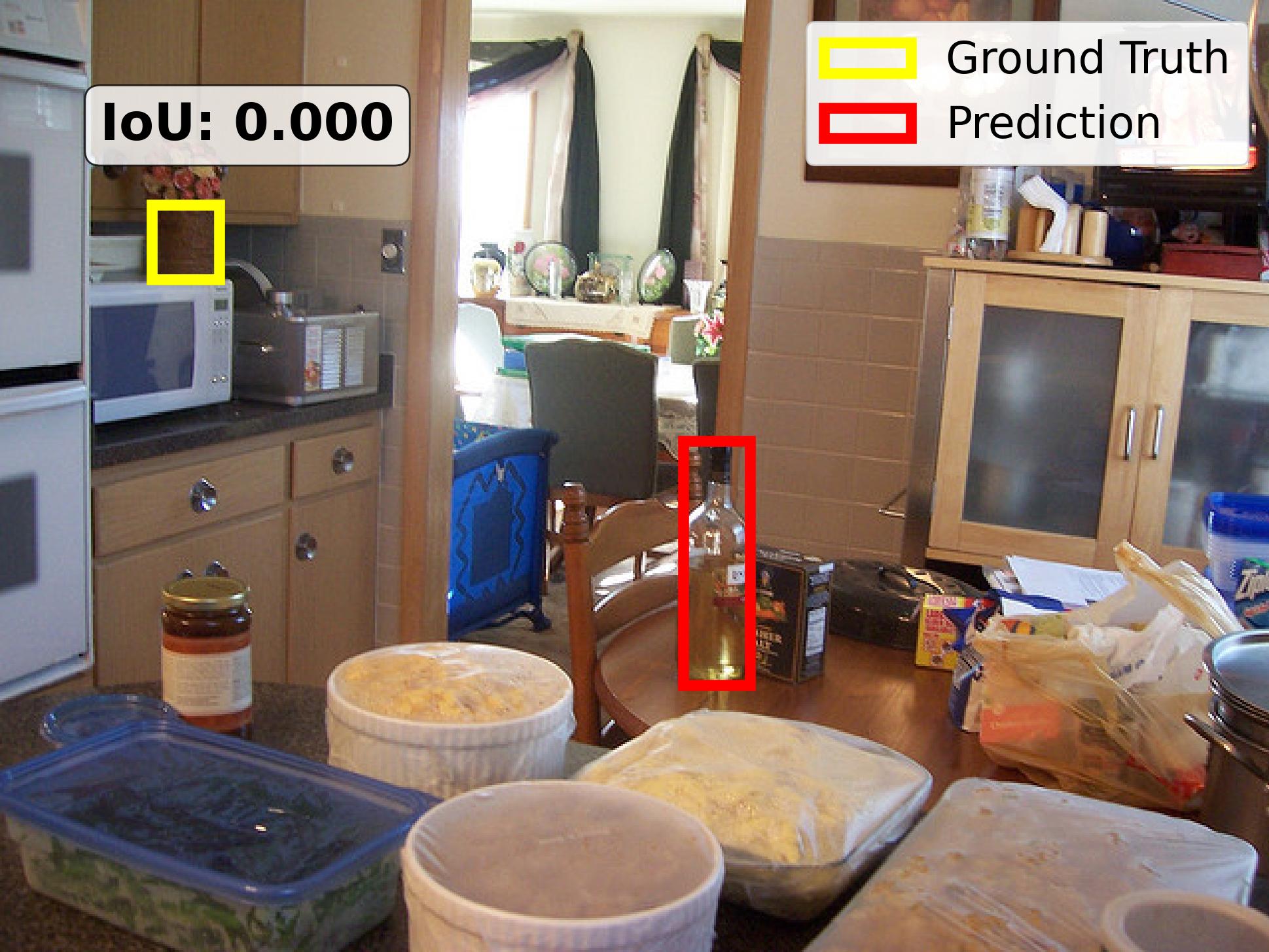} &
\includegraphics[width=0.31\textwidth]{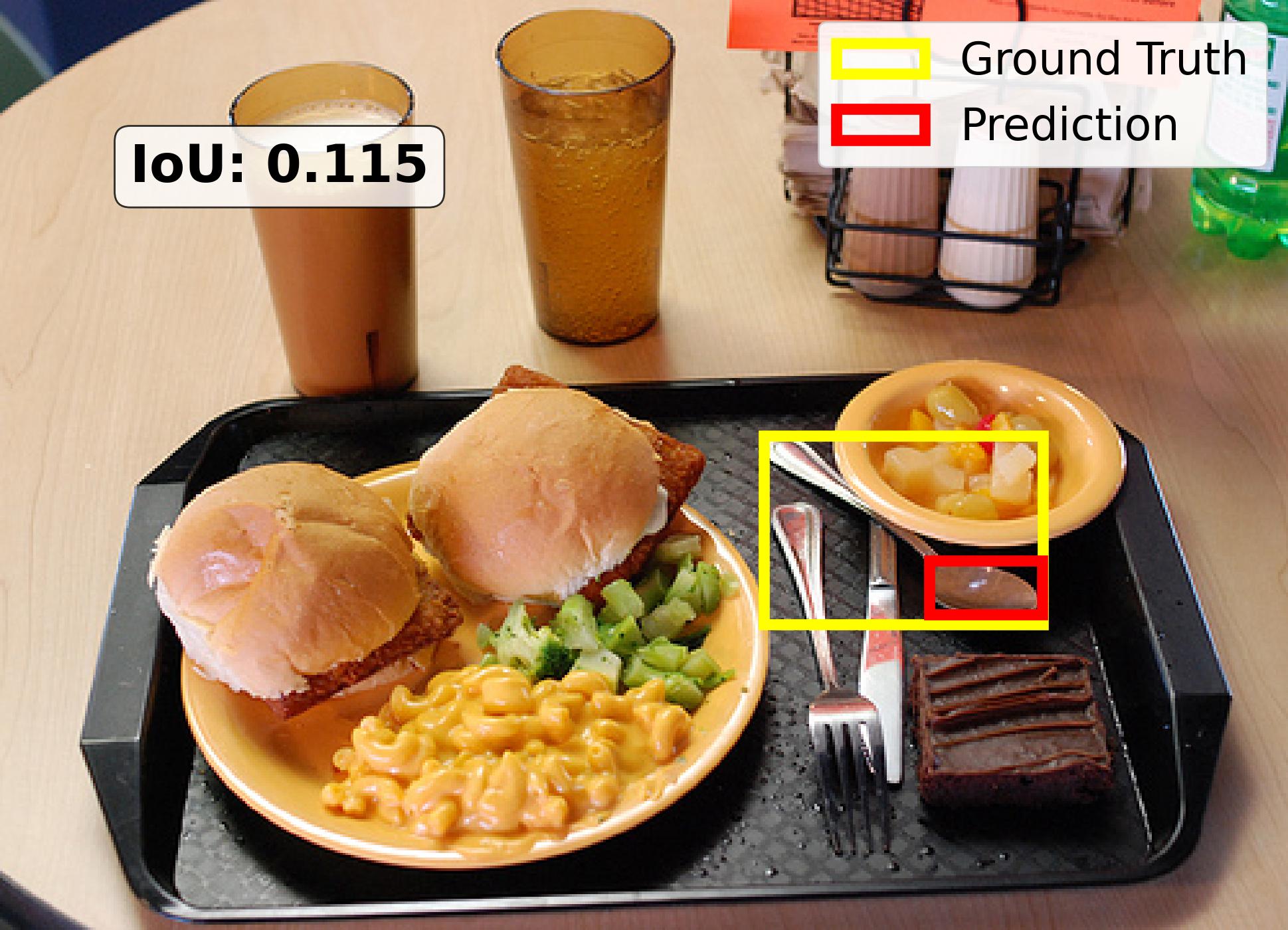} &
\includegraphics[width=0.31\textwidth]{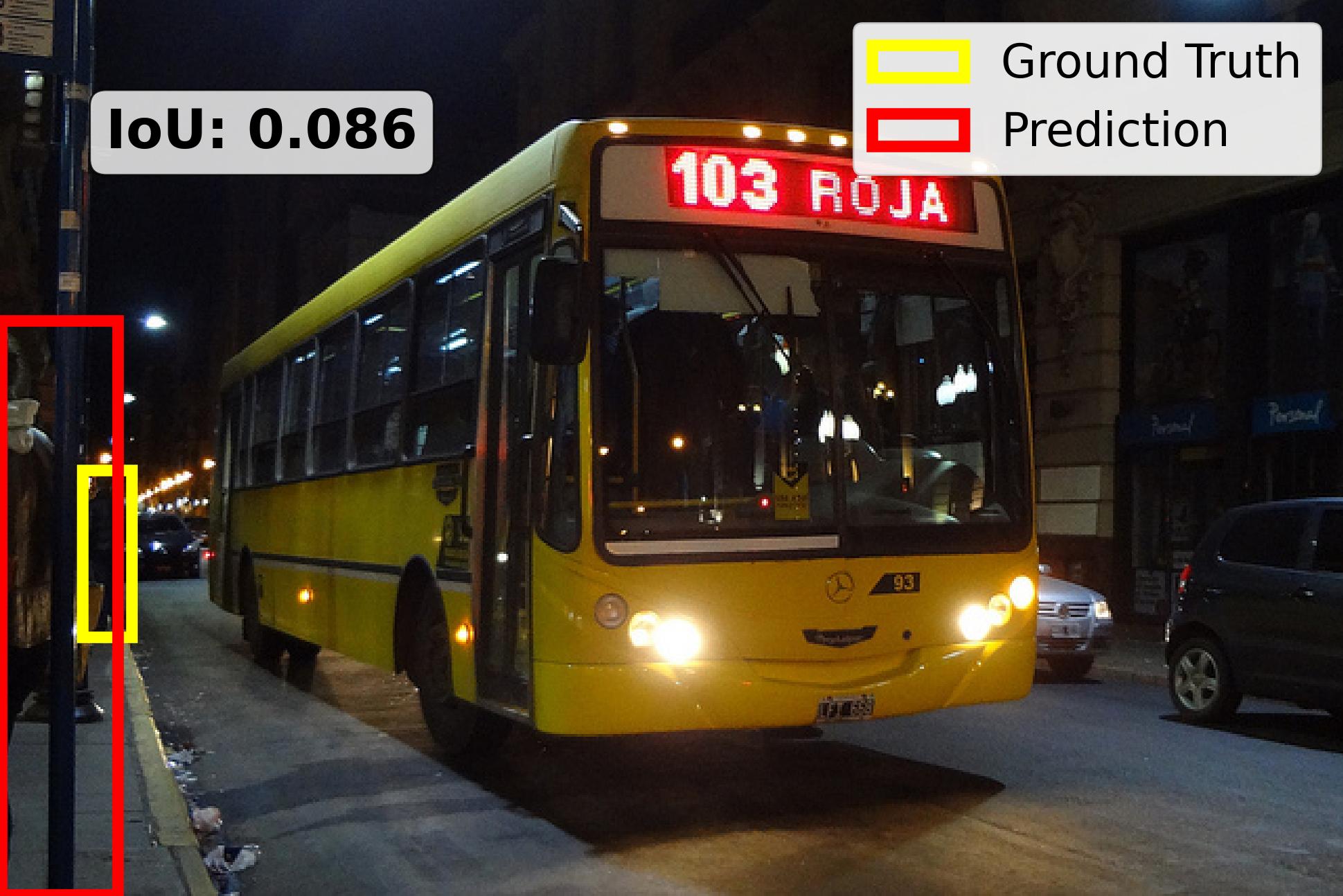} \\[2pt]
\includegraphics[width=0.31\textwidth]{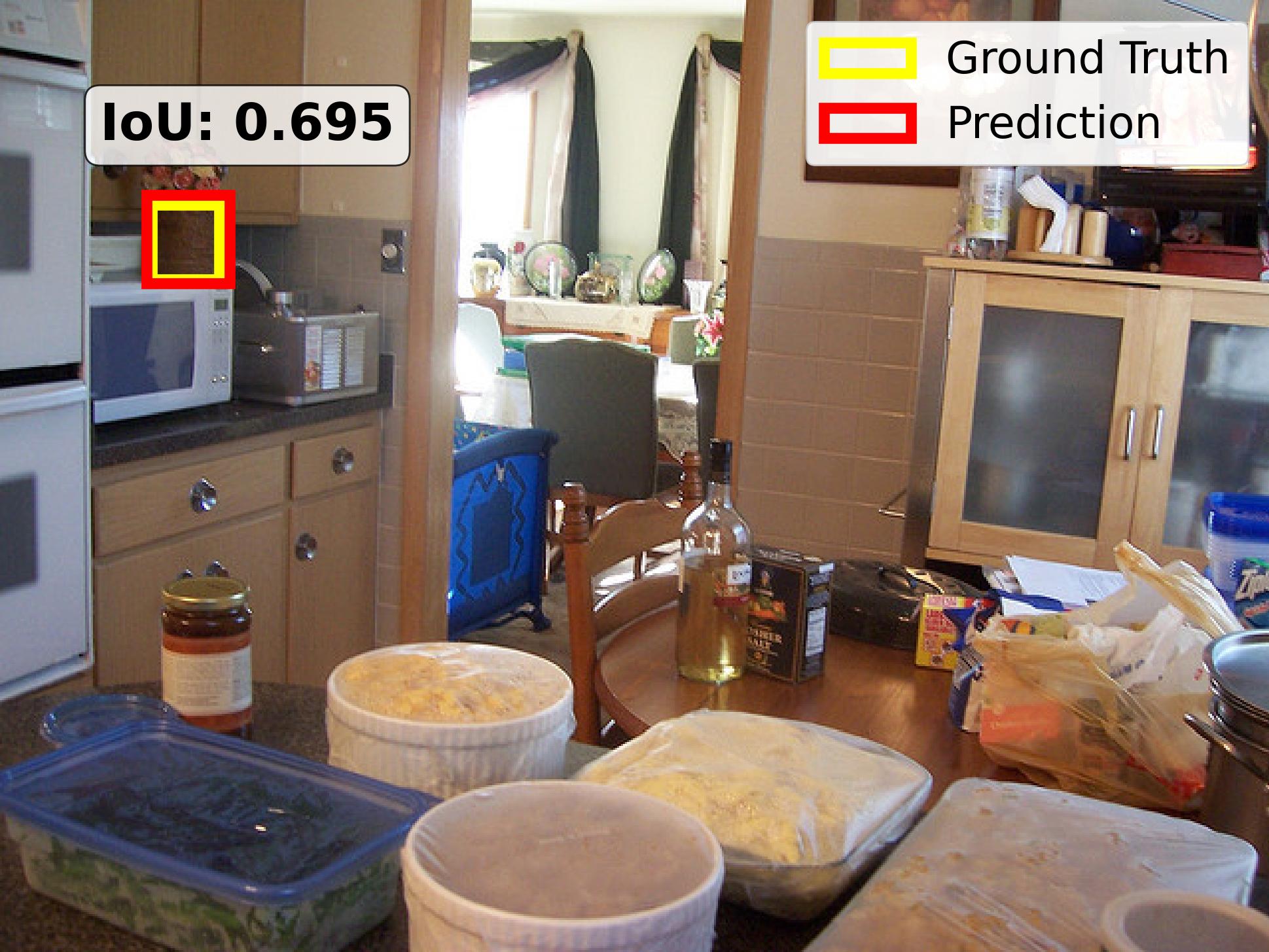} &
\includegraphics[width=0.31\textwidth]{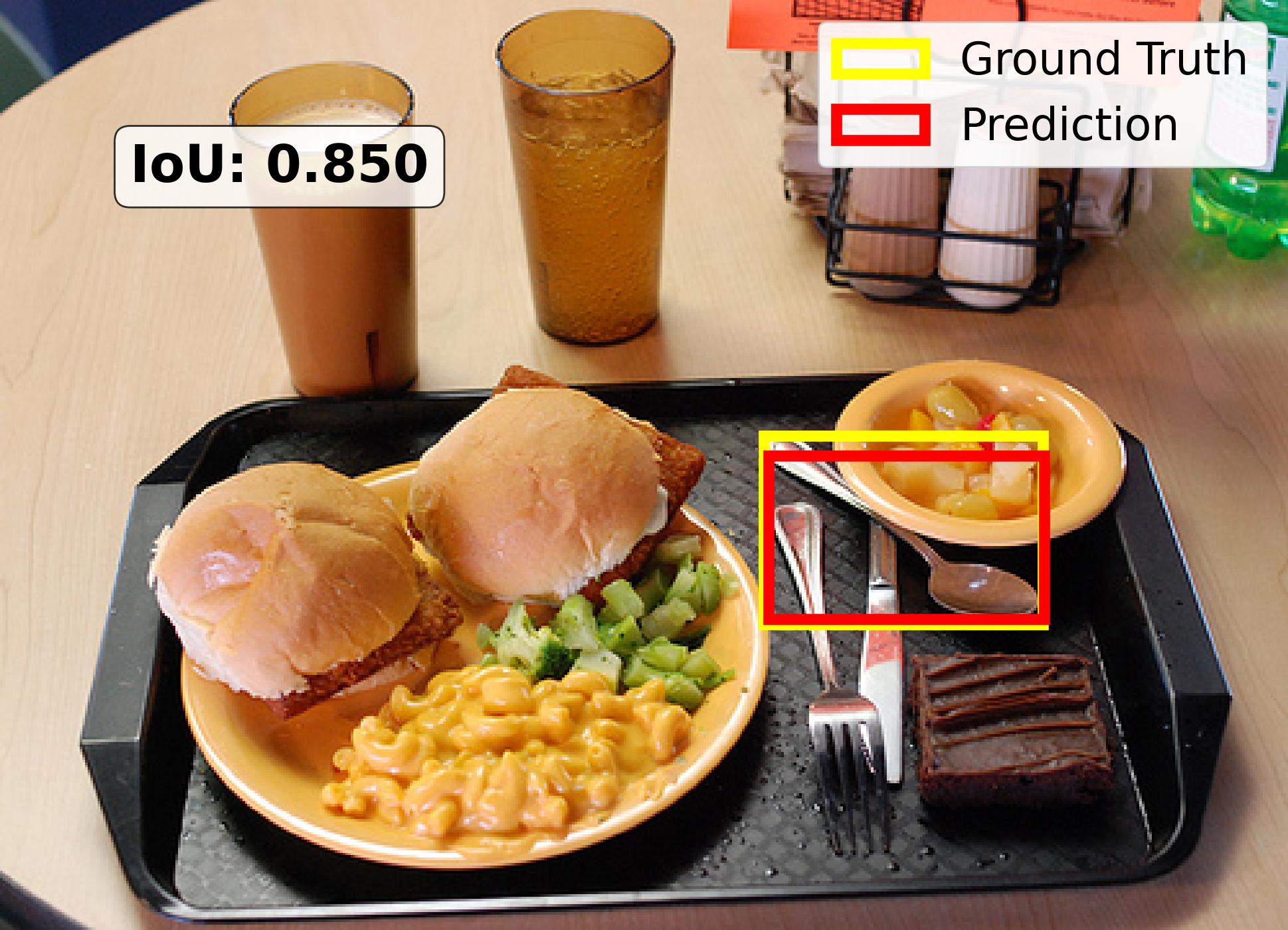} &
\includegraphics[width=0.31\textwidth]{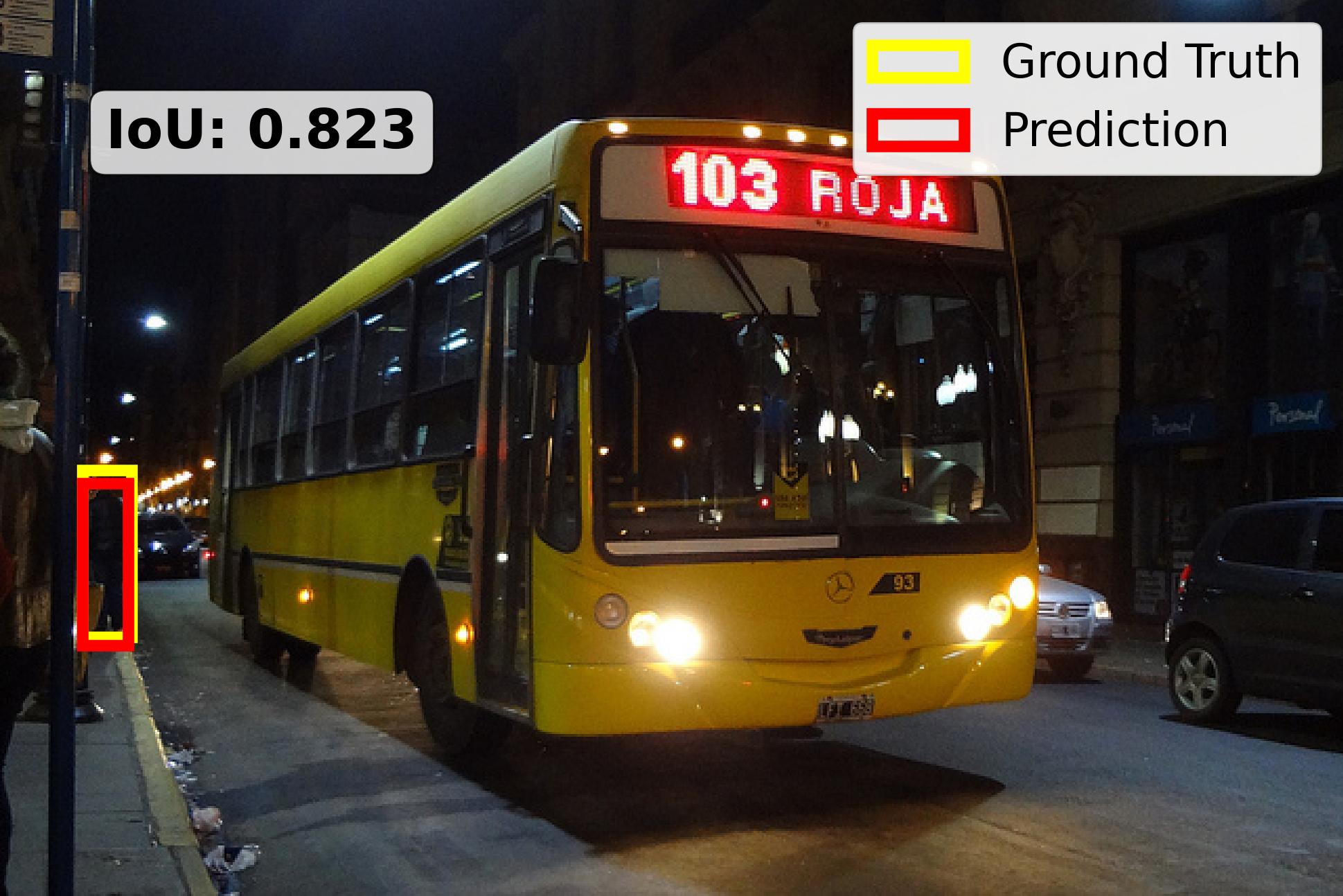} \\
\end{tabular}

\caption{
Qualitative comparison of localization.
Yellow boxes denote the ground-truth bounding box, and red boxes denote the predicted bounding box.
Each column corresponds to a different object category.
Top: Greedy Decoding; Bottom: Ours.
}
\label{fig:qualitative_results}
\end{figure}

\noindent\textbf{LVLMs and sampling details.}
We evaluate on Qwen2.5-VL-7B and InternVL-3.5-8B.
The IoU regressor is trained on the COCO training split; evaluation uses held-out instances from both COCO and Objects365.
We generate training/validation data using a temperature of $\tau = 1.0$, and all analyses of attention layers and heads are conducted on the validation subset.
For inference, we sample $N{=}10$ responses using
$\tau \in \{0.5, 0.7, 1.0\}$ with top-$p = 0.9$ and top-$k = 100$.
Based on gradient analysis, we focus on layers 14-21 for Qwen2.5-VL and layers 17-26 for InternVL-3.5 as localization-critical layers.
Our ACS-Free uses $n{=}3$ discriminative attention heads per coordinate.

\noindent\textbf{Attention extraction.}
\model generates tokens in raw outputs.
For each predicted box \([x_1,y_1,x_2,y_2]\), we locate the token
for each coordinate and use these four attention maps as the input of $f_\theta$.
We collect the attention assigned to the corresponding visual tokens of LVLMs during generation. More details are in Appendix~\ref{app:arm_settings}.

\noindent\textbf{Baselines.}
Since our approach performs self-improvement using only the LVLM's internal representations without retraining or external localization modules, we compare against baselines that similarly rely on the model's native outputs.
We start with (1) \textbf{Greedy Decoding (Greedy)}: standard deterministic decoding without sampling.
The remaining baselines all draw $N$ sampled responses and differ in how they pick a single output:
(2) \textbf{Pass@1} always returns response 1 with no validity check, capturing raw single-sample quality;
(3) \textbf{FirstValid} selects the first response containing a parseable bounding box;
(4) \textbf{TokEntropy}~\cite{xiong2023can, shi2025searchrag} ranks responses by the summed vocabulary entropy of the four bbox coordinate tokens and picks the lowest;
(5) \textbf{MajVote} synthesizes a bounding box by per-coordinate majority vote across candidates;
(6) \textbf{MeanBBox} synthesizes a bounding box from the per-coordinate arithmetic mean;
(7) \textbf{Smallest} selects the smallest-area candidate, designed to verify that our entropy-based selection is not simply favoring smaller boxes.




\noindent\textbf{Metrics.}
For single-object localization, we report Acc@0.5~\cite{kang2025your, wang2025internvl3} and Mean IoU (mIoU).
For the IoU regressor, we report MAE and Pearson $r$.
For the multi-object setting, we evaluate Precision, Recall, and F1.

\subsection{Regressor Performance}

\begin{wraptable}{r}{0.43\linewidth}
\centering
\vspace{-2.4\baselineskip}
\footnotesize
\setlength{\tabcolsep}{4pt}
\caption{IoU regression performance on COCO val.}
\label{tab:regression_performance}
\begin{tabular}{c|cc|cc}
\toprule
\multirow{2}{*}{\textbf{Temp}}
& \multicolumn{2}{c|}{\textbf{Qwen2.5-VL}}
& \multicolumn{2}{c}{\textbf{InternVL-3.5}} \\
& MAE & Pearson $r$
& MAE & Pearson $r$ \\
\midrule
0.5 & 0.165 & 0.737 & 0.180 & \textbf{0.682} \\
0.7 & 0.159 & 0.751 & 0.178 & 0.680 \\
1.0 & \textbf{0.148} & \textbf{0.780} & \textbf{0.176} & 0.677 \\
\bottomrule
\end{tabular}
\vspace{-1.5\baselineskip}
\end{wraptable}
Table~\ref{tab:regression_performance} reports IoU regressor performance. The strong correlation (Pearson $r > 0.67$ across all settings) validates that attention alone encodes a meaningful localization signal---a proof-of-concept before we examine downstream localization gains.

\subsection{Self-Improved Localization Results}

Table~\ref{tab:main_results} presents results on COCO and Objects365. We discuss results in three parts.

\noindent\textbf{ACS-Learned: attention-based selection works.}
ACS-Learned consistently outperforms all baselines across configurations, confirming that the learned regressor captures a genuine localization-quality signal. On COCO, ACS-Learned improves Acc@0.5 from 61.4\% to 65.3\% for Qwen2.5-VL (\textcolor{teal}{+6.35\%}) and from 49.1\% to 58.6\% for InternVL-3.5 (\textcolor{teal}{+19.35\%}) over Greedy decoding. On the more challenging Objects365, ACS-Learned maintains stable improvements: \textcolor{teal}{+4.88\%} for Qwen2.5-VL and \textcolor{teal}{+16.44\%} for InternVL-3.5, demonstrating robustness across diverse object types. 

\noindent\textbf{ACS-Free: best training-free method.}
ACS-Free, which distills the regressor's gradient analysis into a parameter-free entropy rule, has the strongest overall performance among all training-free methods, producing the underlined best results in 7 out of 8 column comparisons. At $\tau{=}0.5$, ACS-Free improves Acc@0.5 over greedy by \textcolor{teal}{+3.26\%} for Qwen2.5-VL and \textcolor{teal}{+8.15\%} for InternVL-3.5 on COCO, demonstrating that entropy on regressor-identified heads captures most of the localization-quality signal without any learned component. The one exception---the weaker InternVL COCO Acc@0.5 at $\tau{=}0.5$, where MajVote slightly exceeds ACS-Free---
reflects that performance gain is affected by the base performance of a model.

\noindent\textbf{TokEntropy: a negative result.}
TokEntropy, which ranks candidates by the summed vocabulary entropy of bbox coordinate tokens, consistently underperforms even Pass@1 across most settings. This is a clear negative result: token-level confidence does not capture spatial grounding quality. The signal that matters for localization is not how uncertain the model is about which token to generate, but how its spatial attention is organized---precisely the signal ACS-Free exploits.

\noindent\textbf{Greedy vs. Sampling Trade-offs.}
While Greedy Decoding is simple, it often produces invalid or unusable outputs such as ``cannot see the target object'' or incomplete bounding boxes. This problem is particularly severe for InternVL, where greedy decoding fails to generate any valid bounding box for many cases,
leading to reduced localization accuracy.
Temperature sampling increases the chance of producing a valid box, which explains why FirstValid can outperform greedy decoding on InternVL, but it remains suboptimal at higher temperatures without a principled attention-based selector.

\subsection{Effect of Temperature and Sampling Size}

\begin{wraptable}{r}{0.4\linewidth}
\centering
\vspace{-\baselineskip}
\footnotesize
\caption{Performance saturation across different $N$ (Qwen2.5-VL, COCO, $\tau=1.0$).}
\label{tab:saturation}
\setlength{\tabcolsep}{4pt}
\begin{tabular}{l|cccccc}
\toprule
\textbf{N} & 3 & 5 & 10 & 15 & 20 & 30 \\
\midrule
\textbf{Acc@0.5} & 61.2 & 62.5 & \textbf{64.0} & 63.4 & 63.1 & 63.0 \\
\textbf{mIoU}    & 46.3 & 51.4 & \textbf{53.0} & 52.9 & 52.6 & 52.6 \\
\bottomrule
\end{tabular}
\vspace{0.5\baselineskip}
\end{wraptable}

Temperature and sampling size jointly control the diversity-quality tradeoff. Lower temperatures ($\tau=0.5$) yield more conservative and accurate boxes, while higher temperatures introduce more noisy candidates, causing FirstValid to degrade substantially (e.g., 60.9\% to 54.2\% Acc@0.5 on COCO for Qwen2.5-VL). As shown in Tables~\ref{tab:saturation} and \ref{tab:sampling_temperature_all}, increasing $N$ quickly improves ACS-Learned, but performance saturates once $N$>9; even $N$=3--5 gives substantial gains across temperatures. Thus, $N$=10 provides a strong balance between candidate diversity, accuracy, and computational cost.

\begin{table}[H]
\centering
\small
\setlength{\tabcolsep}{5pt}
\caption{Effect of sampling size N across temperatures on Qwen2.5-VL (COCO).}
\label{tab:sampling_temperature_all}
\begin{tabular}{c|cc|cc|cc}
\toprule
\multirow{2}{*}{\textbf{N}}
& \multicolumn{2}{c|}{$\tau = 1.0$}
& \multicolumn{2}{c|}{$\tau = 0.7$}
& \multicolumn{2}{c}{$\tau = 0.5$} \\
& Acc@0.5 & mIoU & Acc@0.5 & mIoU & Acc@0.5 & mIoU \\
\midrule
1  & 53.6 & 46.3 & 60.0 & 51.0 & 60.2 & 51.6 \\
3  & 61.2 & 51.4 & 64.3 & 54.1 & 63.7 & 54.2 \\
5  & 62.5 & 52.3 & \textbf{65.2} & 54.5 & 64.8 & 54.8 \\
10 & \textbf{64.0} & \textbf{53.0} & 64.9 & \textbf{54.7} & \textbf{65.3} & \textbf{55.2} \\
\bottomrule
\end{tabular}
\end{table}





\subsection{Multiple and Medium/Large Objects}
\begin{wraptable}{r}{0.44\linewidth}
\centering
\vspace{-\baselineskip}
\footnotesize
\setlength{\tabcolsep}{4pt}
\caption{ACS for Qwen2.5-VL on COCO multi-object.}
\label{tab:multi_object}
\begin{tabular}{lccc}
\toprule
\textbf{Setting} & \textbf{Prec. (\%)} & \textbf{Rec. (\%)} & \textbf{F1 (\%)} \\
\midrule
Greedy (w/o NMS)      & 47.21 & 40.19 & 43.42 \\
Greedy + NMS (0.5)    & \textbf{57.13} & 40.01 & 47.06 \\
Greedy + NMS (0.8)    & 56.29 & 40.10 & 46.83 \\
ACS ($\tau = 0.5$)    & 49.10 & \textbf{50.18} & 49.63 \\
ACS ($\tau = 0.7$)    & 50.93 & 50.13 & \textbf{50.53} \\
ACS ($\tau = 1.0$)    & 50.22 & 47.09 & 48.60 \\
\bottomrule
\end{tabular}
\vspace{-\baselineskip}
\end{wraptable}
ACS also works on images with multiple objects and medium/large objects settings.

\noindent\textbf{Multiple objects}.
Sampled bounding boxes are grouped per image via IoU-based Union--Find clustering, and the candidate with the highest predicted IoU from each cluster is selected (global best if no cluster forms). Table~\ref{tab:multi_object} shows Greedy decoding often produces many near-duplicate boxes that reduce recall; NMS suppresses redundancy and improves precision but cannot recover recall. In contrast, ACS-Learned uses temperature-based sampling to produce more diverse hypotheses, increasing recall at a modest cost to precision---a natural precision--recall trade-off. Our method achieves the highest F1 score, showing that candidate selection is an effective solution for multi-object localization.

\noindent\textbf{Medium/Large objects}.
We further validate on medium/large objects (occupying $>5\%$ of image area, roughly with width and height both $>22\%$) using an equal number of cases as the small-object setting. For Qwen2.5-VL, both Greedy and ACS-Learned reach 87\% Acc@0.5---the model is already near-optimal here, leaving little headroom for selection. For InternVL-3.5, ACS-Learned still improves Acc@0.5 from 64\% to 70\%, showing the method can help even on larger objects when the base model has weaker localization capability. Since LVLMs generally perform better on large than small objects, we focus on the more challenging small-object setting, where attention-based selection has the most impact.

\section{Conclusion}
We discovered that internal attention structure in LVLMs encodes object grounding quality, validated by an IoU regressor from attention maps alone. Building on this finding, our Attention-based Candidate Selection (ACS) framework improves small-object localization through two variants: ACS-Learned, which exploits the learned regressor signal, and ACS-Free, which distills it into a parameter-free entropy rule for interpretable and deployment-friendly use. Both variants achieve substantial gains with only 3--5 samples and minimal overhead. Our findings also advance understanding of how LVLMs process spatial reasoning.

\section{Limitations}
Our approach relies on internal attention maps and therefore requires white-box access to the LVLM. Closed-API models such as GPT-4V or Gemini do not expose attention activations, and so ACS is not directly applicable in those settings. This trade-off is inherent to attention-based methods and is what enables our mechanistic analysis.

As a sampling-based method, ACS incurs additional inference cost relative to greedy decoding. Performance saturates at 3--5 samples and the overhead is modest---attention is already computed during the forward pass and the regressor is lightweight---but latency-sensitive deployments may still prefer single-pass decoding.

The Attention-IoU Regressor in ACS-Learned is trained per LVLM and is not directly transferable across architectures with different attention shapes. Training is lightweight (3.6--4.4M parameters; minutes on a single GPU), and once a regressor is trained, the resulting head analysis enables ACS-Free, which requires no learned component at inference.

\section{Ethical Considerations}

\paragraph{Data and models.} All experiments use public benchmarks
(MS COCO~\cite{lin2014microsoft} and Objects365~\cite{shao2019objects365})
and open-source LVLMs (Qwen2.5-VL and InternVL-3.5) under their respective
licenses. No new data was collected and no personally identifiable
information is involved.

\paragraph{Intended use and misuse.} Our method improves small-object
localization in LVLMs and is intended for benign applications such as
assistive vision, robotics, and autonomous-driving safety. As with any
object-detection technique, the improved reliability could in principle
be repurposed for surveillance; the contribution is methodological and
is not tailored toward such use cases.

\paragraph{Use of LLMs.} Large language models were used for language
polishing (grammar and phrasing) and minor coding assistance (e.g.,
shell scripts, plotting). All research ideas, methodology, experimental
design, analyses, and claims are the authors' own.

\bibliography{main}

\newpage
\appendix
\clearpage
\section*{APPENDIX}
\addcontentsline{toc}{section}{Appendix}

\vspace{0.8em}

\begin{table}[h]
\centering
\small
\setlength{\tabcolsep}{6pt}
\renewcommand{\arraystretch}{1.25}
\caption*{\textbf{Appendix Table of Contents}}
\begin{tabular}{p{0.80\textwidth}@{\hfill}r}
\toprule
\textbf{Appendix Section} & \textbf{Page} \\
\midrule

\textbf{A. Training Setup for the Attention-IoU Regressor}
& \pageref{app:arm_settings} \\

\addlinespace[0.4em]

\textbf{B. Differential Entropy of a Gaussian Random Variable}
& \pageref{sec:differential-entropy} \\

\addlinespace[0.4em]

\textbf{C. Complete Attention Map Visualizations}
& \pageref{sec:complete_attention_maps} \\

\addlinespace[0.4em]

\textbf{D. Prompt for Object Localization}
& \pageref{sec:appendix-prompt} \\

\addlinespace[0.4em]

\textbf{E. Rank-based Entropy Selection}
& \pageref{sec:appendix-ranking} \\

\addlinespace[0.4em]

\textbf{F. Entropy Analysis in Localization-Critical Layers}
& \pageref{sec:appendix-critical-layers} \\

\addlinespace[0.4em]

\textbf{G. Additional Qualitative Results}
& \pageref{sec:appendix-qualitative} \\

\bottomrule
\end{tabular}
\end{table}

\section{Training Setup for the Attention-IoU Regressor}
\label{app:arm_settings}
We focus on Qwen2.5-VL and InternVL-3.5
because both LVLMs
possess strong object-localization capabilities and, critically, can directly
output bounding box coordinates in their responses.
The training data for the IoU regressor are constructed from 5,000 image-category pairs.
For each pair, each LVLM was queried with temperature $\tau = 1.0$ to produce 10 sampled responses, and all valid predicted bounding boxes in these responses are collected.
Each candidate box is paired with its four coordinate-specific attention maps based on $(x_{1}, y_{1}, x_{2}, y_{2})$, extracted from all layers and heads of the underlying LVLM. For a coordinate value such as \([15,\, 240,\, 35,\, 460]\), we extract the
attention maps by selecting the \emph{first token} of each coordinate.
Concretely, for example,
for a predicted box
$[15,\,240,\,35,\,460]$, we take the first token of
each coordinate:
the first coordinate uses the attention of
token $1$ in $15$ for $A_{x_1}$, the second coordinate uses the attention of token $2$ in $240$
for $A_{y_1}$, the third uses token $3$ in $35$ for $A_{x_2}$, and the fourth uses token
$4$ in $460$ for $A_{y_2}$. Thus, although the coordinate value is $240$, its associated
attention map comes from token $2$.

The IoU regressor takes these attention
maps
as input and predicts the IoU between the candidate box and the ground-truth box.

\subsection{Model configuration}
For Qwen2.5-VL, attention maps contain 28 layers with 28 heads per layer, while InternVL-3.5 provides 36 layers and 32 heads.
Each coordinate branch uses a CNN encoder with hidden dimensions $\{64,128,256\}$ and dropout rate $0.3$, followed by a fusion MLP of dimension $256$.
A sigmoid activation constrains the output to $[0,1]$.

\subsection{Optimization}
All IoU regressors are trained with Adam as the optimizer, a learning rate of 3e-4, a weight decay of 1e-4, and a batch size of \textbf{64}.
Training runs for up to \textbf{100 epochs}, using MSE loss and cosine learning-rate scheduling.
Early stopping is applied with patience $15$ and minimum improvement $0.001$.

\subsection{Attention extraction}
For Qwen2.5-VL, attention maps are computed over all image tokens produced by the vision encoder, and all heads are stacked along the channel dimension; the maps are then reshaped to a spatial size of $24\times24$.
For InternVL-3.5, we compute attention over the last 256 vision tokens, because these tokens correspond to the model's thumbnail representation; all heads are similarly stacked as channels, while the native $16\times16$ resolution is preserved.

\subsection{Model Architecture and Computational Efficiency}

The detailed architecture and computational cost of the regressor are summarized in Tables~\ref{tab:arm_layers_qwen}--\ref{tab:arm_memory_internvl}. Specifically, Tables~\ref{tab:arm_layers_qwen} and \ref{tab:arm_params_qwen} present the layer-wise structure and parameter statistics for the regressor trained on Qwen2.5-VL attention features, while Tables~\ref{tab:arm_layers_internvl} and \ref{tab:arm_memory_internvl} provide the corresponding details for InternVL-3.5.

Despite having 3.6M and 4.4M parameters of the regressors for Qwen2.5-VL and InternVL-3.5 respectively, regressor inference is highly efficient: processing 2,225 test cases takes only 5--6 minutes on a single A6000 GPU, which is negligible compared to the LVLM's own inference time. This efficiency makes the IoU regressor a practical and scalable solution for real-time bounding box selection in production environments.
\begin{table}[h]
\centering
\small
\caption{\textbf{Layer Summary of the Attention-IoU Regressor trained on Qwen2.5-VL attention maps.}}
\label{tab:arm_layers_qwen}
\begin{tabular}{lcc}
\toprule
\textbf{Layer (type:depth-idx)} & \textbf{Output Shape} & \textbf{Param \#} \\
\midrule

Attention-IoU Regressor & [1, 1] & -- \\

\midrule
\multicolumn{3}{l}{\textbf{CoordinateAttentionCNN Branch (repeated 4×):}} \\

Conv2d: 1-1 & [1, 64, 24, 24] & 451,648 \\
BatchNorm2d: 1-2 & [1, 64, 24, 24] & 128 \\
ReLU: 1-3 & [1, 64, 24, 24] & -- \\
MaxPool2d: 1-4 & [1, 64, 12, 12] & -- \\

Conv2d: 1-5 & [1, 128, 12, 12] & 73,856 \\
BatchNorm2d: 1-6 & [1, 128, 12, 12] & 256 \\
ReLU: 1-7 & [1, 128, 12, 12] & -- \\
MaxPool2d: 1-8 & [1, 128, 6, 6] & -- \\

Conv2d: 1-9 & [1, 256, 6, 6] & 295,168 \\
BatchNorm2d: 1-10 & [1, 256, 6, 6] & 512 \\
ReLU: 1-11 & [1, 256, 6, 6] & -- \\
AdaptiveAvgPool2d: 1-12 & [1, 256, 1, 1] & -- \\
Linear: 1-13 & [1, 128] & 32,896 \\

\midrule
\multicolumn{3}{l}{\textbf{Fusion MLP:}} \\

Linear: 2-1 & [1, 256] & 131,328 \\
BatchNorm1d: 2-2 & [1, 256] & 512 \\
ReLU: 2-3 & [1, 256] & -- \\

Linear: 2-4 & [1, 128] & 32,896 \\
ReLU: 2-5 & [1, 128] & -- \\

Linear: 2-6 & [1, 1] & 129 \\

\bottomrule
\end{tabular}
\end{table}

\begin{table}[h]
\centering
\small
\caption{\textbf{Parameters and Memory Usage of the Attention-IoU Regressor trained on Qwen2.5-VL attention maps.}}
\label{tab:arm_params_qwen}
\begin{tabular}{lc}
\toprule
\textbf{Description} & \textbf{Value} \\
\midrule
Total parameters & 3,582,721 \\
Trainable parameters & 3,582,721 \\
Non-trainable parameters & 0 \\
Total mult-adds (G) & 1.13 \\
\midrule
Input size (MB) & 7.23 \\
Forward/backward pass size (MB) & 4.14 \\
Params size (MB) & 14.33 \\
Estimated total size (MB) & 25.69 \\
\bottomrule
\end{tabular}
\end{table}

\begin{table}[h]
\centering
\small
\caption{\textbf{Layer Summary of the Attention-IoU Regressor trained on InternVL-3.5 attention maps.}}
\label{tab:arm_layers_internvl}
\begin{tabular}{lcc}
\toprule
\textbf{Layer (type:depth-idx)} & \textbf{Output Shape} & \textbf{Param \#} \\
\midrule

Attention-IoU Regressor & [1, 1] & -- \\

\midrule
\multicolumn{3}{l}{\textbf{CoordinateAttentionCNN Branch (4× identical branches):}} \\

Conv2d: 1-1 & [1, 64, 16, 16] & 663,616 \\
BatchNorm2d: 1-2 & [1, 64, 16, 16] & 128 \\
ReLU: 1-3 & [1, 64, 16, 16] & -- \\
Dropout2d: 1-4 & [1, 64, 16, 16] & -- \\
MaxPool2d: 1-5 & [1, 64, 8, 8] & -- \\

Conv2d: 1-6 & [1, 128, 8, 8] & 73,856 \\
BatchNorm2d: 1-7 & [1, 128, 8, 8] & 256 \\
ReLU: 1-8 & [1, 128, 8, 8] & -- \\
Dropout2d: 1-9 & [1, 128, 8, 8] & -- \\
MaxPool2d: 1-10 & [1, 128, 4, 4] & -- \\

Conv2d: 1-11 & [1, 256, 4, 4] & 295,168 \\
BatchNorm2d: 1-12 & [1, 256, 4, 4] & 512 \\
ReLU: 1-13 & [1, 256, 4, 4] & -- \\
AdaptiveAvgPool2d: 1-14 & [1, 256, 1, 1] & -- \\

Flatten: 1-15 & [1, 256] & -- \\
Dropout: 1-16 & [1, 256] & -- \\
Linear: 1-17 & [1, 128] & 32,896 \\

\midrule
\multicolumn{3}{l}{\textbf{Fusion MLP:}} \\

Linear: 2-1 & [1, 256] & 131,328 \\
ReLU: 2-2 & [1, 256] & -- \\
BatchNorm1d: 2-3 & [1, 256] & 512 \\
Dropout: 2-4 & [1, 256] & -- \\

Linear: 2-5 & [1, 128] & 32,896 \\
ReLU: 2-6 & [1, 128] & -- \\
Dropout: 2-7 & [1, 128] & -- \\

Linear: 2-8 & [1, 1] & 129 \\

\bottomrule
\end{tabular}
\end{table}

\begin{table}[h]
\centering
\small
\caption{\textbf{Parameters and Memory Usage of the Attention-IoU Regressor trained on InternVL-3.5 attention maps.}}
\label{tab:arm_memory_internvl}
\begin{tabular}{lc}
\toprule
\textbf{Description} & \textbf{Value} \\
\midrule
Total parameters & 4,430,593 \\
Trainable parameters & 4,430,593 \\
Non-trainable parameters & 0 \\
Total mult-adds (M) & 717.64 \\
\midrule
Input size (MB) & 4.72 \\
Forward/backward pass size (MB) & 1.84 \\
Params size (MB) & 17.72 \\
Estimated total size (MB) & 24.29 \\
\bottomrule
\end{tabular}
\end{table}


\section{Differential Entropy of a Gaussian Random Variable}
\label{sec:differential-entropy}
This section supplements the result used in Section 3.2 of the main paper by providing the full derivation of the conditional Gaussian entropy.

We derive the standard result used in the main text:
\[
H\!\left(Y \mid f_\theta(\mathbf{A})\right) = \frac{1}{2}\log(2\pi e \sigma^2),
\]
under the standard regression noise model where
\[
Y \mid f_\theta(\mathbf{A}) = z \sim \mathcal{N}(z,\sigma^2).
\]
That is, the conditional distribution of $Y$ given the prediction $z = f_\theta(\mathbf{A})$ is Gaussian with mean $z$ and variance $\sigma^2$.

\paragraph{Derivation.}
By definition, the conditional differential entropy is
\[
H(Y \mid Z) = - \int p(y \mid z)\,\log p(y \mid z)\,dy,
\]
where \(Z = f_\theta(\mathbf{A})\).
If \(Y\mid Z=z \sim \mathcal{N}(z,\sigma^2)\), then
\[
\begin{aligned}
p(y\mid z)
&= \frac{1}{\sqrt{2\pi\sigma^2}}
   \exp\!\left(-\frac{(y-z)^2}{2\sigma^2}\right), \\[3pt]
\log p(y\mid z)
&= -\tfrac12 \log(2\pi\sigma^2)
   - \frac{(y-z)^2}{2\sigma^2}.
\end{aligned}
\]

Substituting into the entropy integral:

\[
\begin{aligned}
H(Y\mid Z=z)
&= -\int p(y\mid z)\,\log p(y\mid z)\,dy \\[3pt]
&= \int p(y\mid z)\Big[
    \tfrac12\log(2\pi\sigma^2)
    + \tfrac{(y-z)^2}{2\sigma^2}
\Big]dy \\[3pt]
&= \tfrac12\log(2\pi\sigma^2)
   + \tfrac{1}{2\sigma^{2}}
     \int p(y\mid z)(y-z)^2 dy.
\end{aligned}
\]

Because \(\int p(y\mid z)dy = 1\) and
\(\int p(y\mid z)(y-z)^2 dy = \sigma^2\),
\[
H(Y\mid Z=z)
= \tfrac{1}{2}\log(2\pi\sigma^2)+\tfrac12
= \tfrac12\log(2\pi e \sigma^2).
\]
Since the expression does not depend on \(z\), the conditional entropy is
\[
H(Y\mid Z)=\tfrac12\log(2\pi e \sigma^2).
\]

\section{Complete Attention Map Visualizations}
\label{sec:complete_attention_maps}
Below we provide the complete attention map visualizations of Qwen2.5-VL and InternVL-3.5 corresponding to Figure~3 in the main paper.
These figures show all layers and coordinate-specific attention maps used in our analysis.
\begin{figure*}[t]
    \centering
    \begin{subfigure}{\textwidth}
        \centering
        \includegraphics[width=\textwidth]{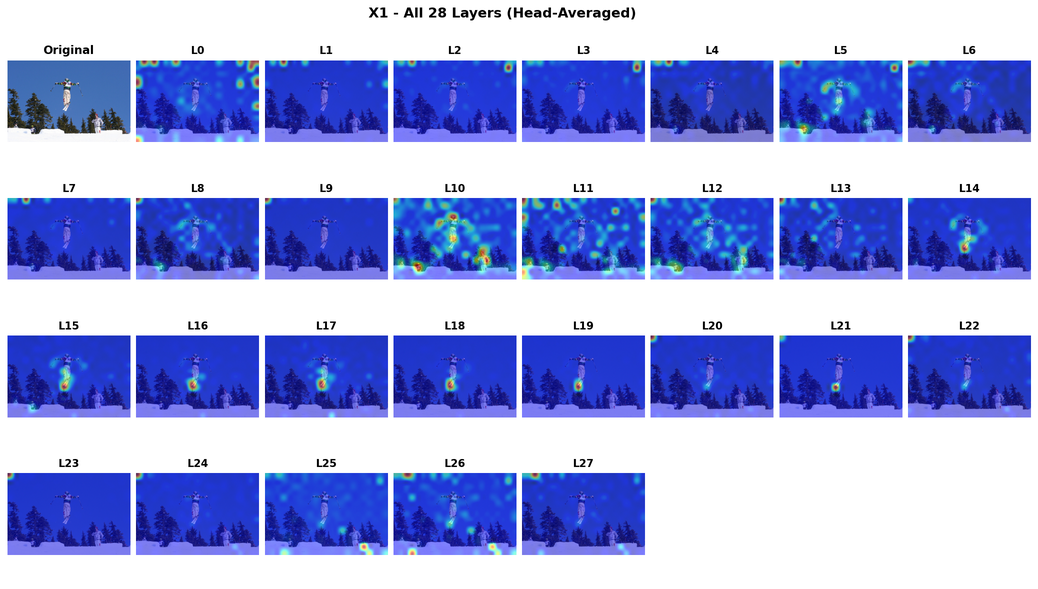}
        \caption{Attention maps across all layers for x1 coordinate prediction.}
        \label{fig:qwen_attention_x1}
    \end{subfigure}

    \vspace{0.5cm} 

    \begin{subfigure}{\textwidth}
        \centering
        \includegraphics[width=\textwidth]{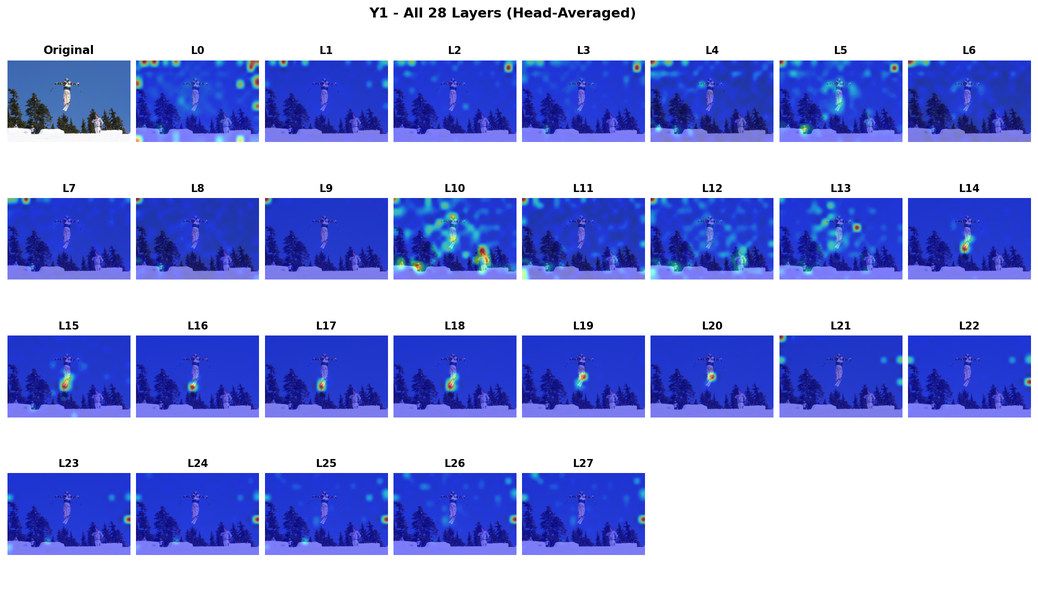}
        \caption{Attention maps across all layers for y1 coordinate prediction.}
        \label{fig:attention_y1}
    \end{subfigure}

    \caption{
    \textbf{Qwen2.5-VL attention visualization for object localization (Part 1).}
    Early layers exhibit diffuse and globally distributed attention, while
    mid-to-late layers progressively focus on the target object for both
    (a) $x_1$ and (b) $y_1$ coordinate prediction. These coordinate-specific
    attention maps serve as the input to the regressor for estimating the IoU of each
    candidate bounding box. For each layer, the visualization represents the
    attention map averaged across all heads. This example is randomly selected, and other samples exhibit similar patterns.
    }

    \label{fig:attention_maps_qwen}
\end{figure*}

\begin{figure*}[t]
    \centering

    \begin{subfigure}{\textwidth}
        \centering
        \includegraphics[width=\textwidth]{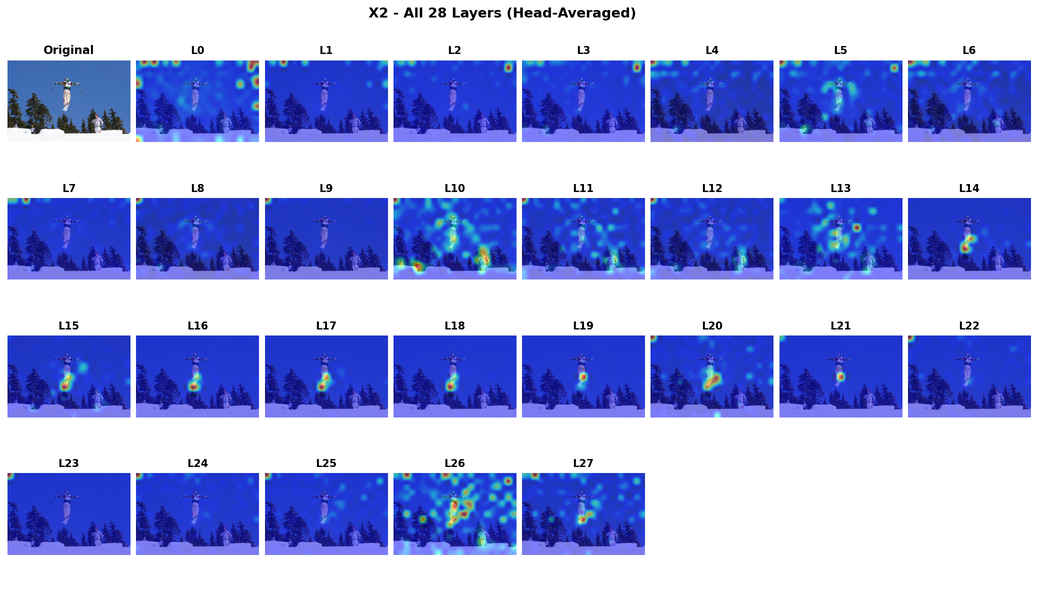}
        \caption{x2 coordinate prediction.}
        \label{fig:qwen_attention_x2}
    \end{subfigure}

    \vspace{0.5cm}

    \begin{subfigure}{\textwidth}
        \centering
        \includegraphics[width=\textwidth]{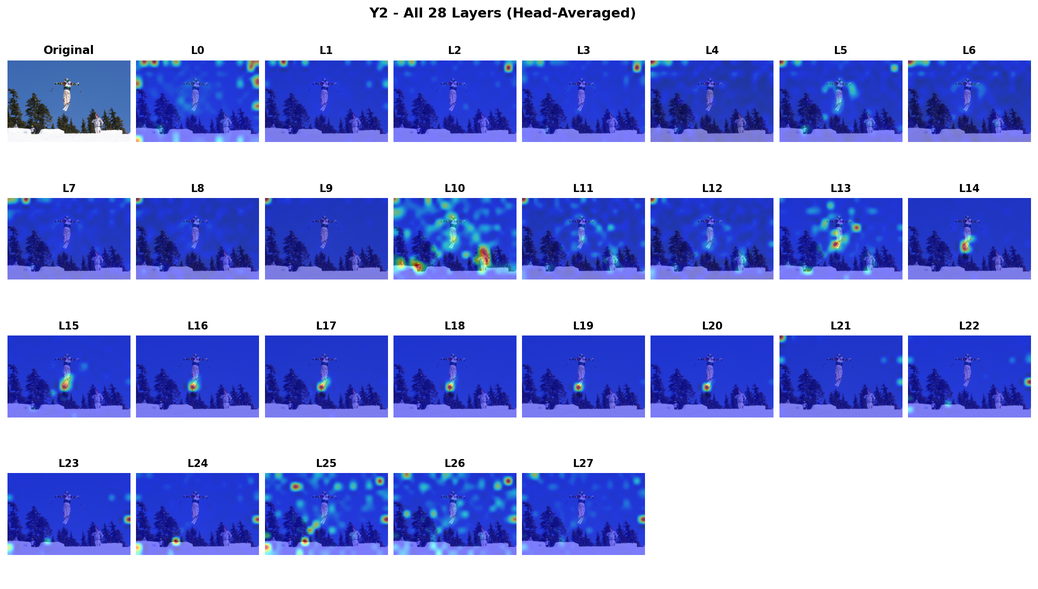}
        \caption{y2 coordinate prediction.}
        \label{fig:qwen_attention_y2}
    \end{subfigure}

    \caption{
    \textbf{Qwen2.5-VL attention visualization for object localization (Part 2).}
    Early layers exhibit diffuse and globally distributed attention, while
    mid-to-late layers progressively focus on the target object for both
    (a) $x_2$ and (b) $y_2$ coordinate prediction. These coordinate-specific
    attention maps serve as the input to the regressor for estimating the IoU of each
    candidate bounding box. For each layer, the visualization represents the
    attention map averaged across all heads. This example is randomly selected, and other samples exhibit similar patterns.
    }
    \label{fig:attention_maps_qwen_part2}
\end{figure*}

\begin{figure*}[t]
    \centering

    \begin{subfigure}{\textwidth}
        \centering
        \includegraphics[width=\textwidth]{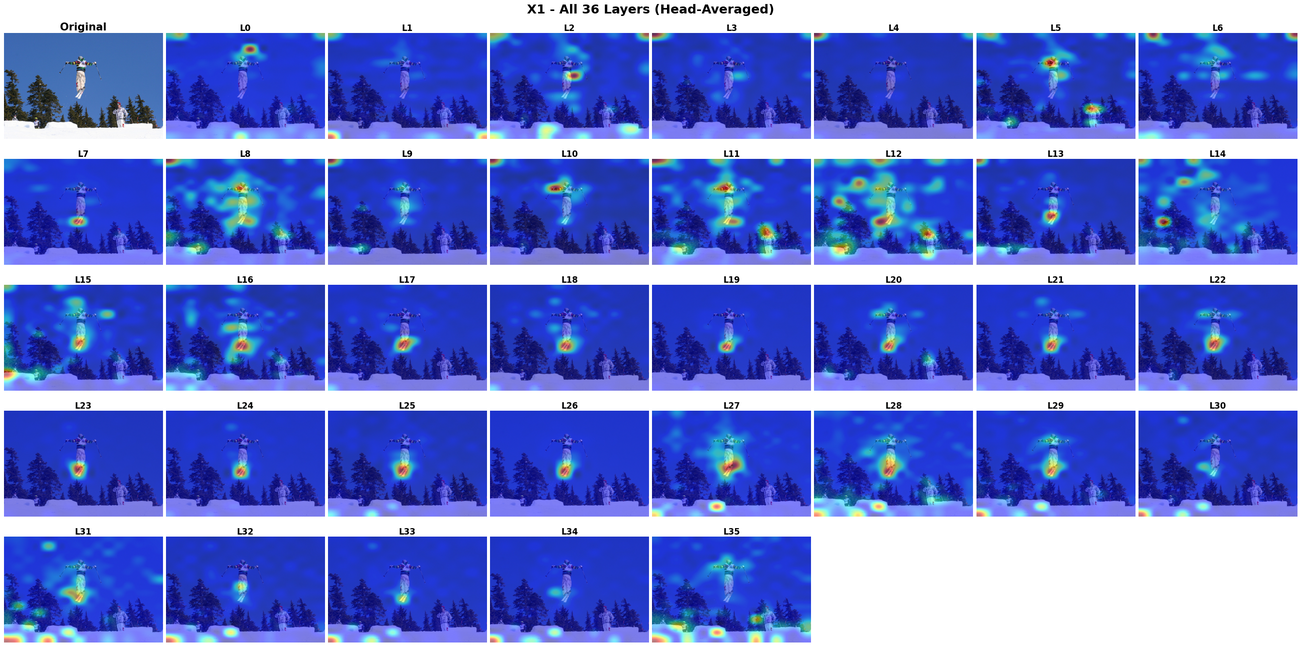}
        \caption{x1 coordinate prediction.}
        \label{fig:attention_x1_internvl}
    \end{subfigure}

    \vspace{0.5cm}

    \begin{subfigure}{\textwidth}
        \centering
        \includegraphics[width=\textwidth]{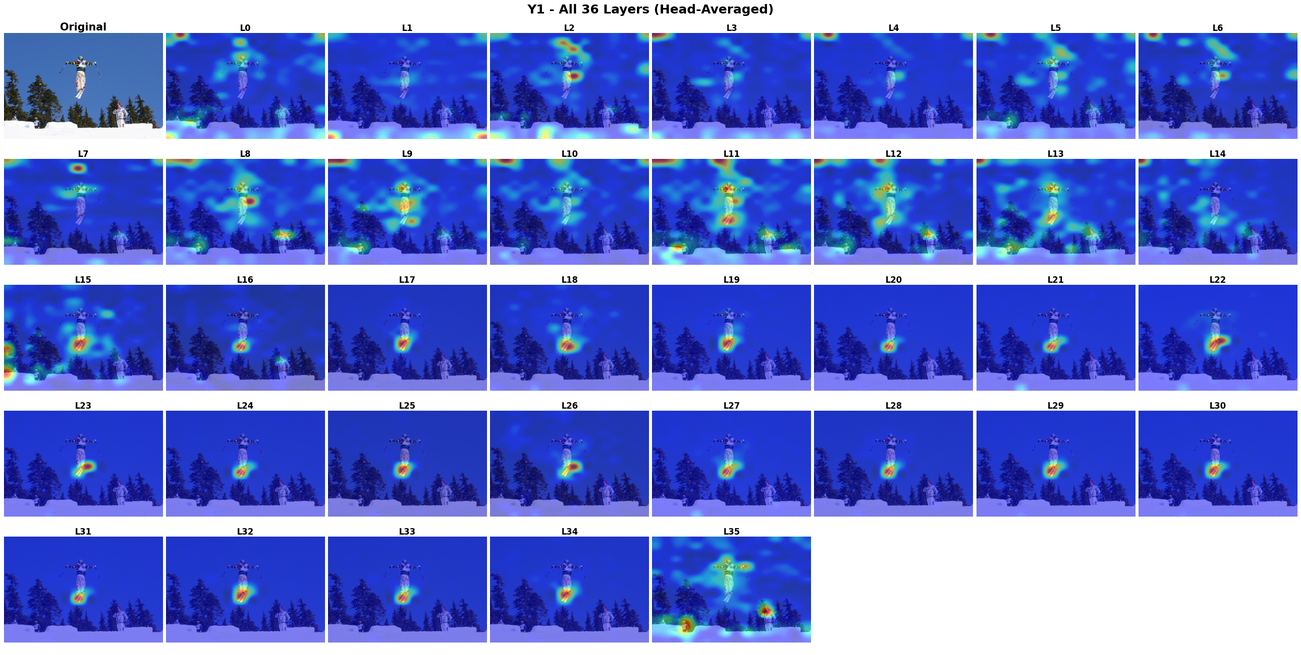}
        \caption{y1 coordinate prediction.}
        \label{fig:attention_y1_internvl}
    \end{subfigure}

    \caption{
    \textbf{InternVL-3.5 attention visualization for object localization (Part 1).}
    Early layers show diffuse and broadly distributed attention, while
    mid-to-late layers gradually focus on the target object for both
    (a) $x_1$ and (b) $y_1$ coordinate prediction. These coordinate-specific
    attention maps serve as the input to the regressor for estimating the IoU of each
    candidate bounding box. For each layer, the visualization represents the
    attention map averaged across all heads. This example is randomly selected, and other samples exhibit similar patterns.
    }
    \label{fig:attention_maps_internvl_part1}
\end{figure*}

\begin{figure*}[t]
    \centering

    \begin{subfigure}{\textwidth}
        \centering
        \includegraphics[width=\textwidth]{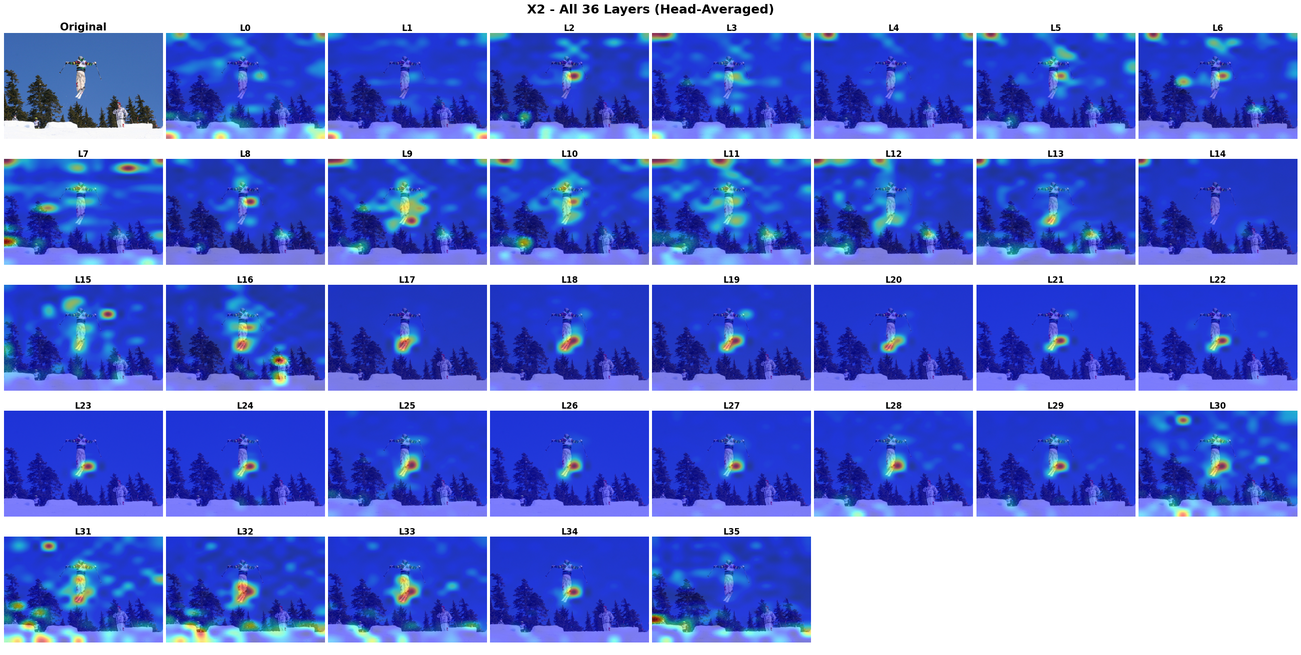}
        \caption{x2 coordinate prediction.}
        \label{fig:attention_x2_internvl}
    \end{subfigure}

    \vspace{0.5cm}

    \begin{subfigure}{\textwidth}
        \centering
        \includegraphics[width=\textwidth]{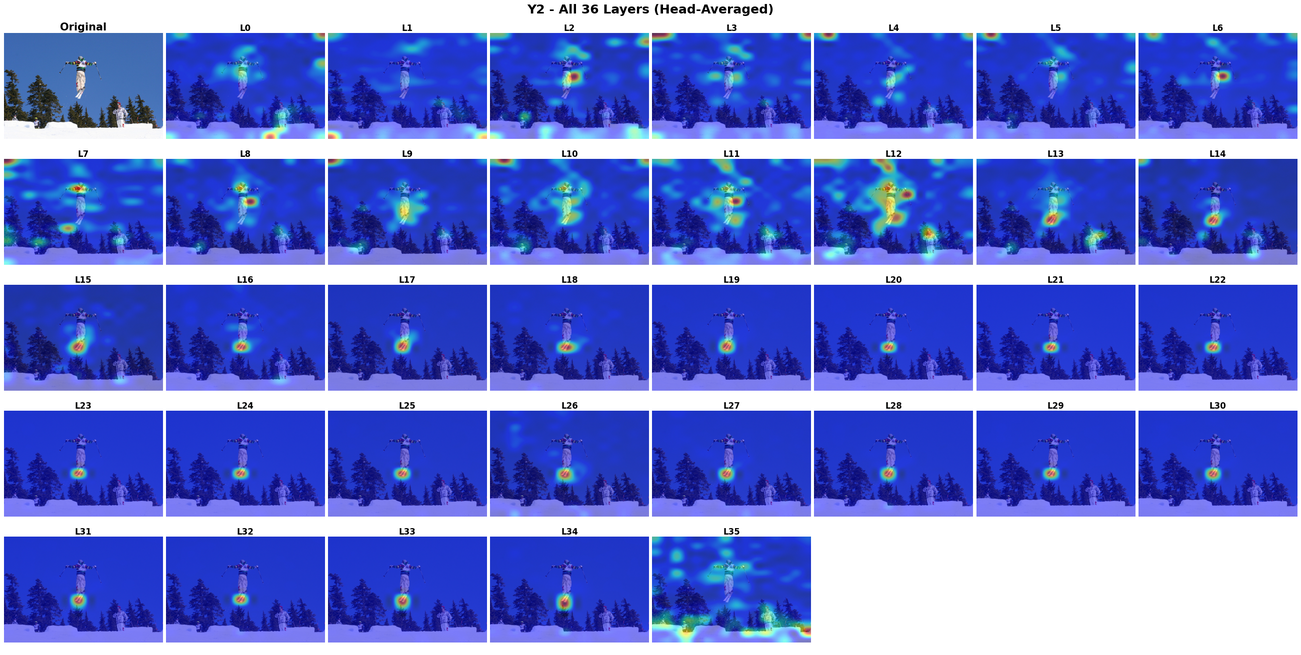}
        \caption{y2 coordinate prediction.}
        \label{fig:attention_y2_internvl}
    \end{subfigure}

    \caption{
    \textbf{InternVL-3.5 attention visualization for object localization (Part 2).}
    Early layers show diffuse and broadly distributed attention, while
    mid-to-late layers gradually focus on the target object for both
    (a) $x_2$ and (b) $y_2$ coordinate prediction. These coordinate-specific
    attention maps serve as the input to the regressor for estimating the IoU of each
    candidate bounding box. For each layer, the visualization represents the
    attention map averaged across all heads. This example is randomly selected, and other samples exhibit similar patterns.
    }
    \label{fig:attention_maps_internvl_part2}
\end{figure*}

\FloatBarrier

\onecolumn

\section{Prompt for Object Localization}
\label{sec:appendix-prompt}
To ensure a fair comparison across models, both Qwen2.5-VL and InternVL-3.5 are
prompted using the same instruction shown below. With this prompt, the LVLM
produces responses that follow the example format and include bounding boxes in
a consistent, structured output.

\begin{figure*}[h]
\centering
\begin{tcolorbox}[title={Prompt for Localization},colback=gray!5,
colframe=gray!50!black,width=0.92\textwidth]
\small
\begin{verbatim}
Detect all instances of {object_name} in the image:

Your task is to detect any {object_name} objects that are visible in the image.

For each detected object, provide:
The bounding box coordinates [x1, y1, x2, y2]

Output format (as a JSON array):
[
  {"bbox_2d": [x1, y1, x2, y2], "label": "{object_name}"},
  {"bbox_2d": [x1, y1, x2, y2], "label": "{object_name}"}
]

Important: Focus on any {object_name} that might be present in the image,
especially those that are still visible and NOT masked out.

Example output:
[
  {"bbox_2d": [200, 150, 280, 220], "label": "{object_name}"},
  {"bbox_2d": [350, 180, 420, 270], "label": "{object_name}"}
]
\end{verbatim}
\end{tcolorbox}
\end{figure*}

\section{Rank-based Entropy Selection}
\label{sec:appendix-ranking}

In the main paper, we describe ACS-Free using rank-based aggregation to select bounding boxes. Here we clarify the ranking procedure. Given a candidate set $\mathcal{B} = \{b_1, b_2, \ldots, b_T\}$ of $T$ bounding boxes for an image, we first compute the average entropy $\bar{H}_c(b_j)$ for each coordinate $c \in \{x_1, y_1, x_2, y_2\}$ across the discriminative attention heads $\mathcal{D}_c$. Then, for each coordinate independently, we rank all candidates based on their entropy values in ascending order (lower entropy receives better rank), yielding coordinate-wise ranks $r_c(b_j) \in \{1, 2, \ldots, T\}$. The total rank sum $R(b_j) = r_{x_1}(b_j) + r_{y_1}(b_j) + r_{x_2}(b_j) + r_{y_2}(b_j)$ aggregates the quality across all four coordinates. Finally, we select the bounding box with the minimum rank sum: $b^* = \arg\min_{b_j \in \mathcal{B}} R(b_j)$. This rank-based approach is more robust than directly using mean entropy values, as it is invariant to entropy scale differences across coordinates and ensures balanced contribution from all four coordinate predictions.

\newpage

\section{Entropy Analysis in Localization-Critical Layers}
\label{sec:appendix-critical-layers}
\begin{figure}[h!]
    \centering

    \begin{subfigure}[b]{0.45\textwidth}
        \centering
        \includegraphics[width=\linewidth]{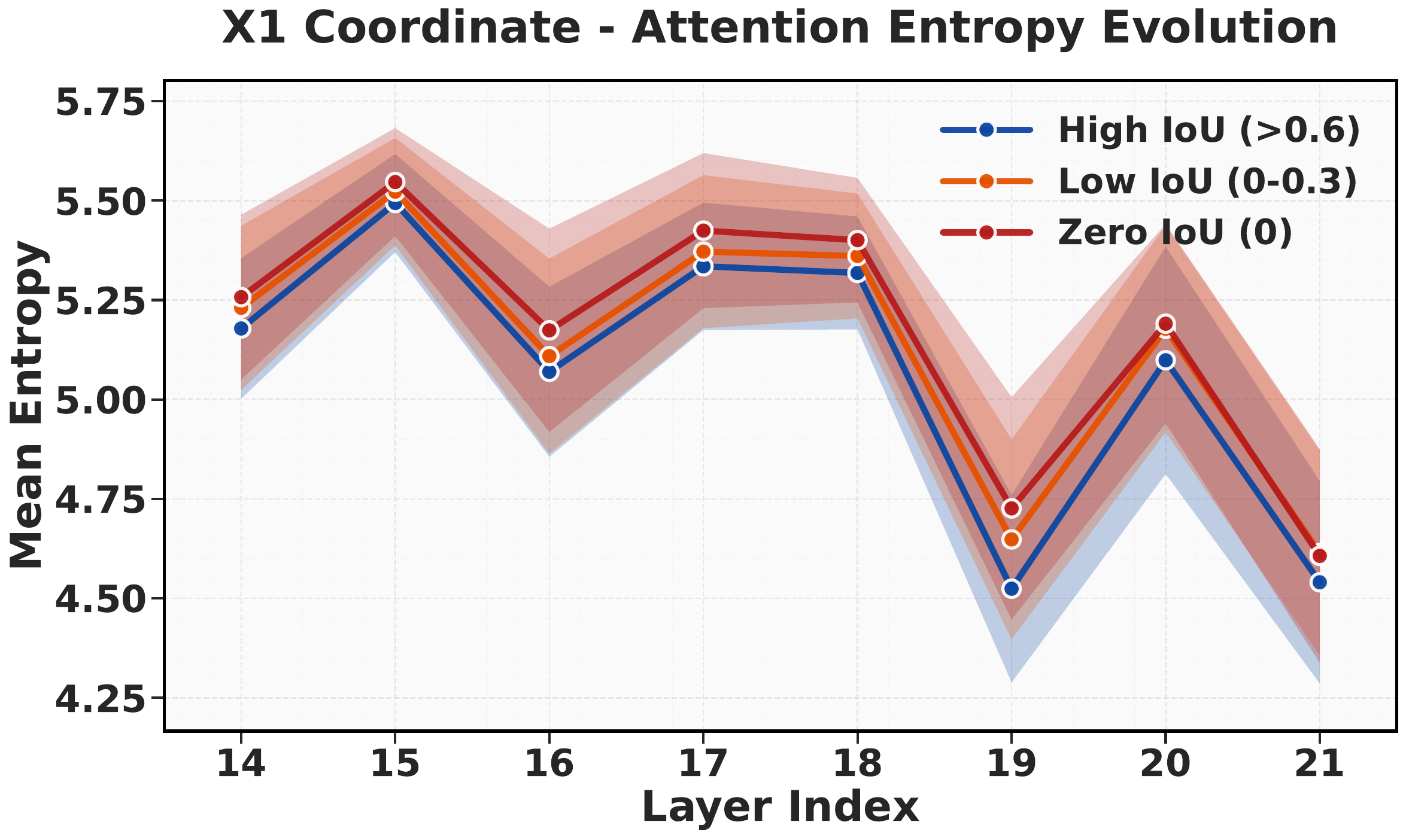}
        \caption*{x1}
    \end{subfigure}
    \hfill
    \begin{subfigure}[b]{0.45\textwidth}
        \centering
        \includegraphics[width=\linewidth]{figs/middle_layer_attention_analysis/y1_entropy_evolution_paper.pdf}
        \caption*{y1}
    \end{subfigure}

    \vspace{0.6em}

    \begin{subfigure}[b]{0.45\textwidth}
        \centering
        \includegraphics[width=\linewidth]{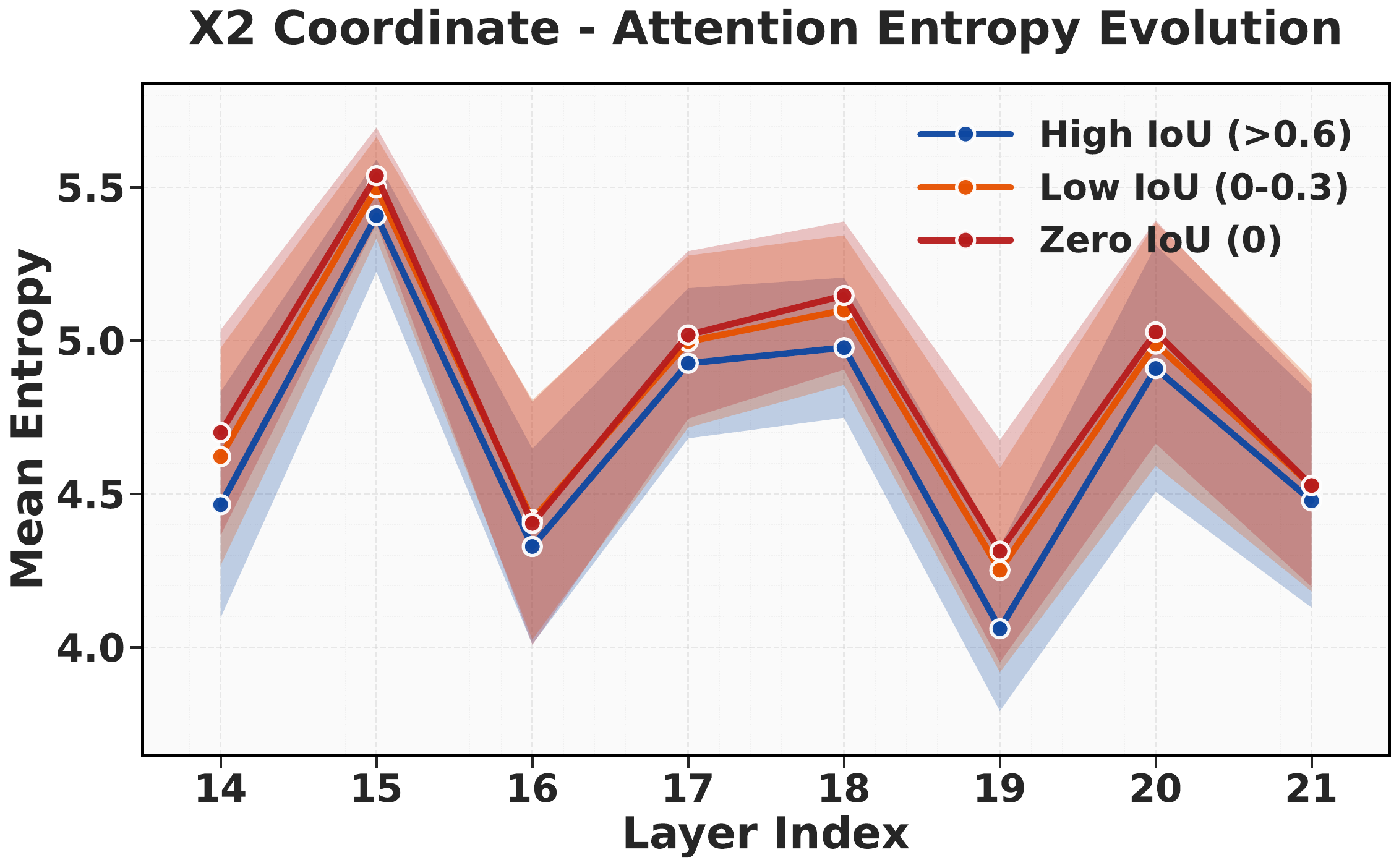}
        \caption*{x2}
    \end{subfigure}
    \hfill
    \begin{subfigure}[b]{0.45\textwidth}
        \centering
        \includegraphics[width=\linewidth]{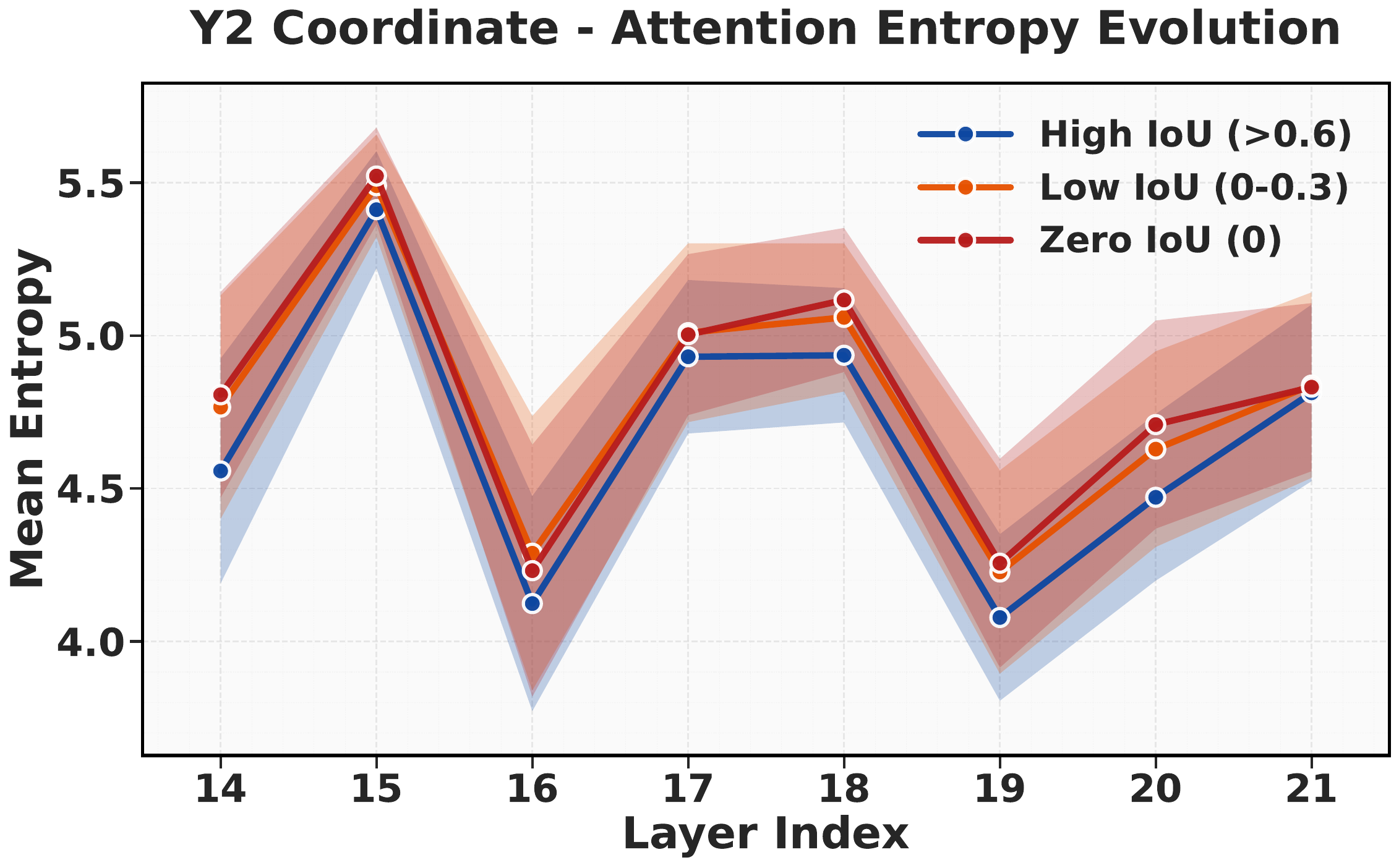}
        \caption*{y2}
    \end{subfigure}

    \caption{Important layers analysis across the attention maps of four bounding box coordinates.}
    \label{fig:full_important_layer_analysis}
\end{figure}

Figure~\ref{fig:full_important_layer_analysis} presents the full version of Figure 4 in the main text, providing a comprehensive visualization of our localization-critical layer analysis, including the entropy patterns of attention maps for all four coordinate tokens. The figure shows that in the layers identified by the IoU regressor, high-IoU predictions exhibit clearly lower entropy than low- or zero-IoU cases, indicating stronger and more focused attention. This consistent separation confirms that these layers contain the dominant localization signal. The visualization also highlights which layers provide the most discriminative patterns, motivating the design of \textbf{ACS-Free}, the training-free variant of our framework that uses entropy on these discriminative heads for bounding box selection.
\paragraph{Implementation detail of ACS-Free.}
For ACS-Free, we construct $\mathcal{D}_c$ as the index set of the top-3 most discriminative heads for the attention map of coordinate $c$; during inference, the attention maps from these heads are used to compute the coordinate-wise entropy $\bar{H}_c(b_j)$ for ranking candidate boxes.

\section{Additional Qualitative Results}
\label{sec:appendix-qualitative}
\begin{figure*}[t]
\centering
\begin{tabular}{ccc}
Traffic Light & Person & Vase\\
\includegraphics[height=1.5in]{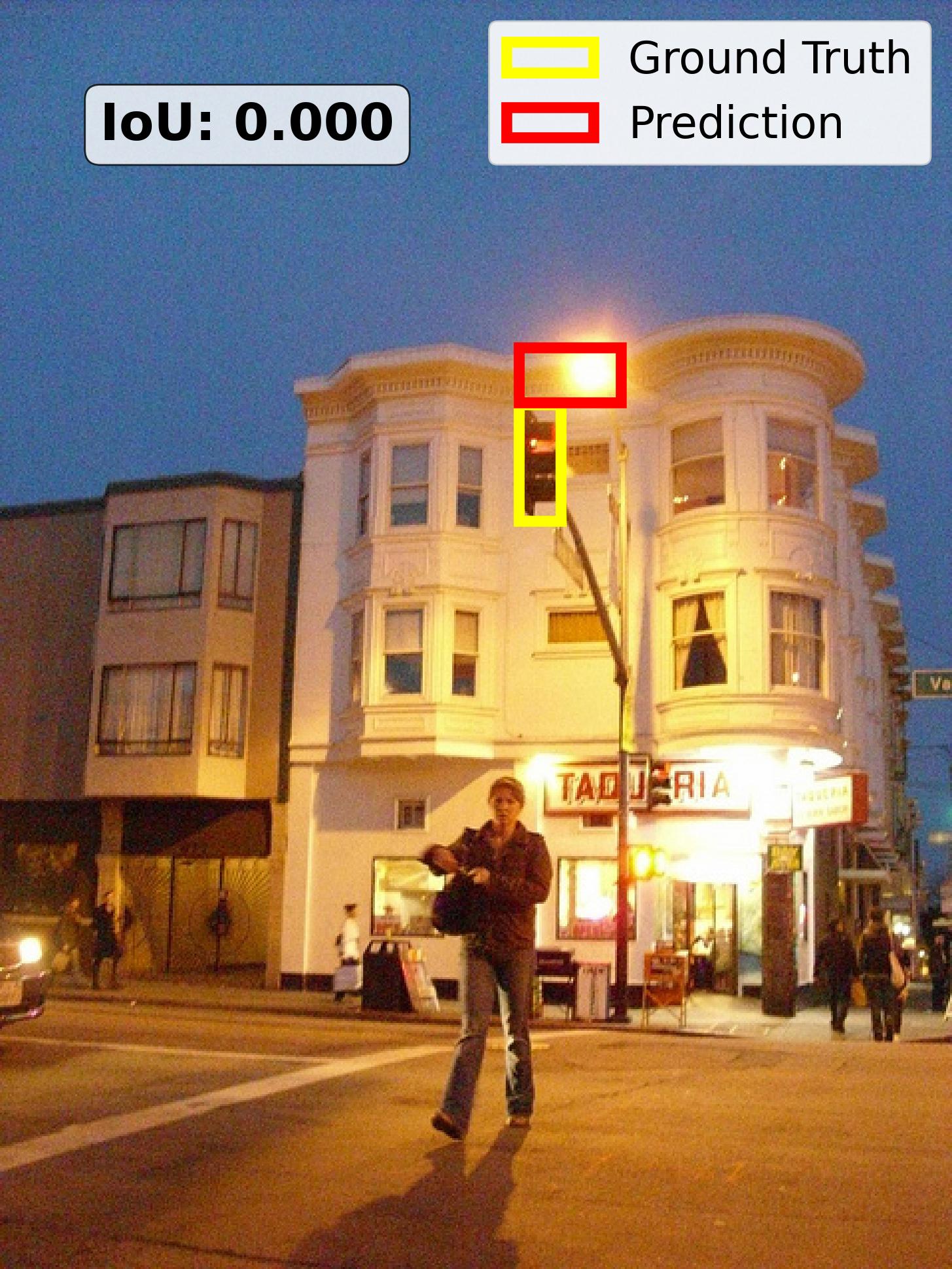} &
\includegraphics[height=1.5in]{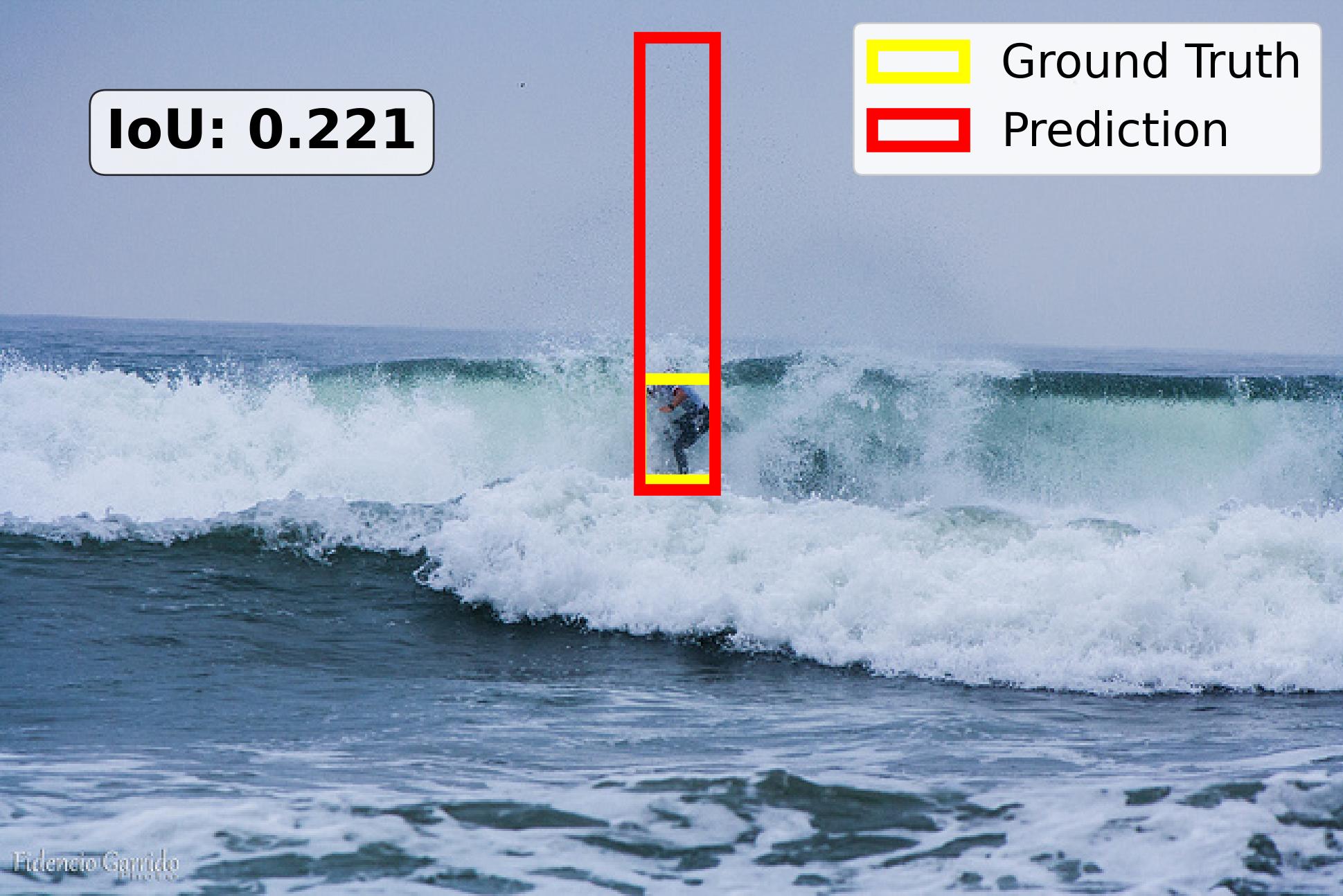} &
\includegraphics[height=1.5in]{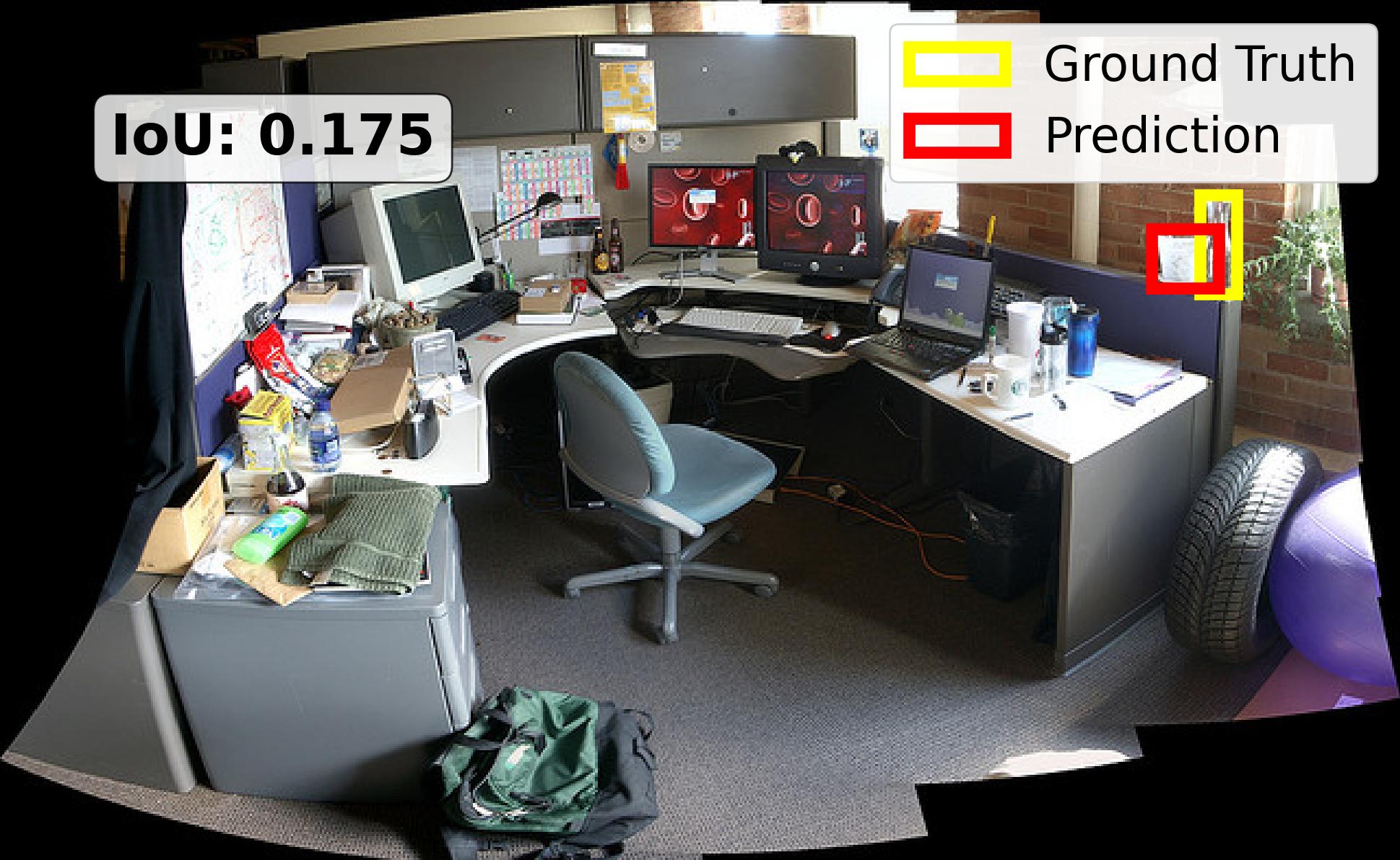}\\
\includegraphics[height=1.5in]{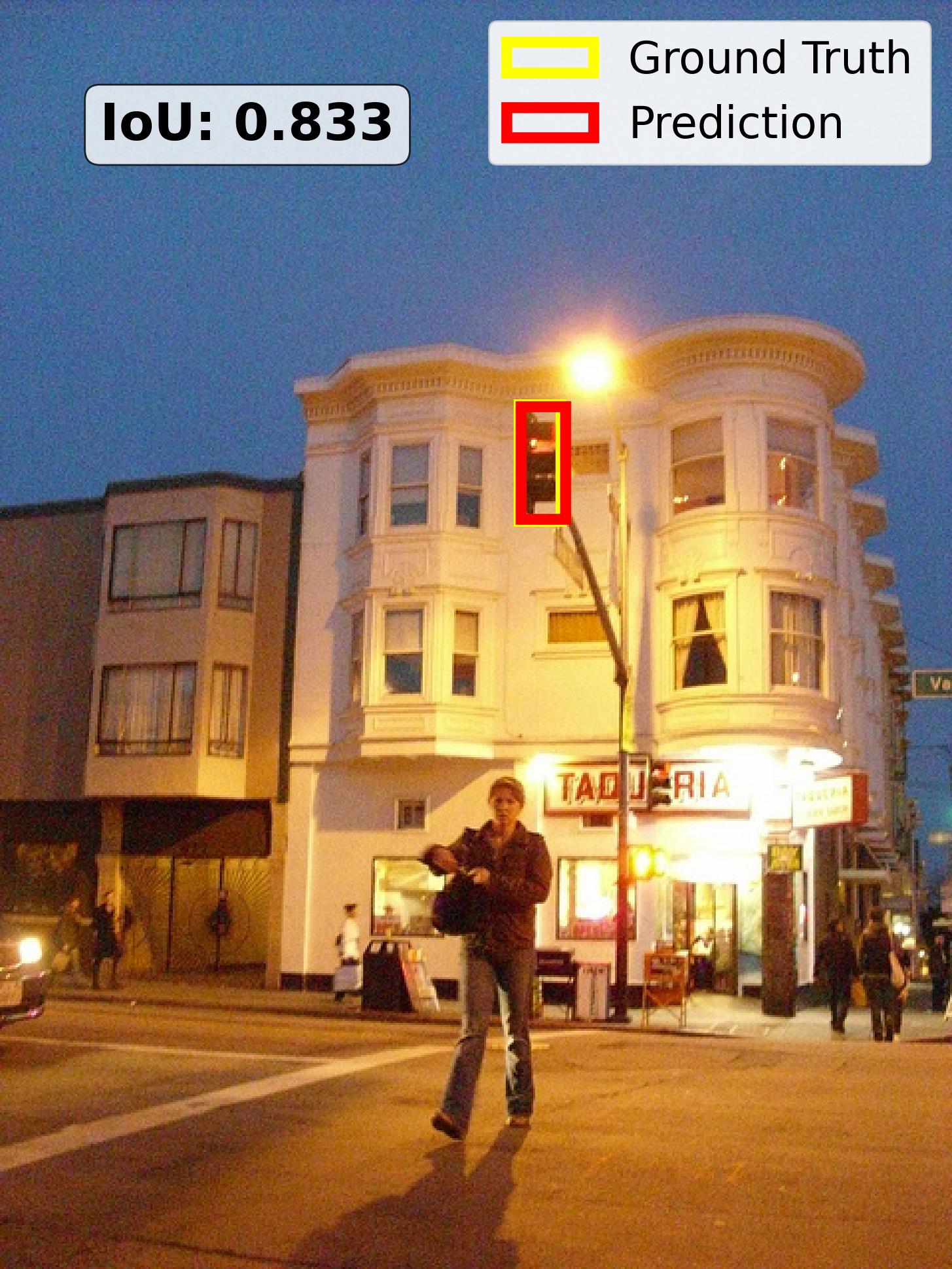} &
\includegraphics[height=1.5in]{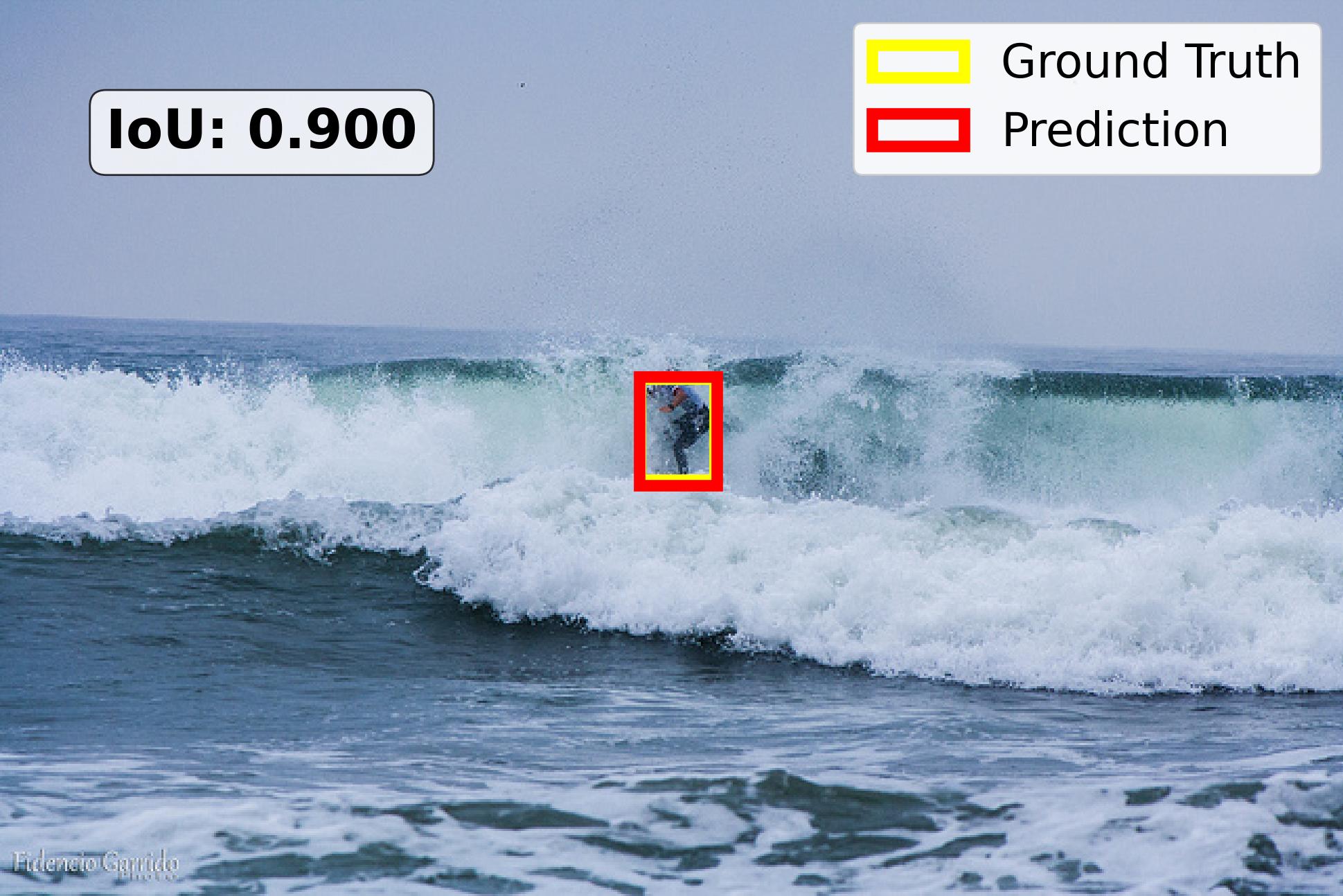} &
\includegraphics[height=1.5in]{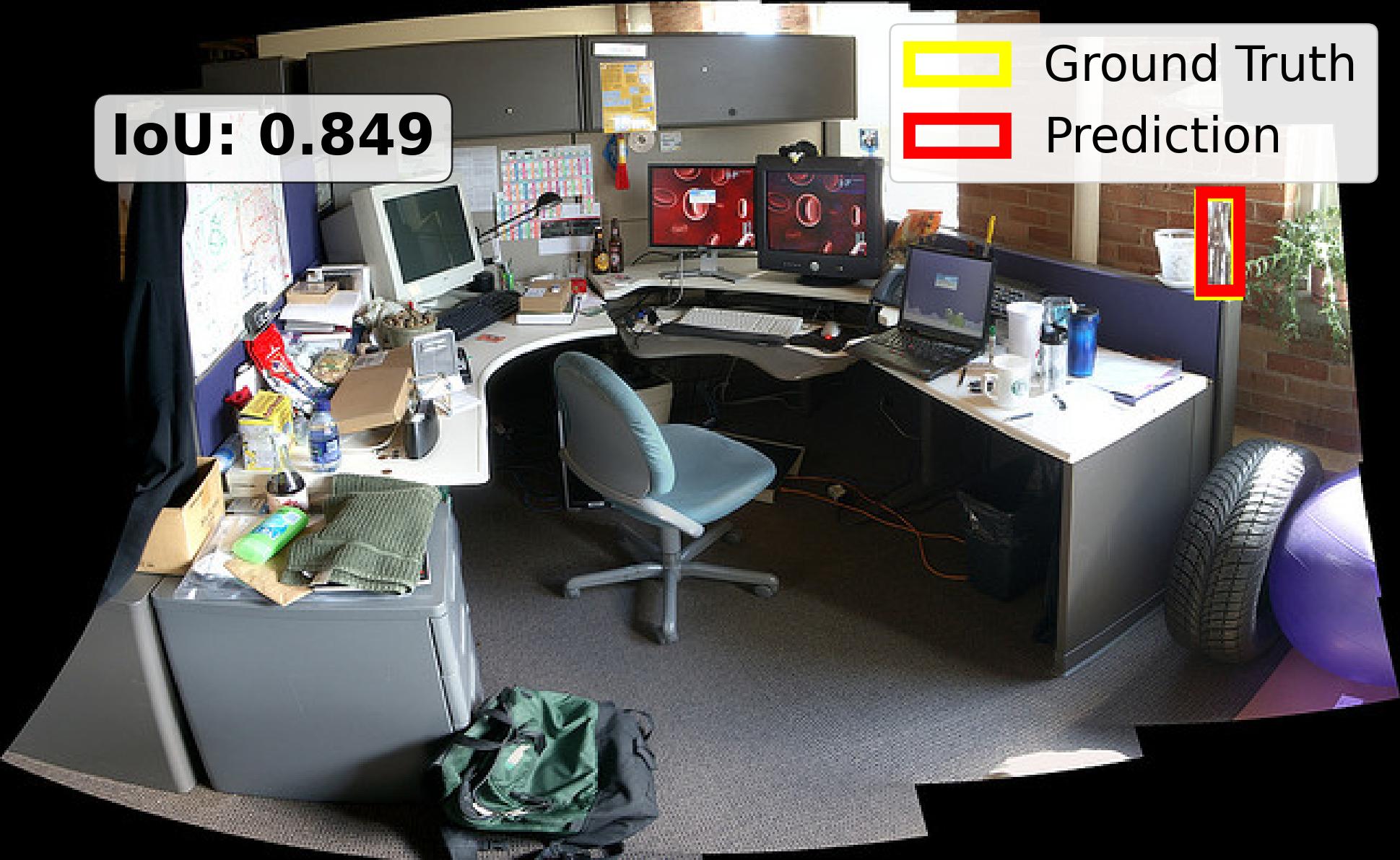}\\
Microwave & Cup & Backpack\\
\includegraphics[height=1.5in]{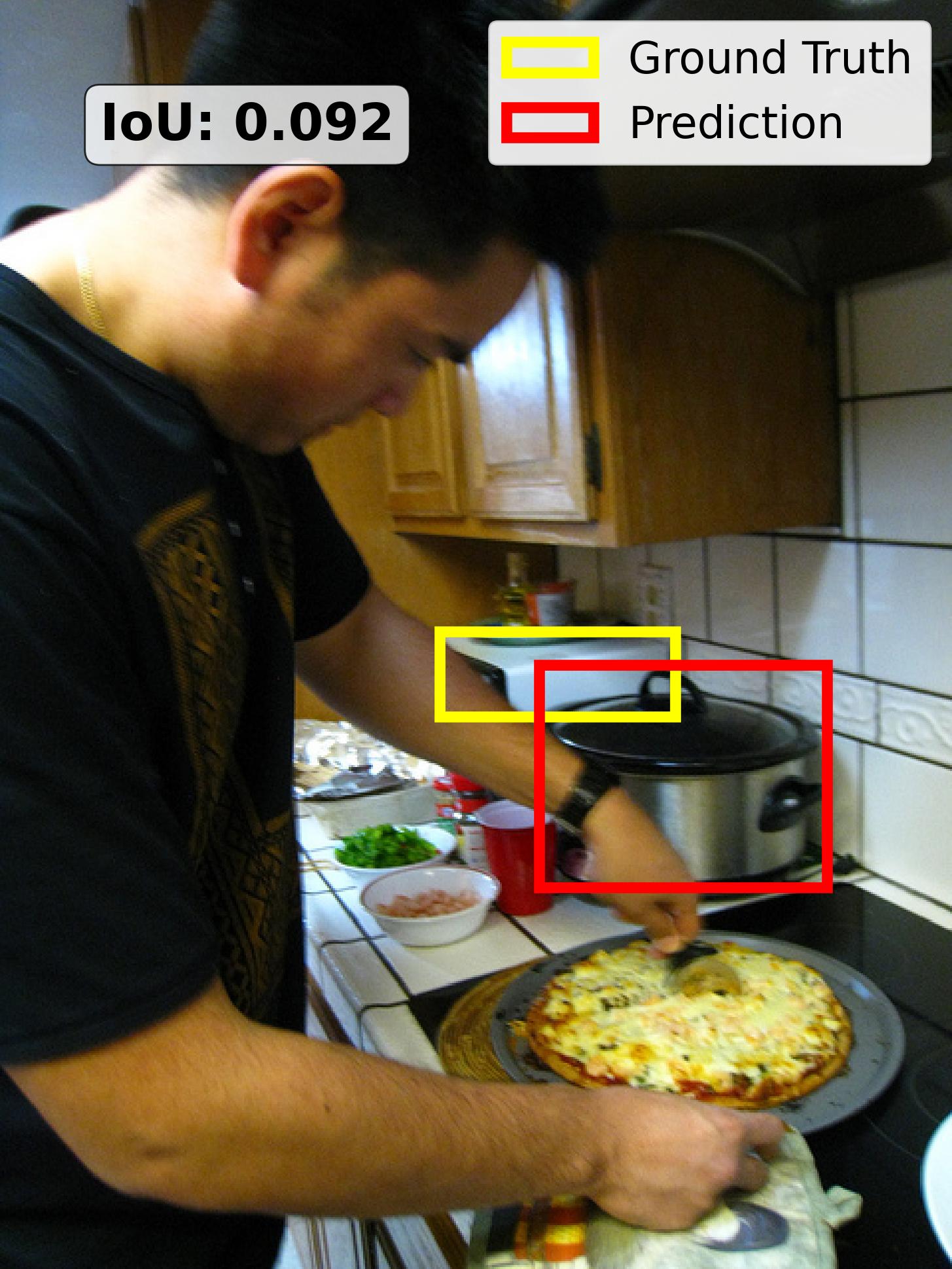}&
\includegraphics[height=1.5in]{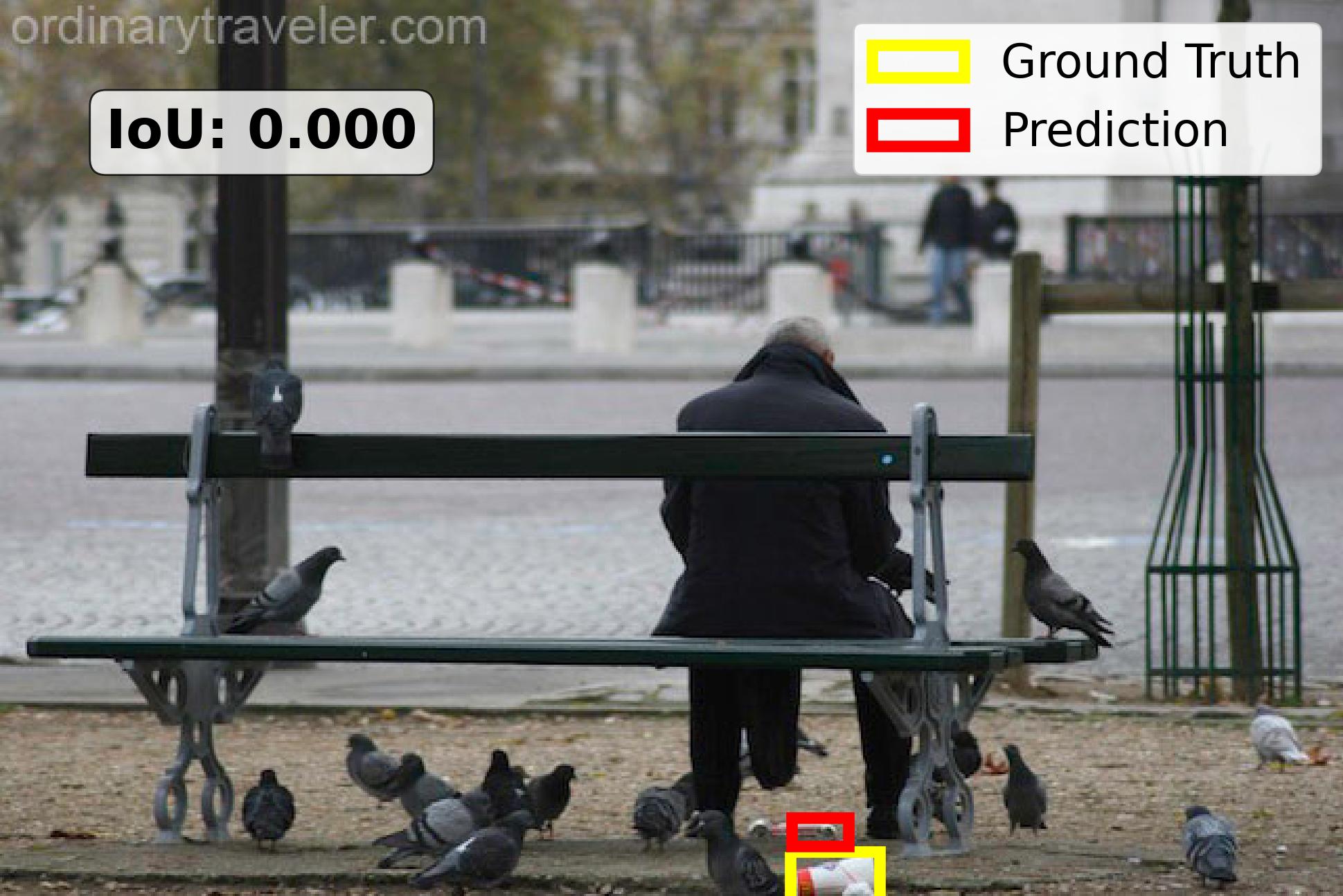}&
\includegraphics[height=1.5in]{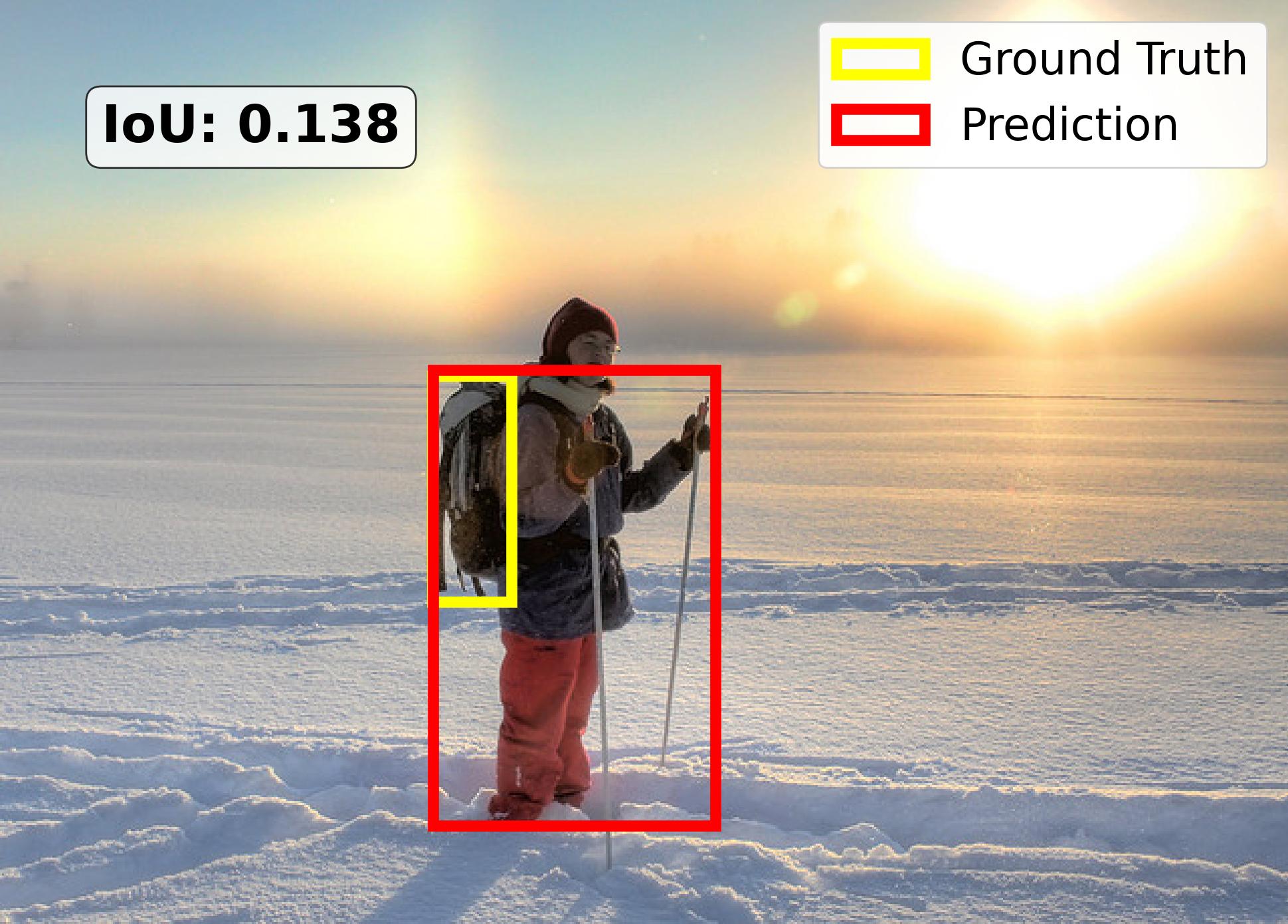}\\
\includegraphics[height=1.5in]{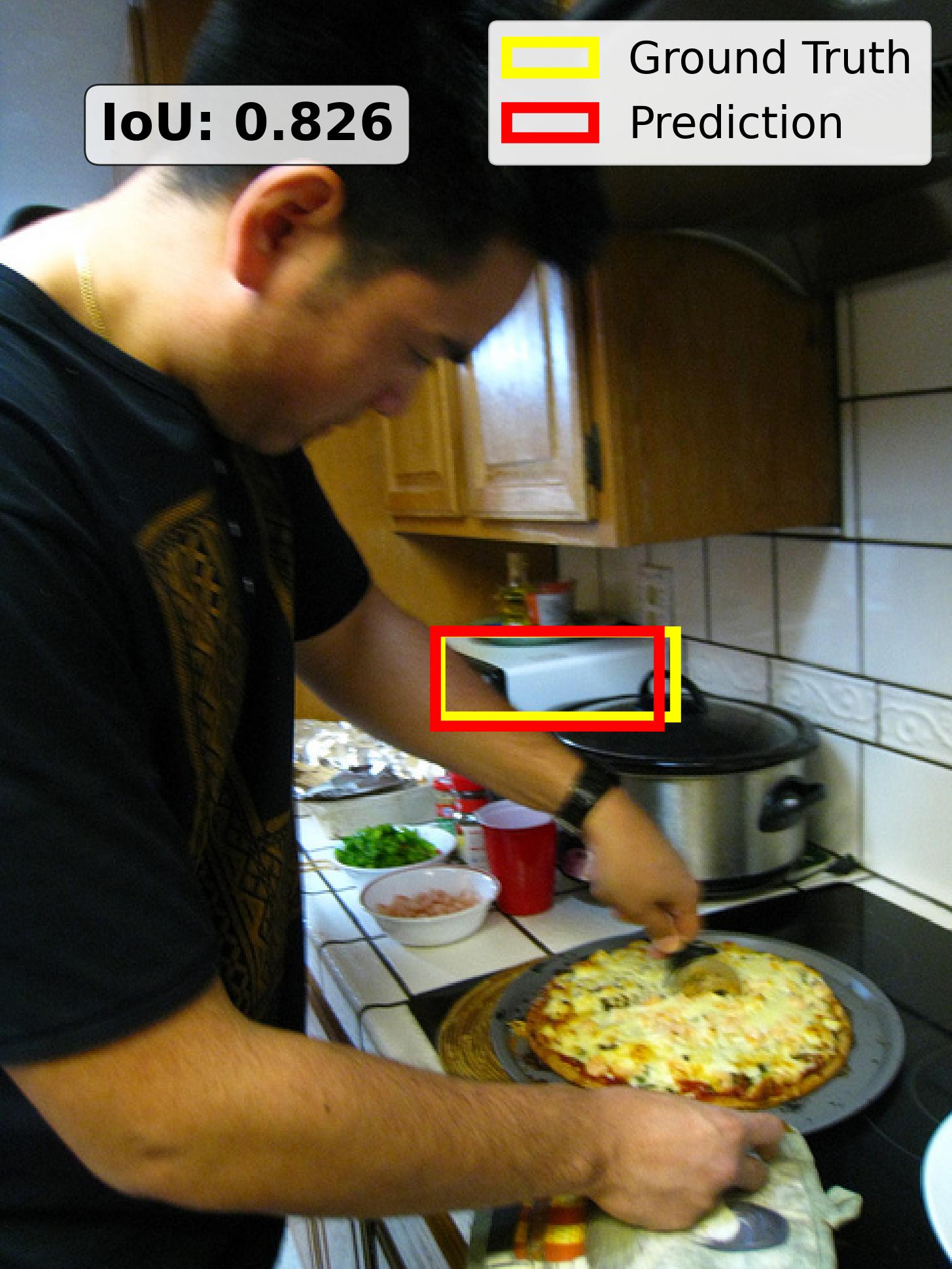}&
\includegraphics[height=1.5in]{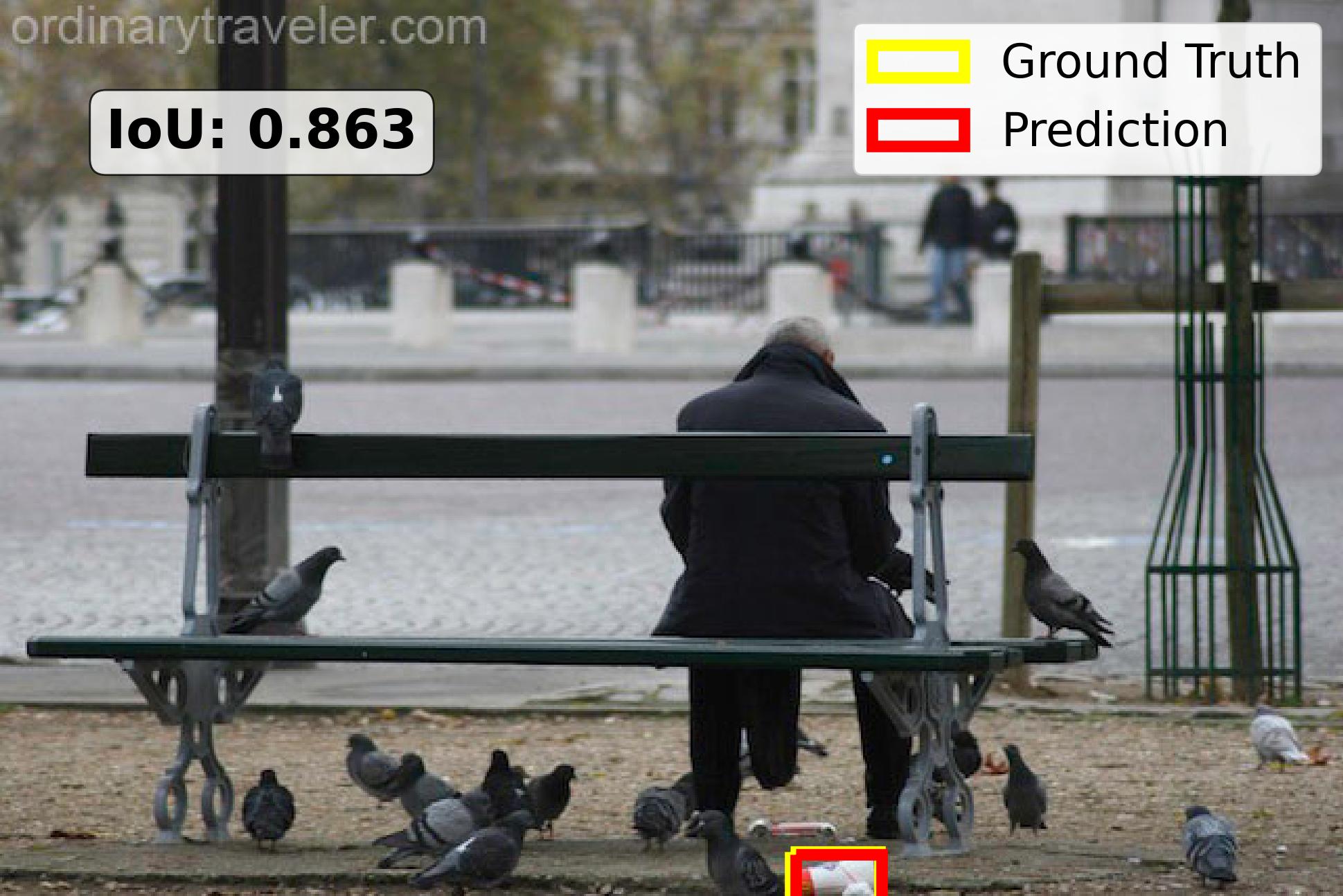}&
\includegraphics[height=1.5in]{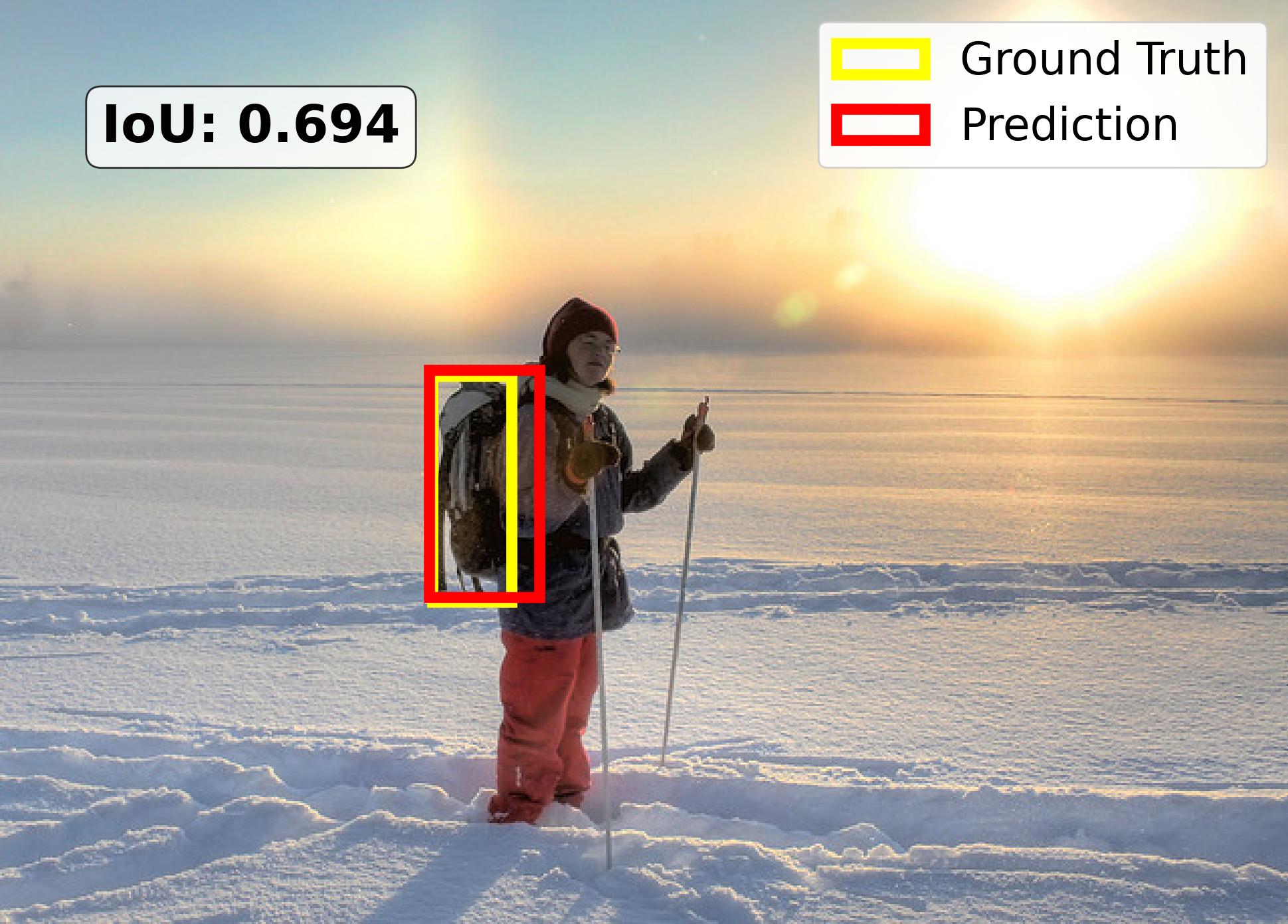}\\
\end{tabular}
\caption{
Additional qualitative comparisons on COCO using Qwen2.5-VL.
Each case shows Greedy (top) and Ours (bottom).
}

\label{fig:qualitative_suppl_page1_coco_qwen}
\end{figure*}

\begin{figure*}[t]
\centering

\begin{tabular}{ccc}
Bus & Tie & Sink\\
\includegraphics[height=1.35in]{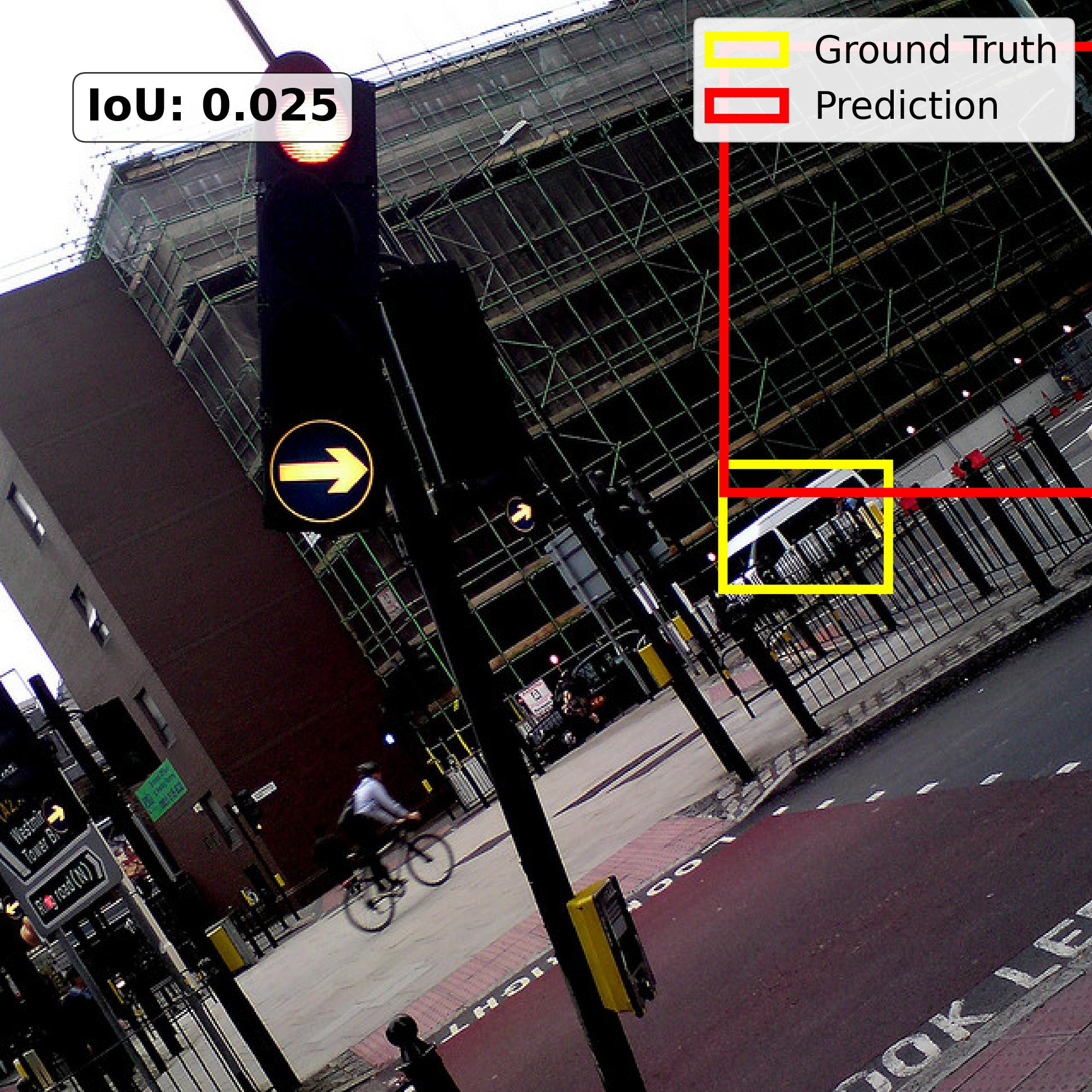} &
\includegraphics[height=1.35in]{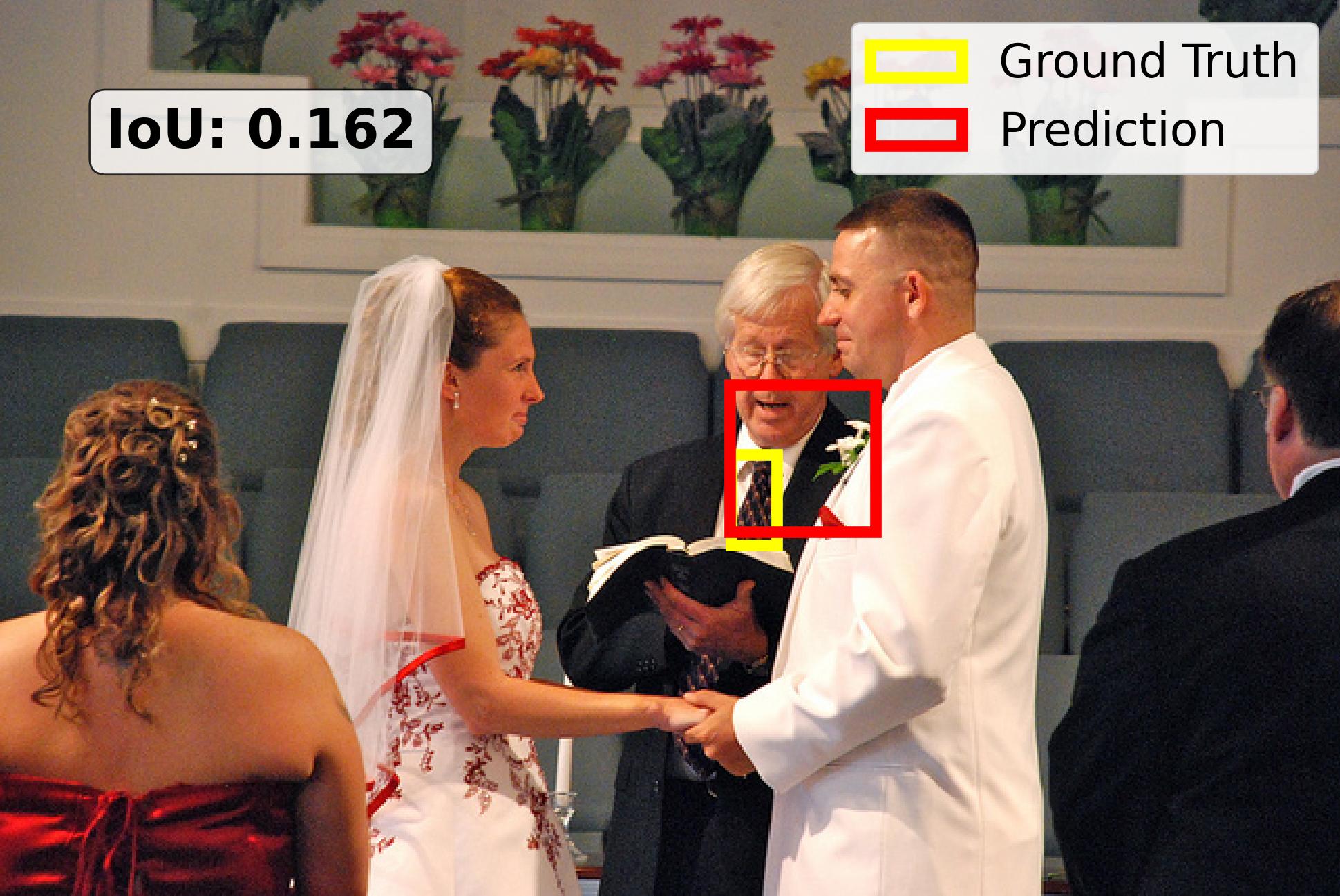} &
\includegraphics[height=1.35in]{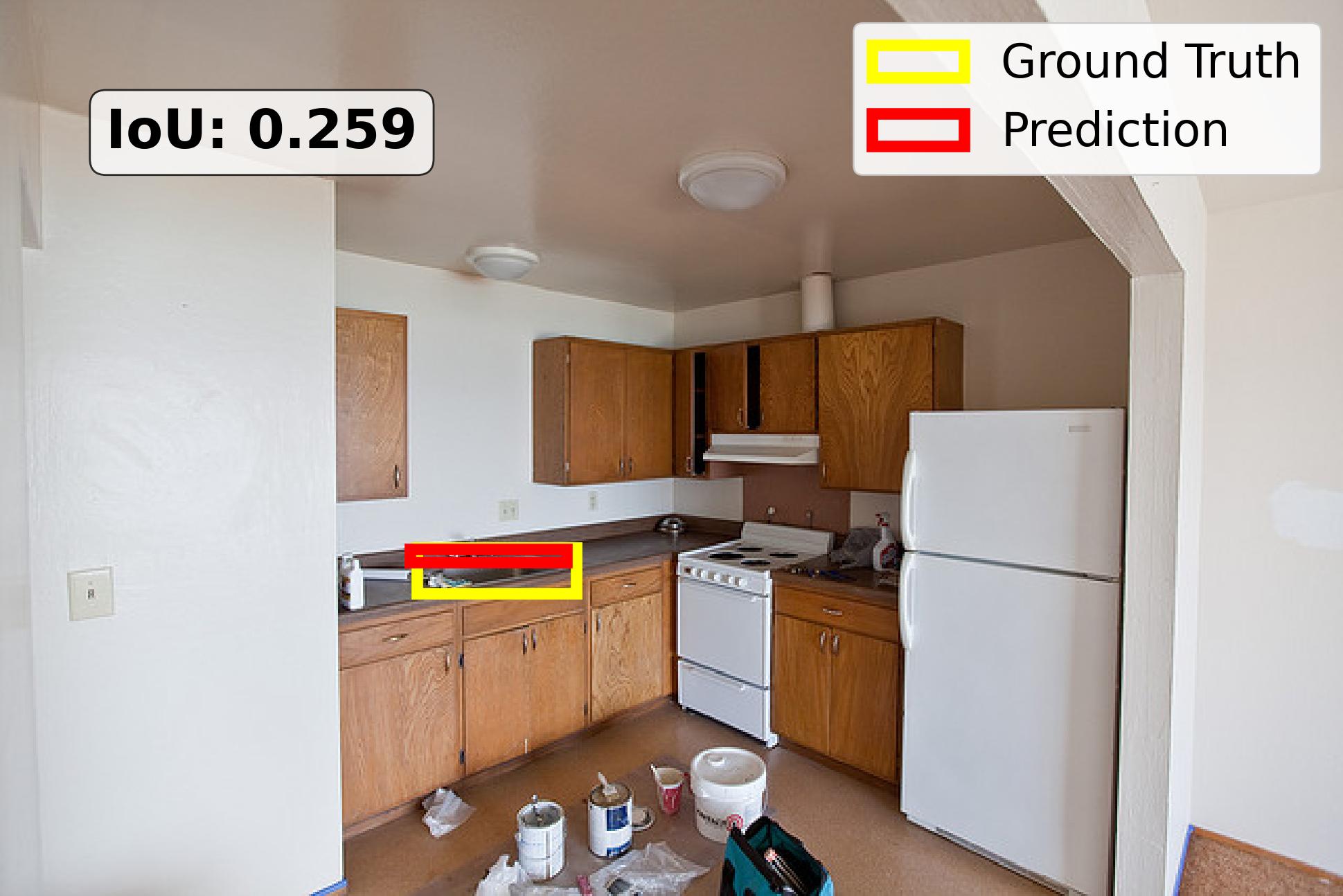}\\
\includegraphics[height=1.35in]{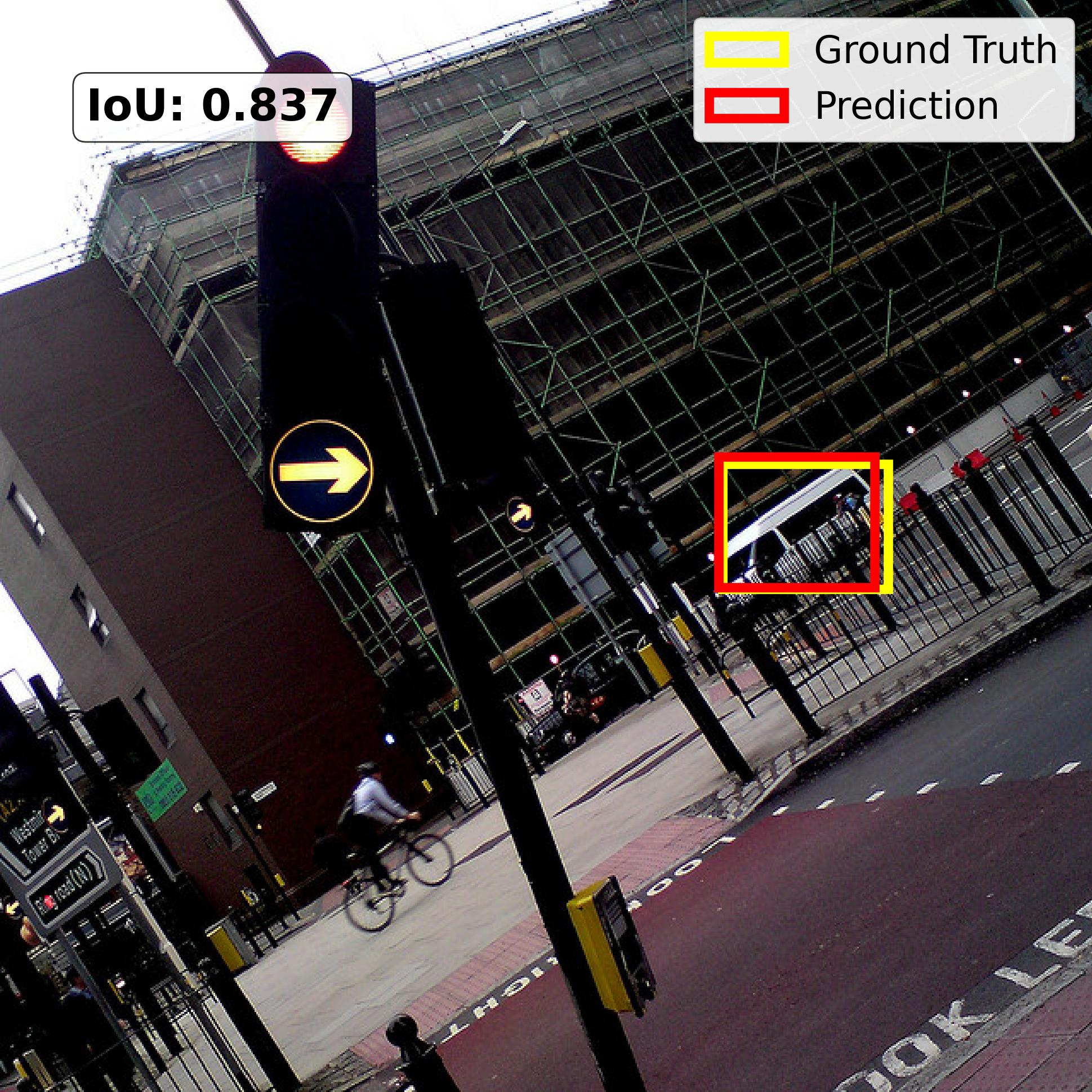} &
\includegraphics[height=1.35in]{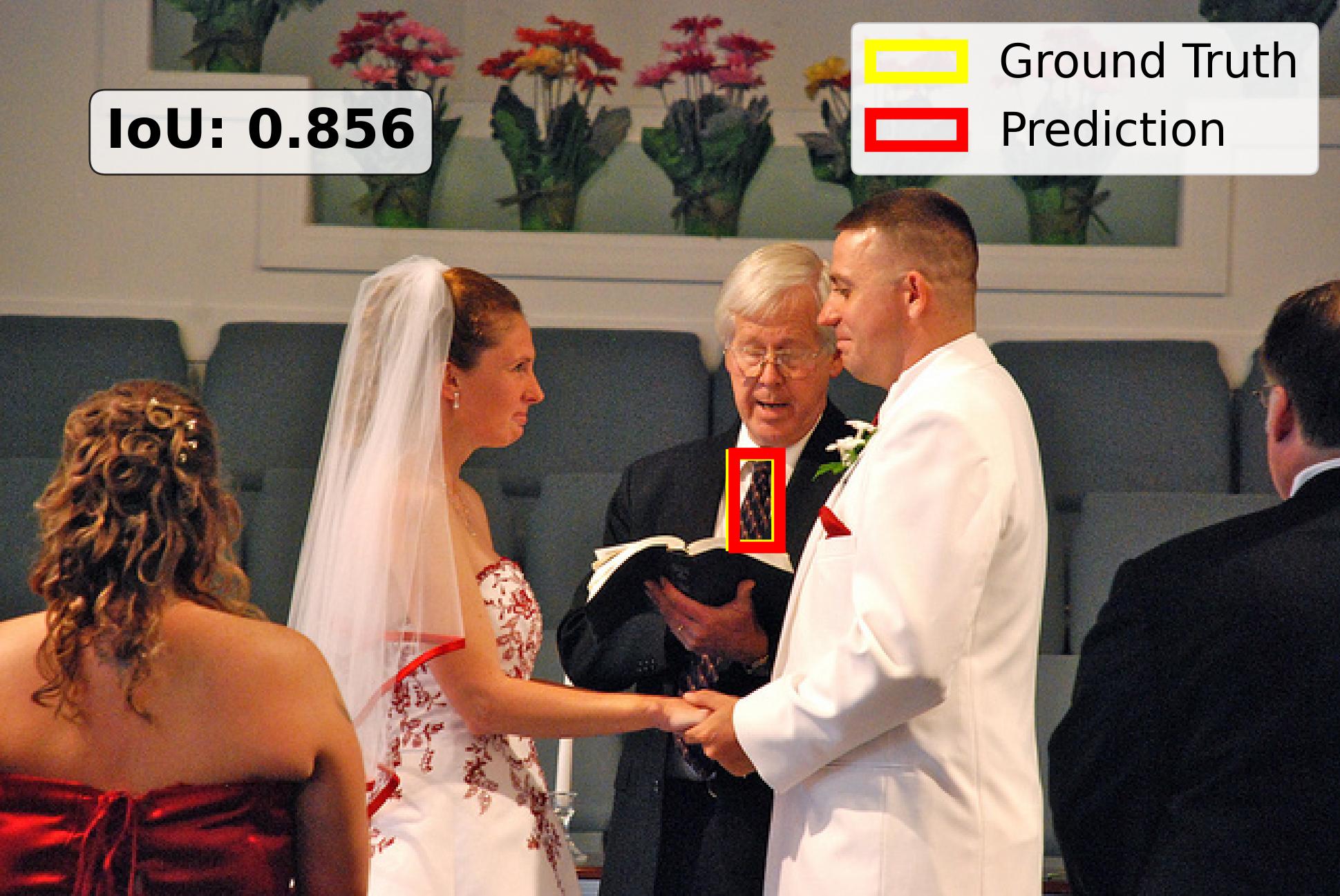} &
\includegraphics[height=1.35in]{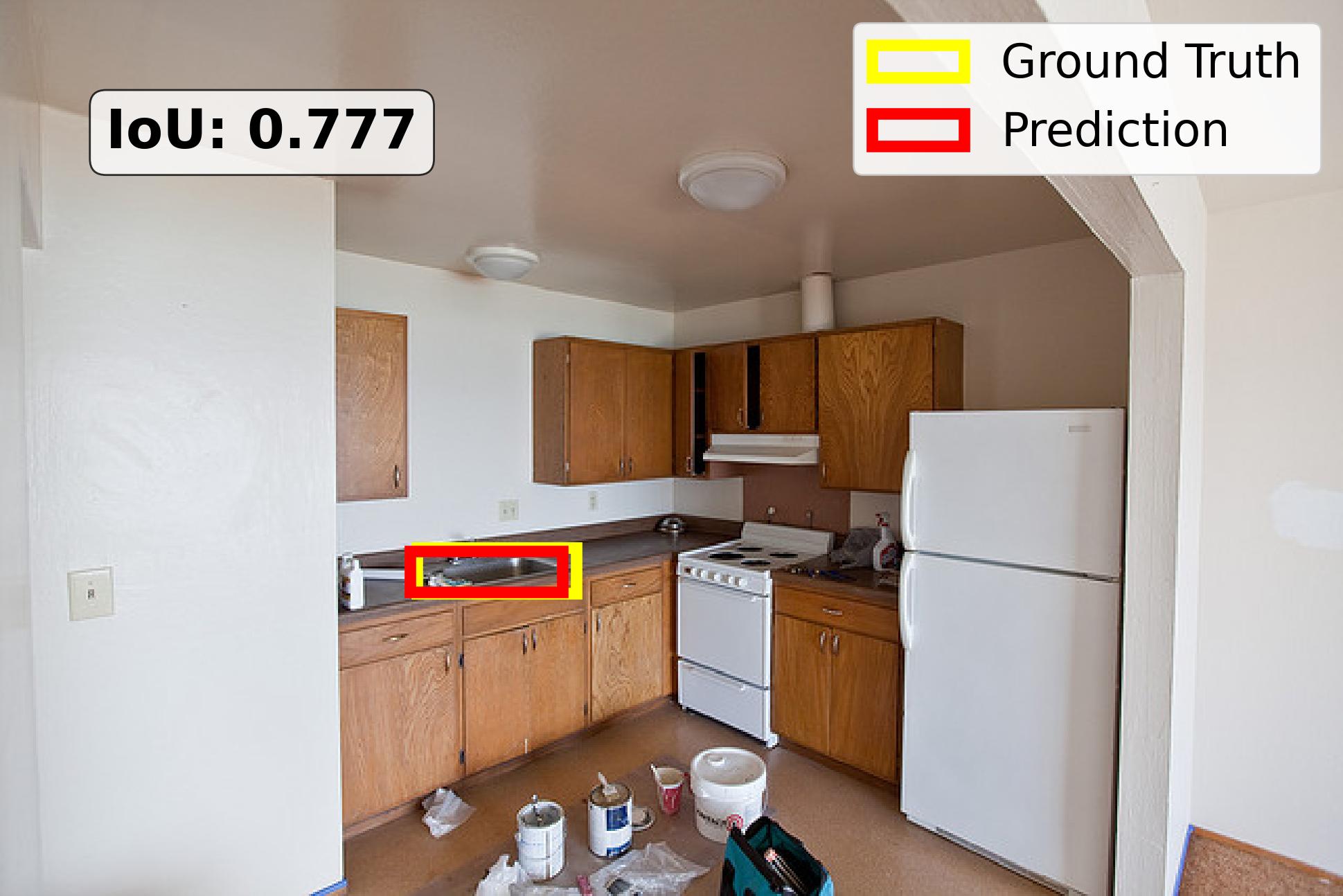}\\
Cup & Bottle & Snowboard\\
\includegraphics[height=1.35in]{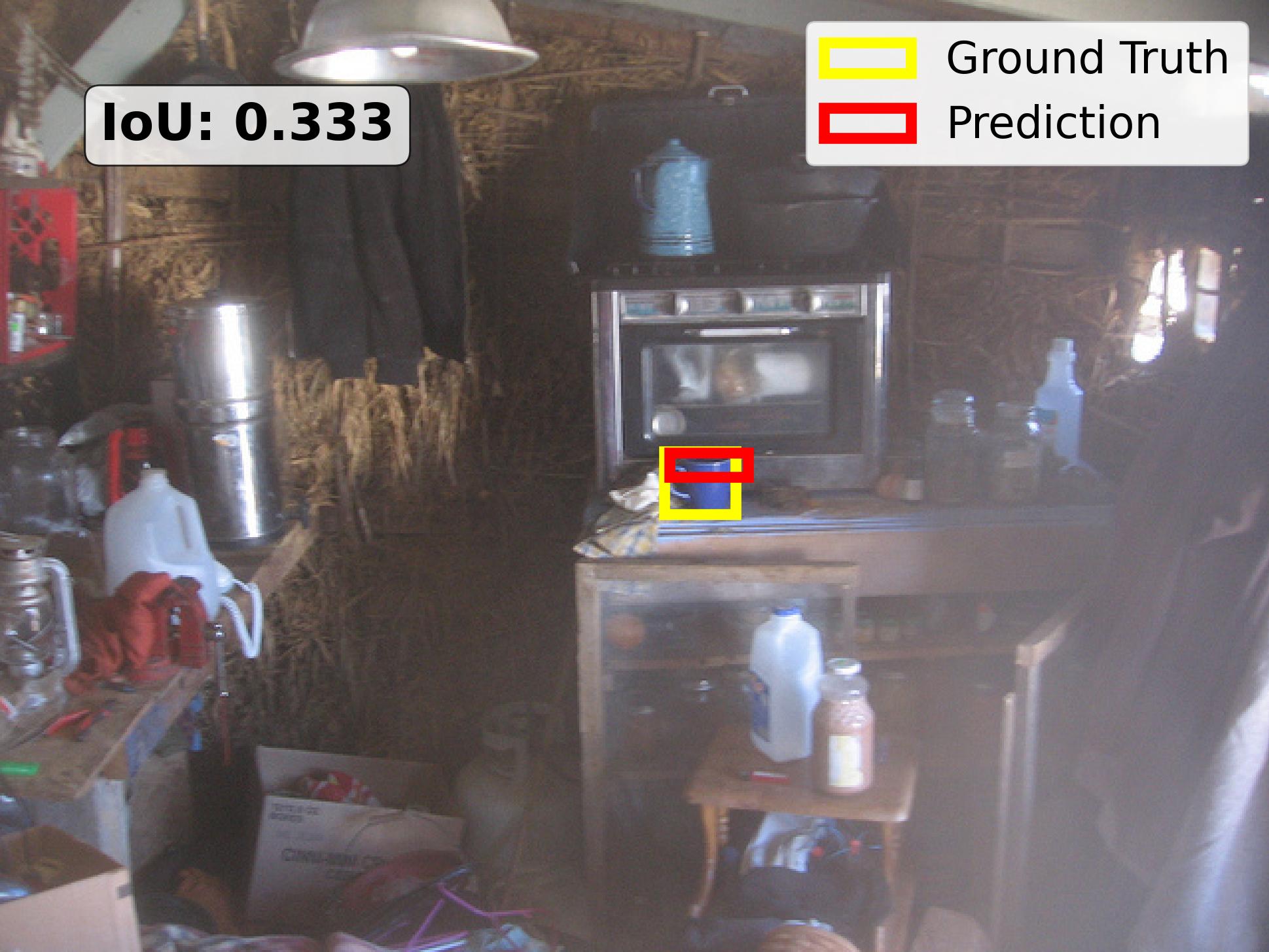}&
\includegraphics[height=1.35in]{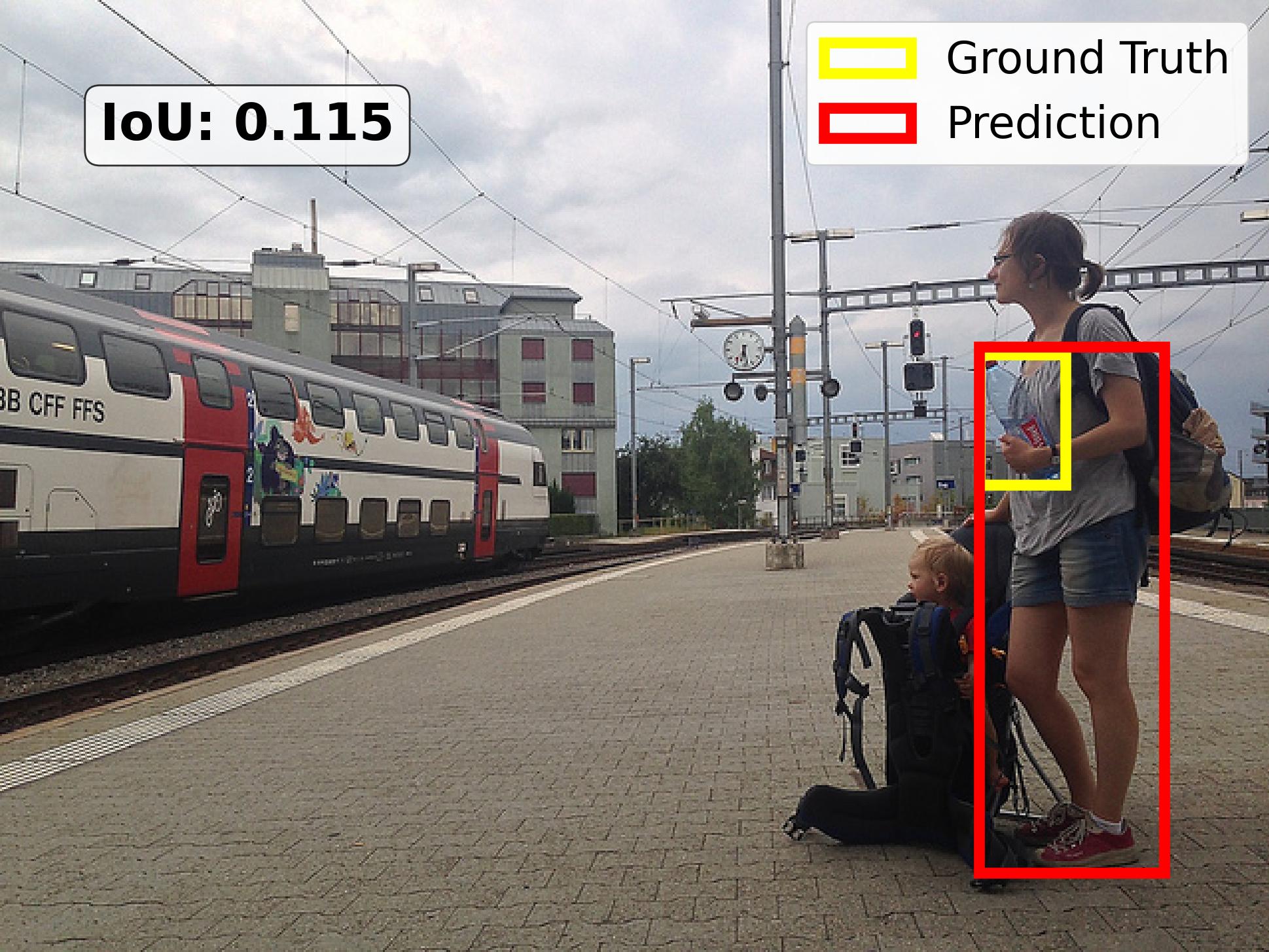}&
\includegraphics[height=1.35in]{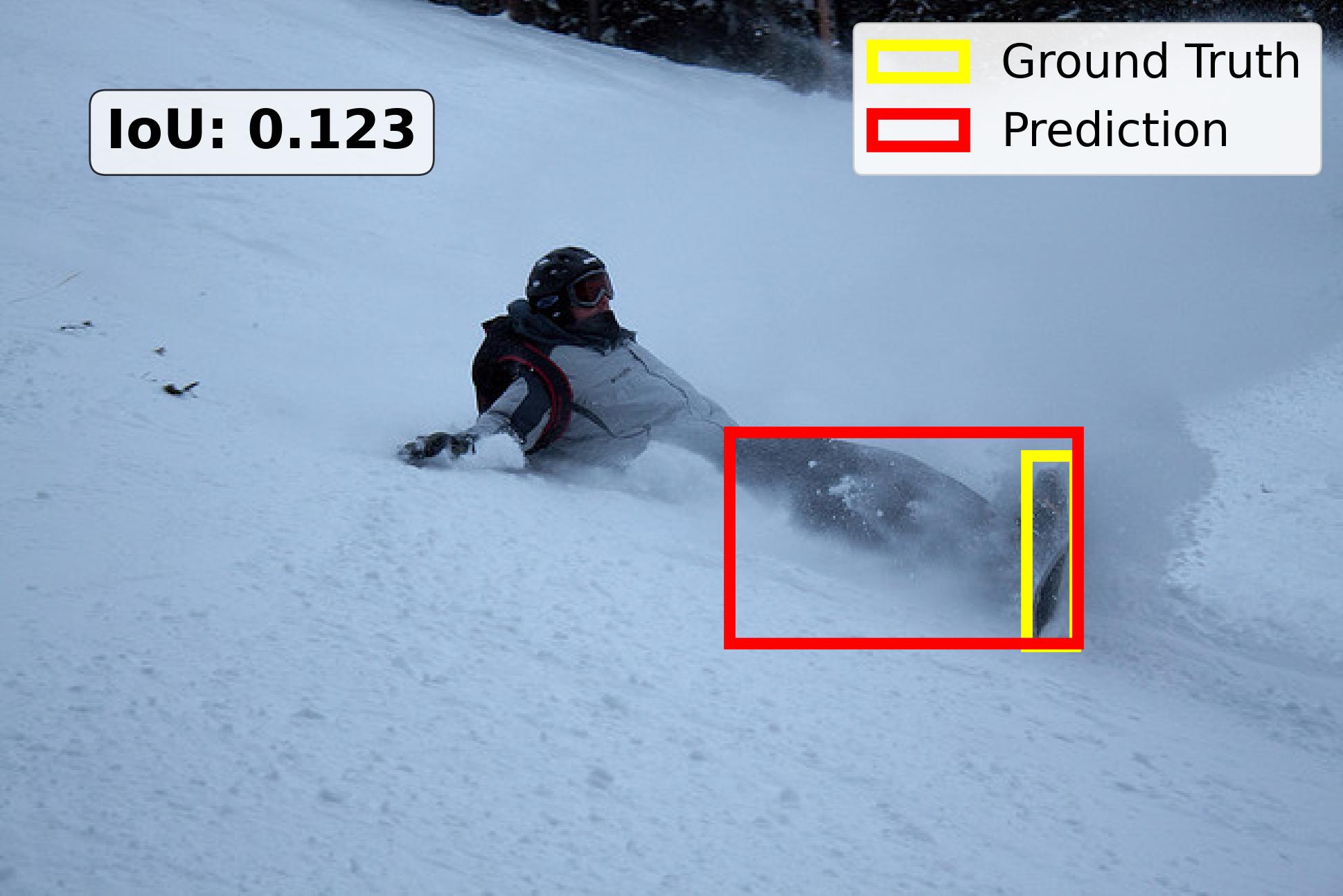}\\
\includegraphics[height=1.35in]{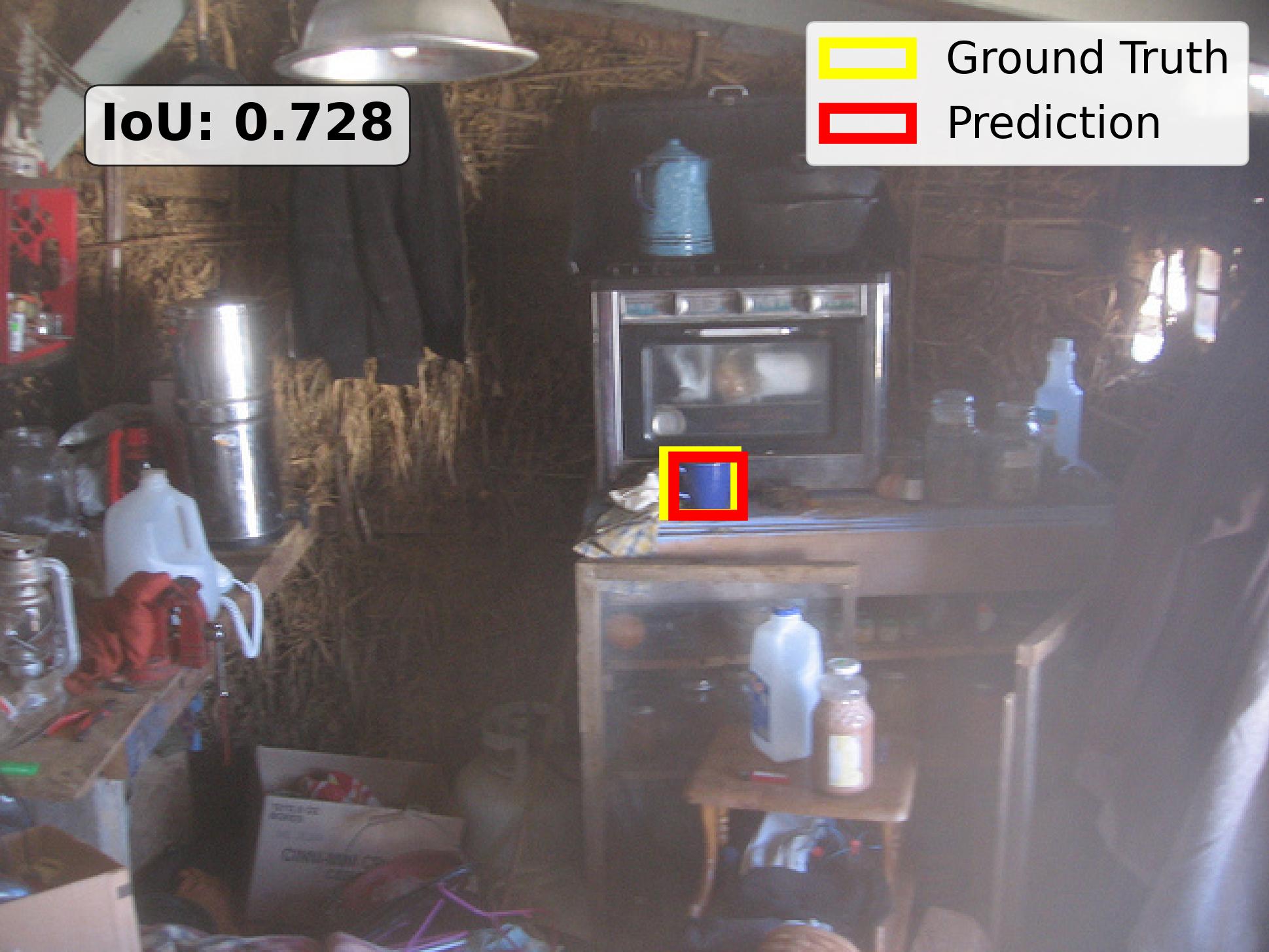}&
\includegraphics[height=1.35in]{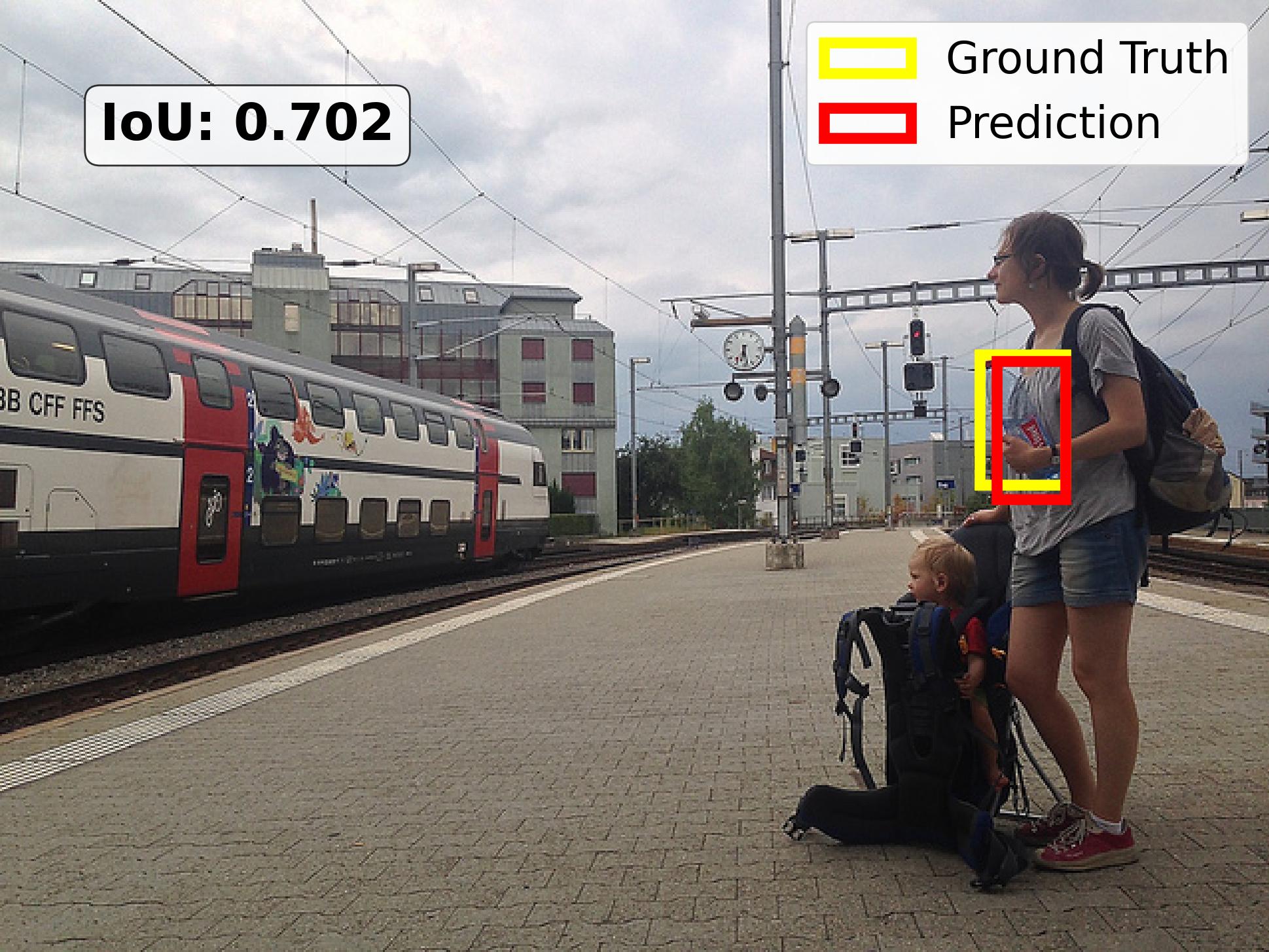}&
\includegraphics[height=1.35in]{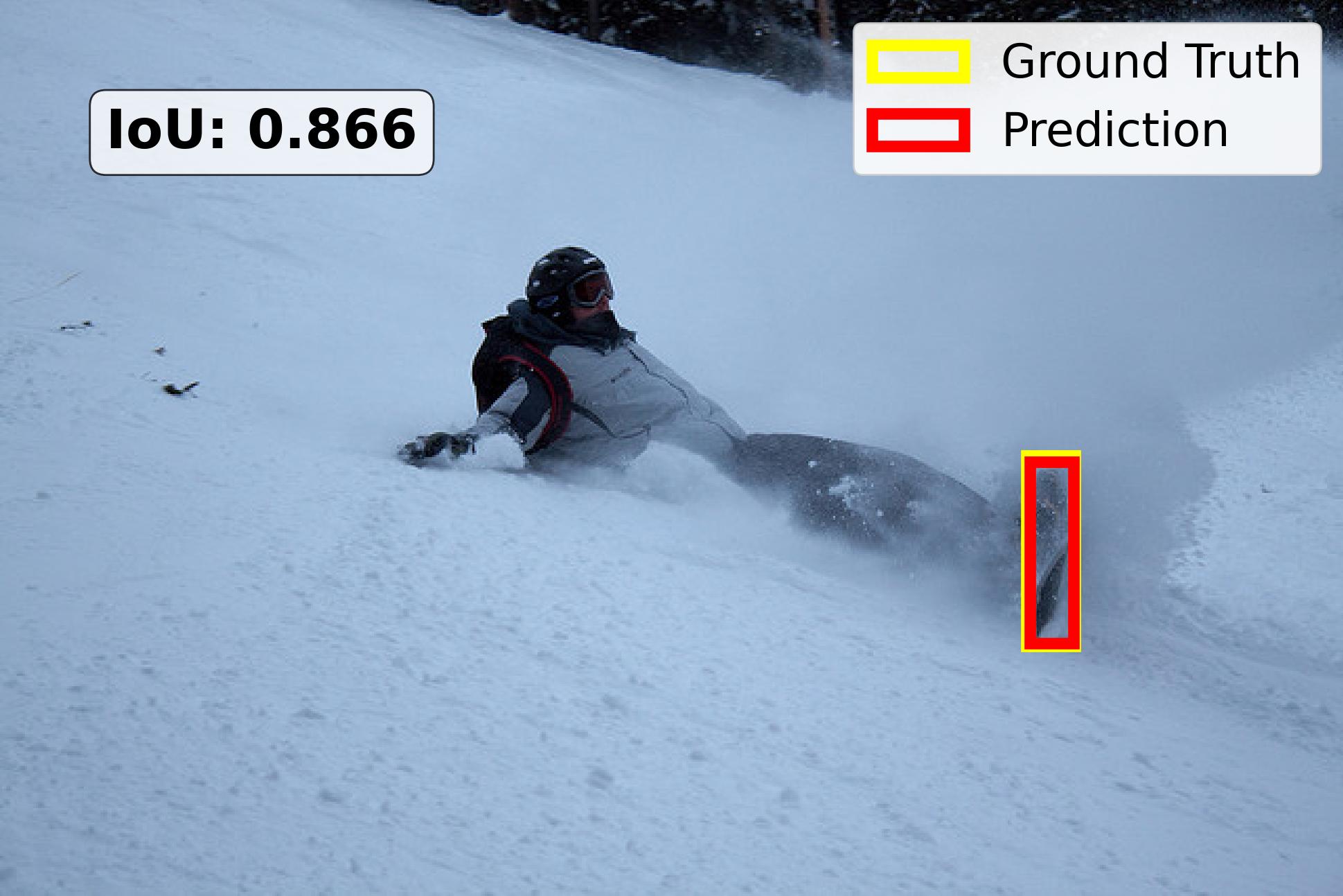}\\
\end{tabular}



\caption{
Additional qualitative comparisons on COCO using Qwen2.5-VL.
Each case shows Greedy (top) and Ours (bottom).
}
\label{fig:qualitative_suppl_page2_coco_qwen}
\end{figure*}

\begin{figure*}[t]
\centering
\begin{tabular}{ccc}
Snowboard & Baseball Glove & Person\\
\includegraphics[height=1.5in]{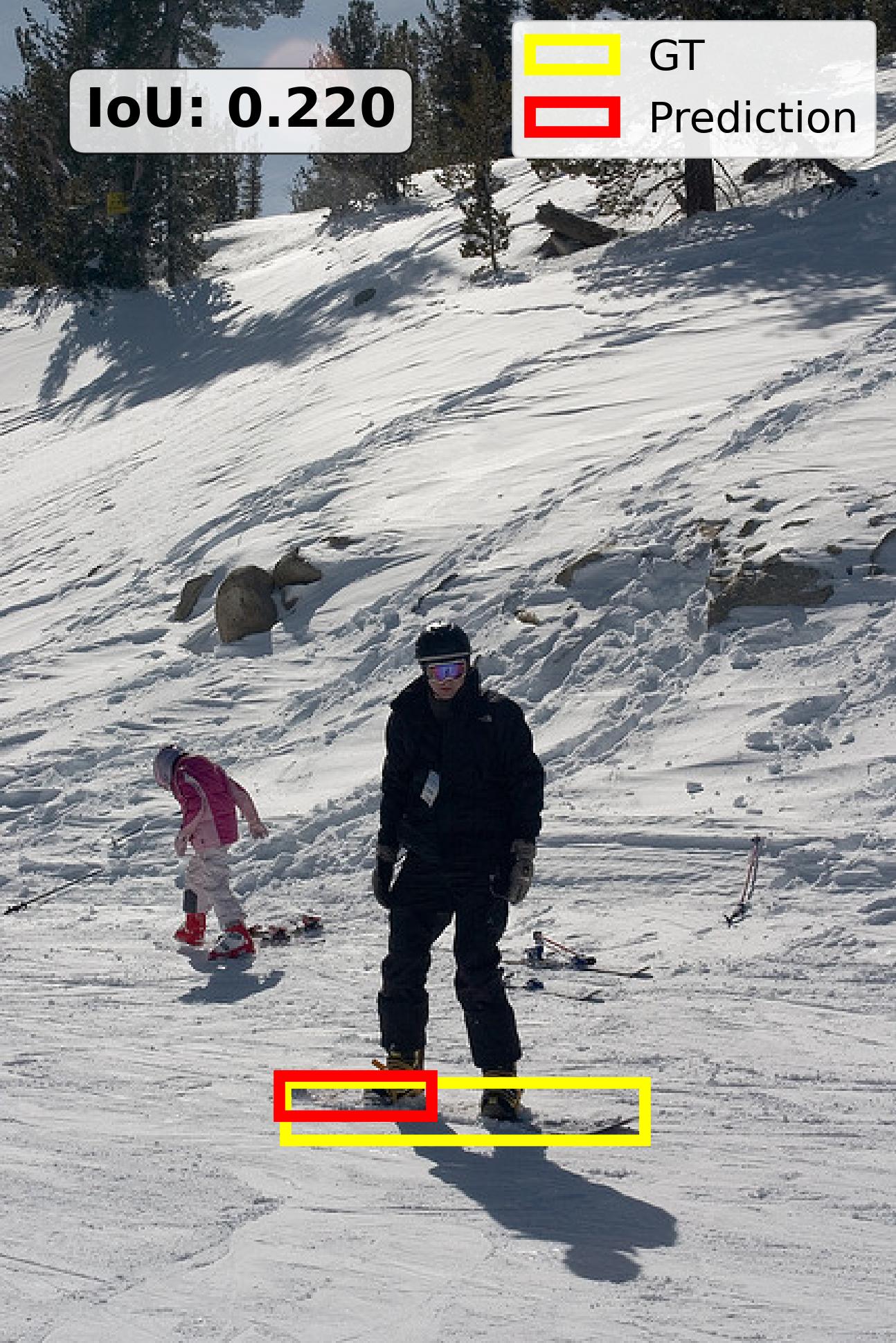} &
\includegraphics[height=1.5in]{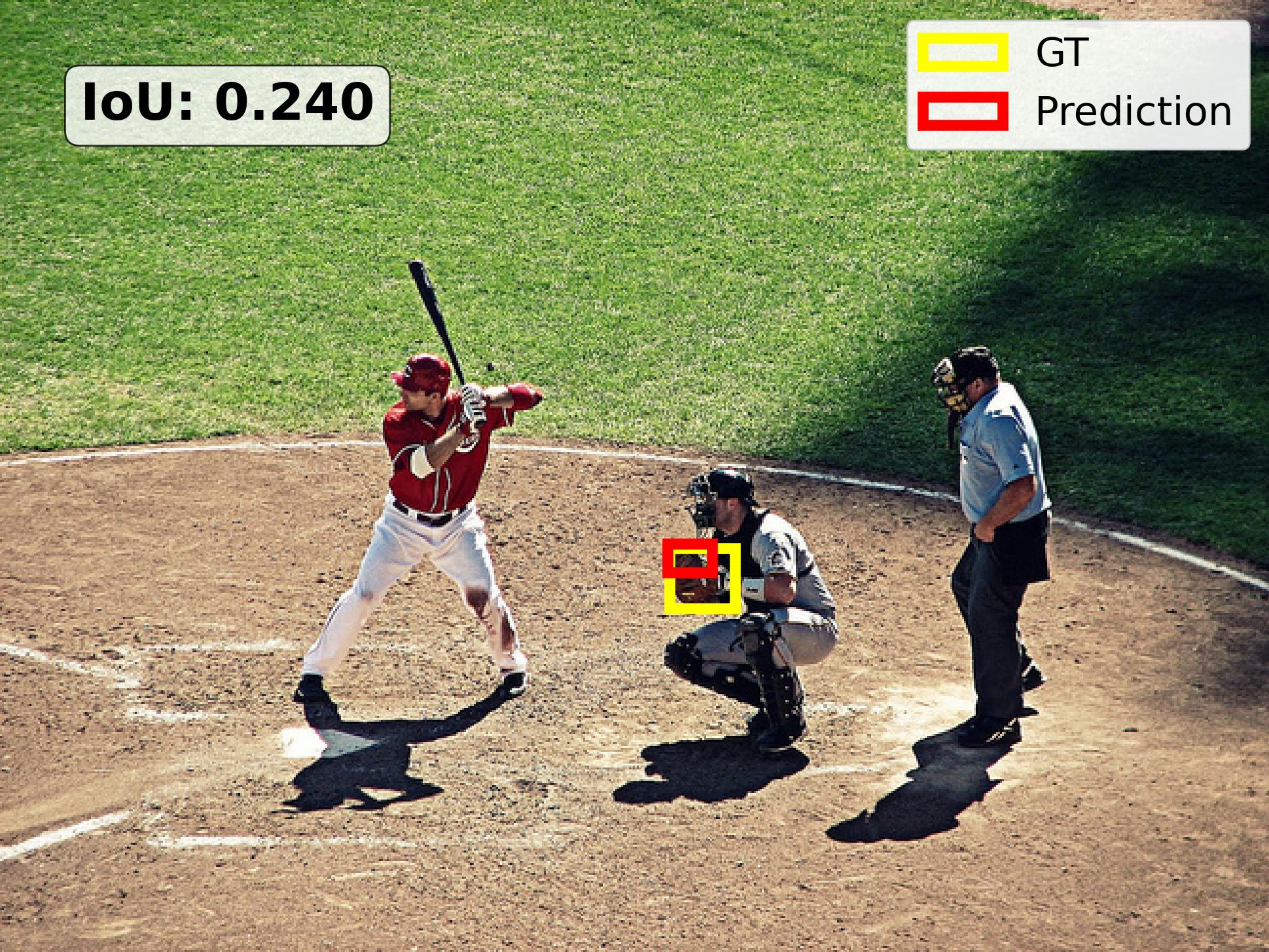} &
\includegraphics[height=1.5in]{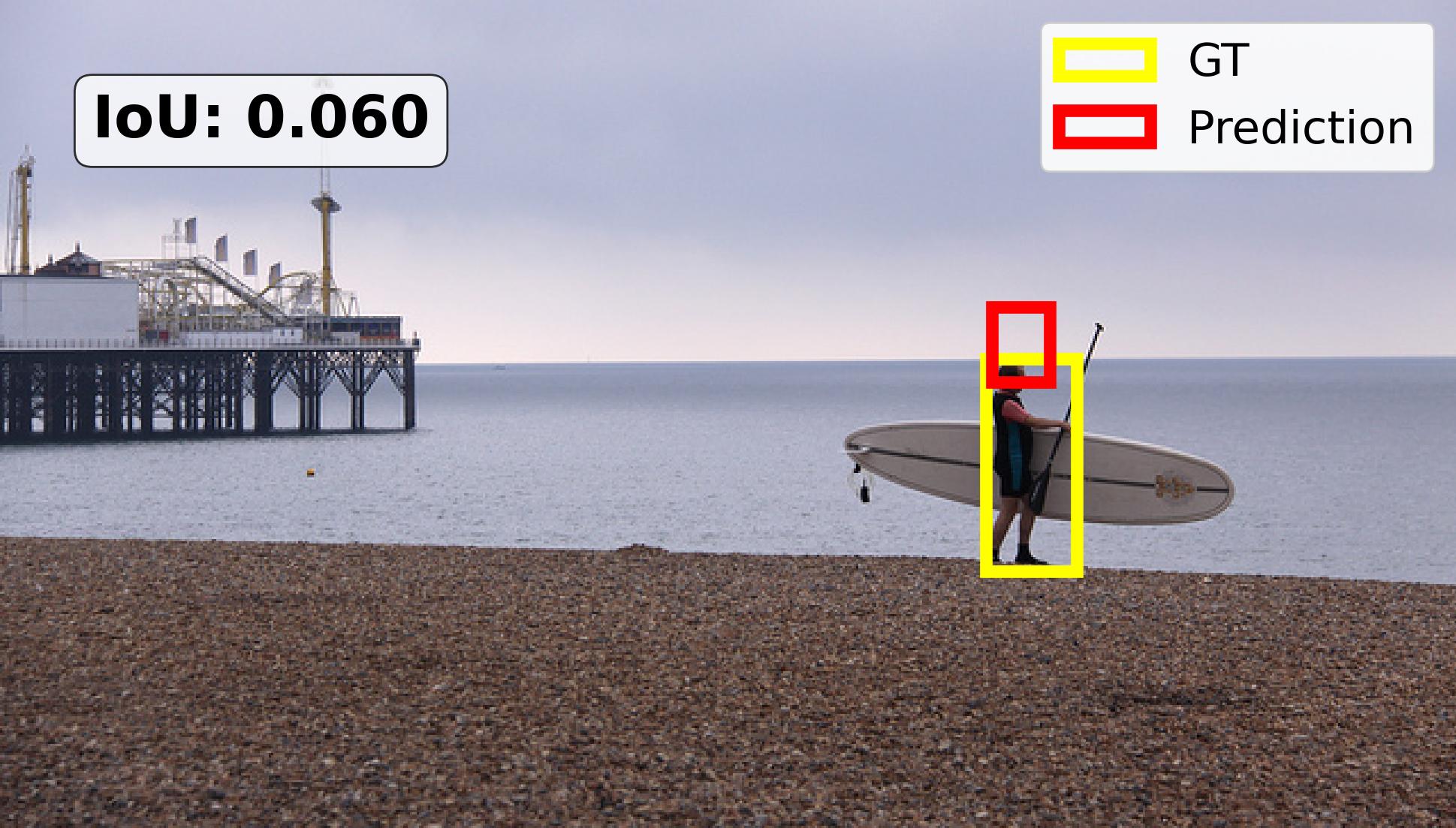}\\
\includegraphics[height=1.5in]{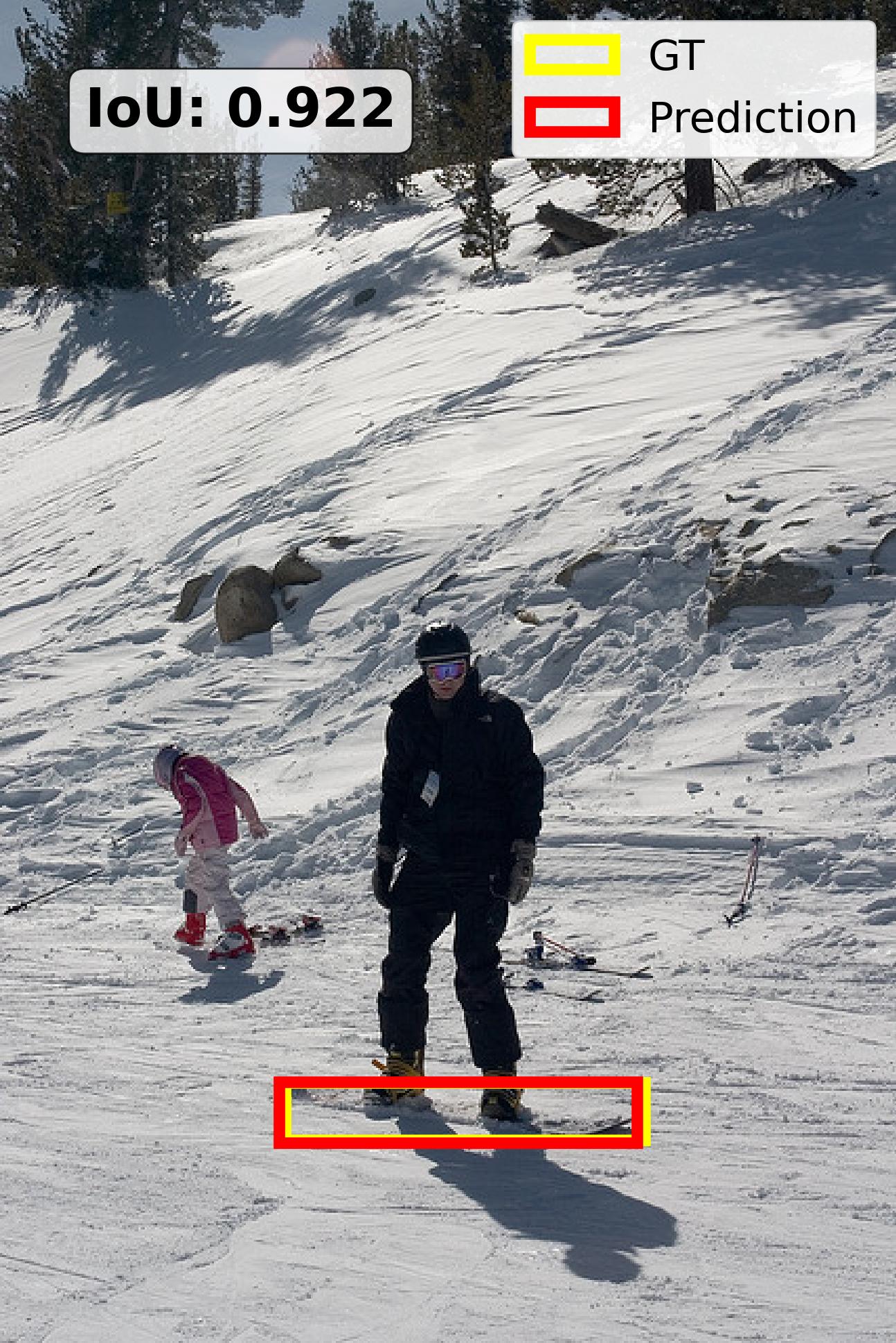} &
\includegraphics[height=1.5in]{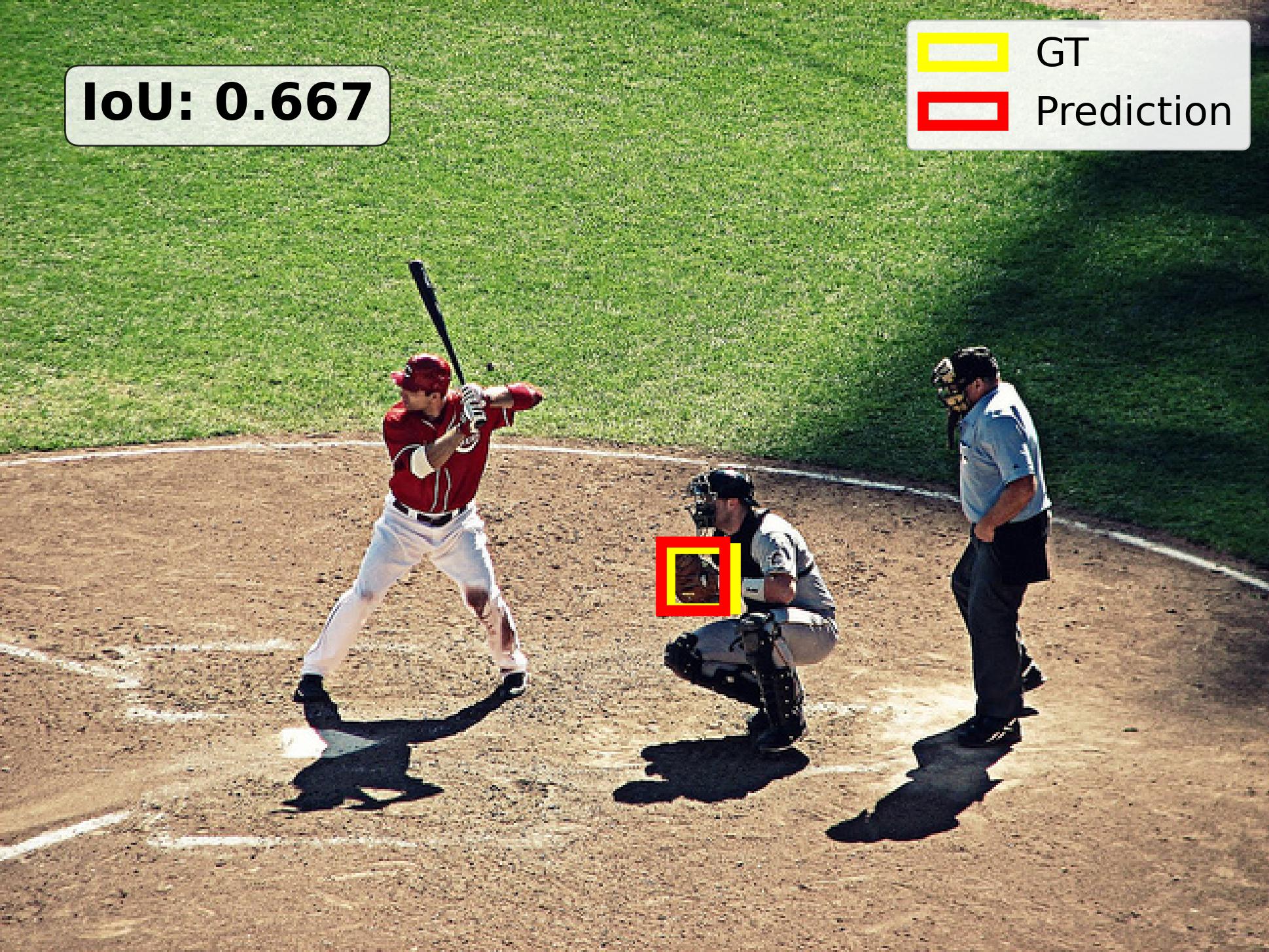} &
\includegraphics[height=1.5in]{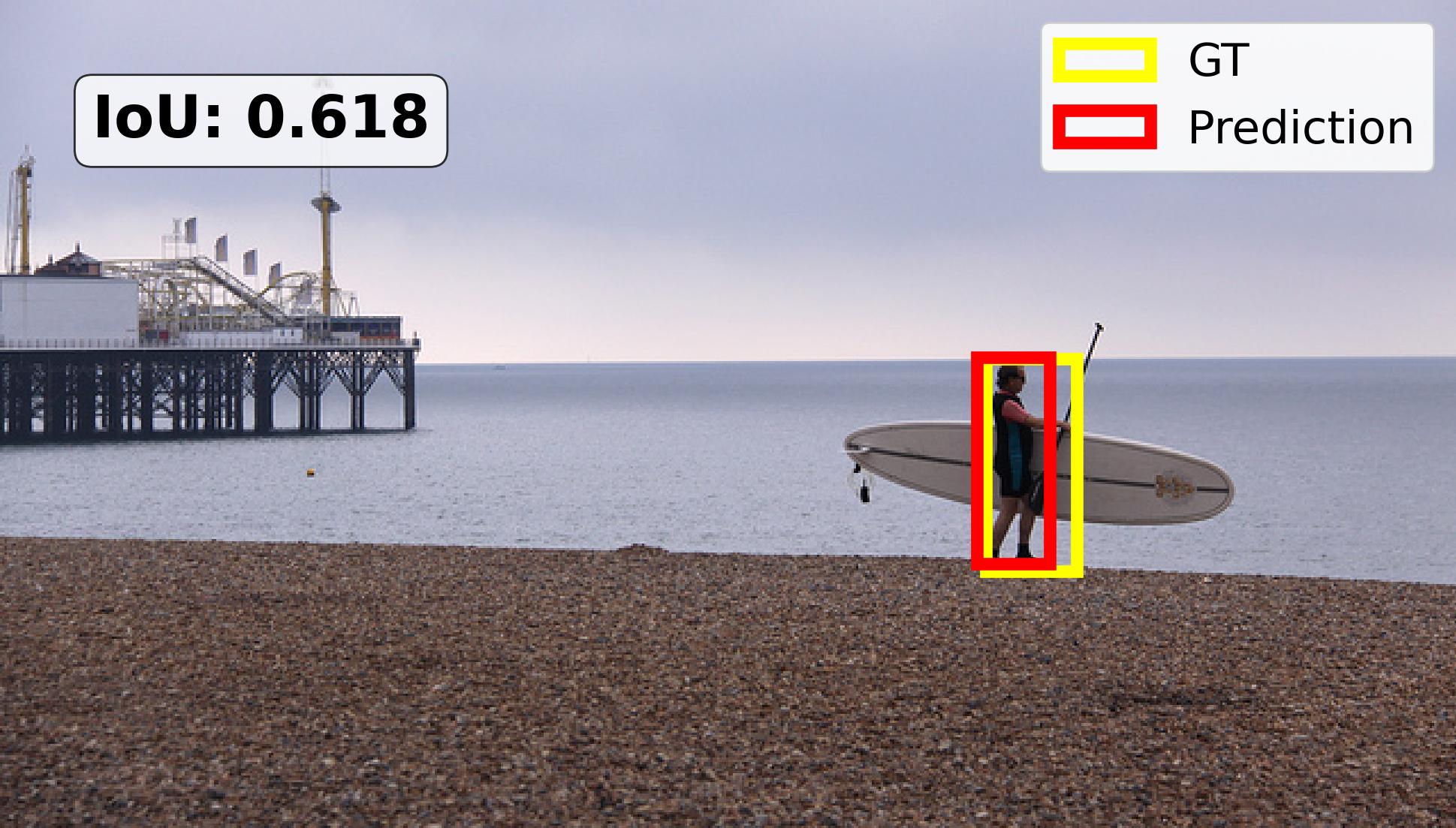}\\

Cell Phone & Clock & Person\\
\includegraphics[height=1.5in]{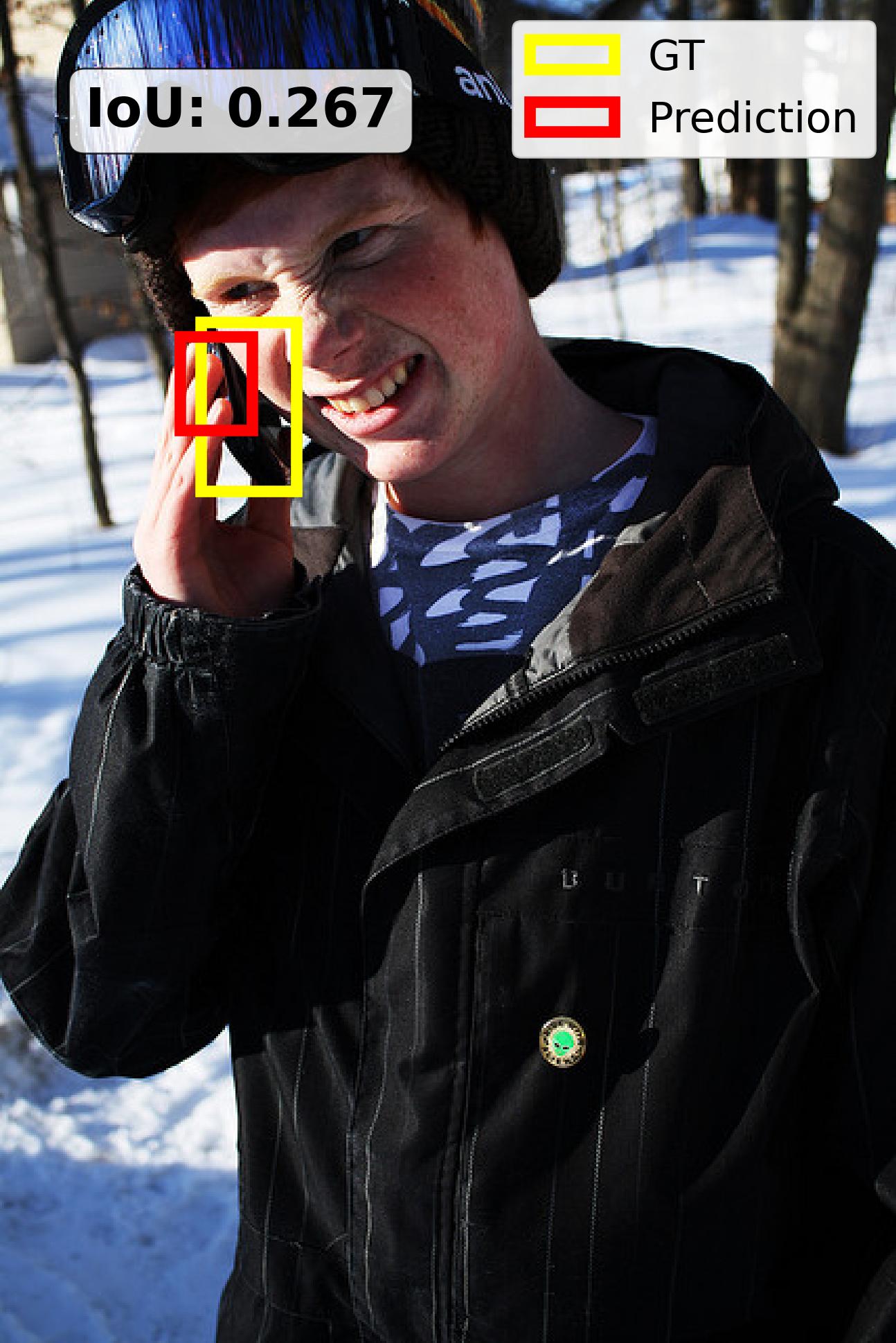}&
\includegraphics[height=1.5in]{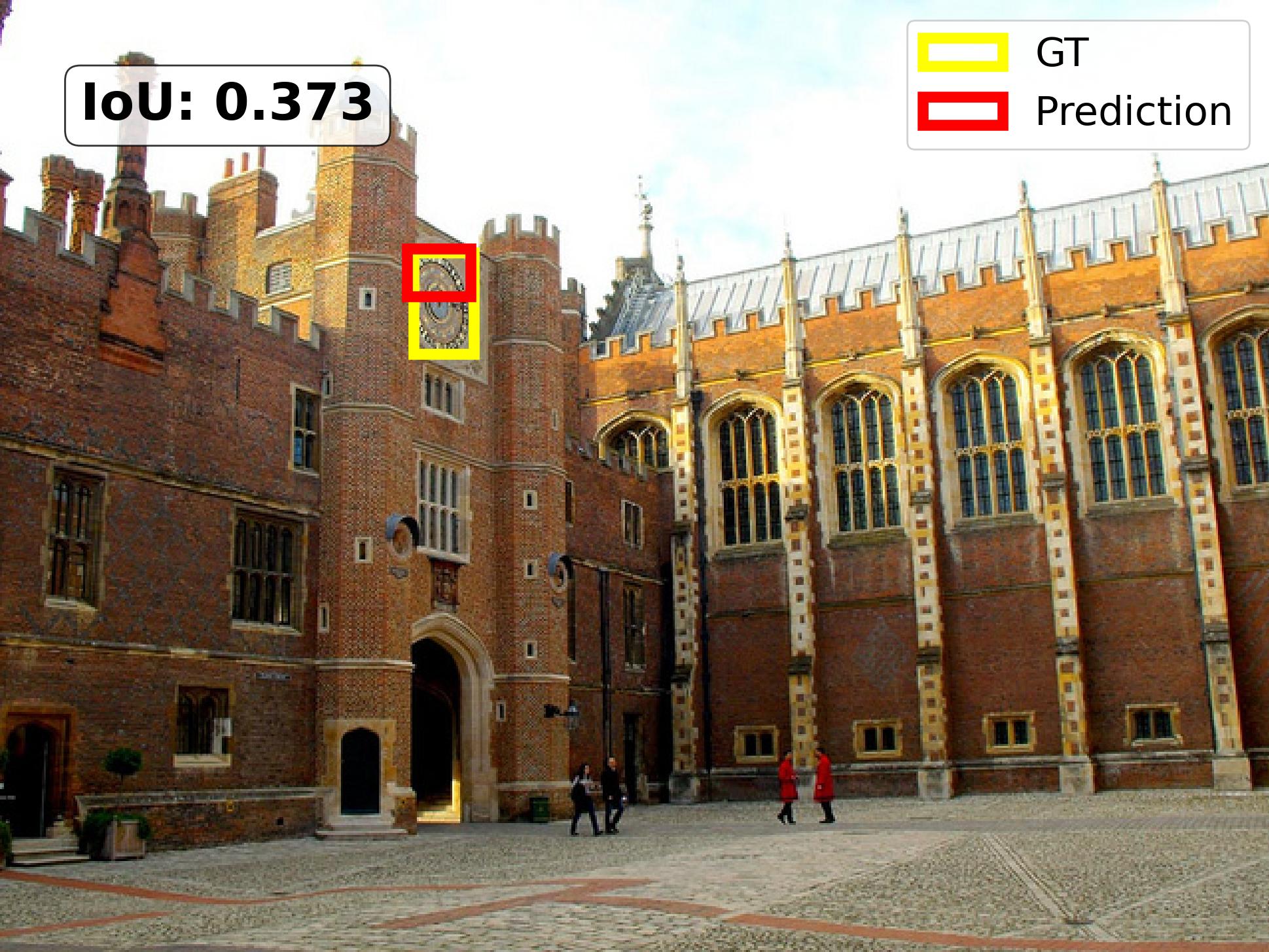}&
\includegraphics[height=1.5in]{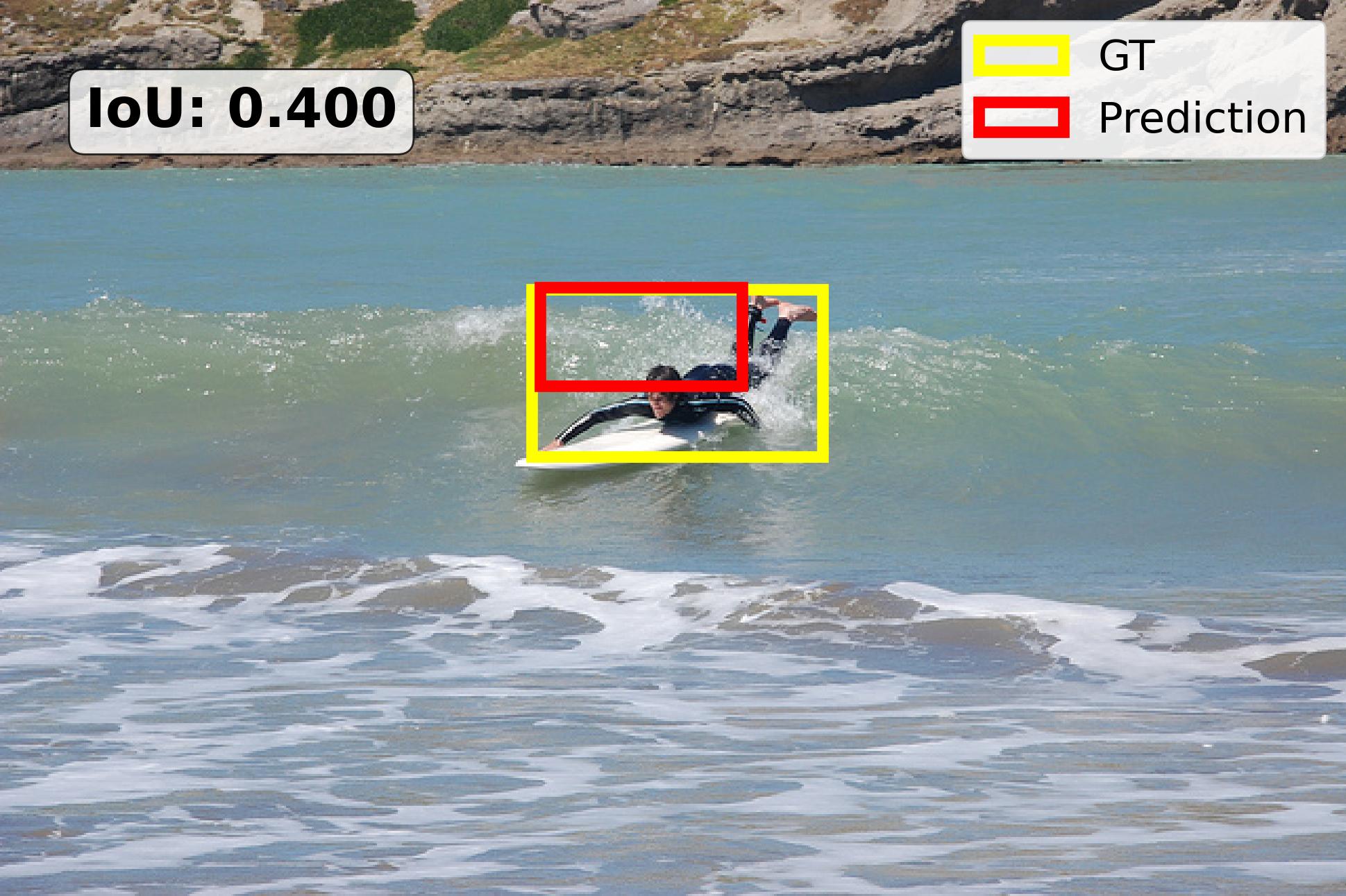}\\
\includegraphics[height=1.5in]{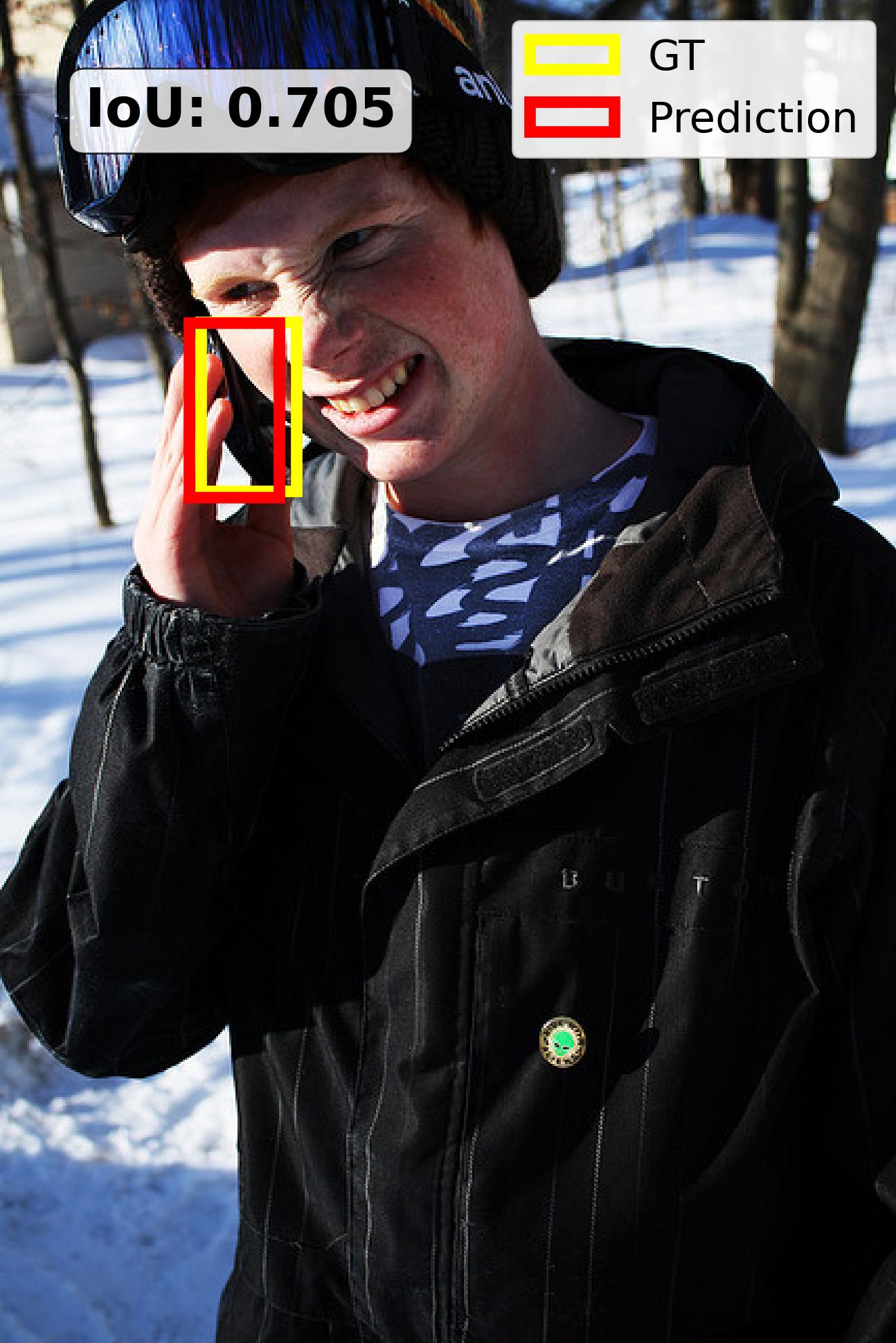}&
\includegraphics[height=1.5in]{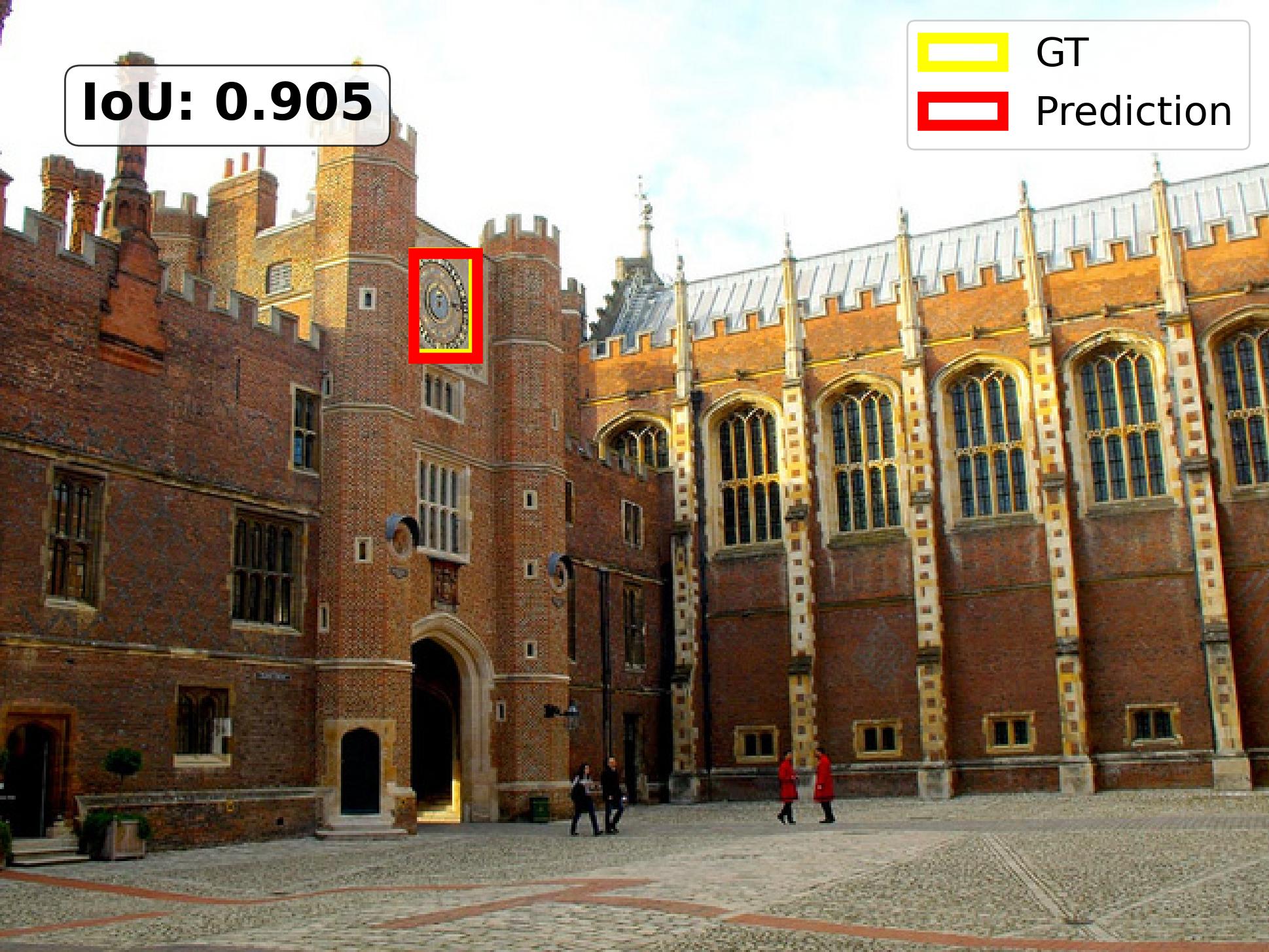}&
\includegraphics[height=1.5in]{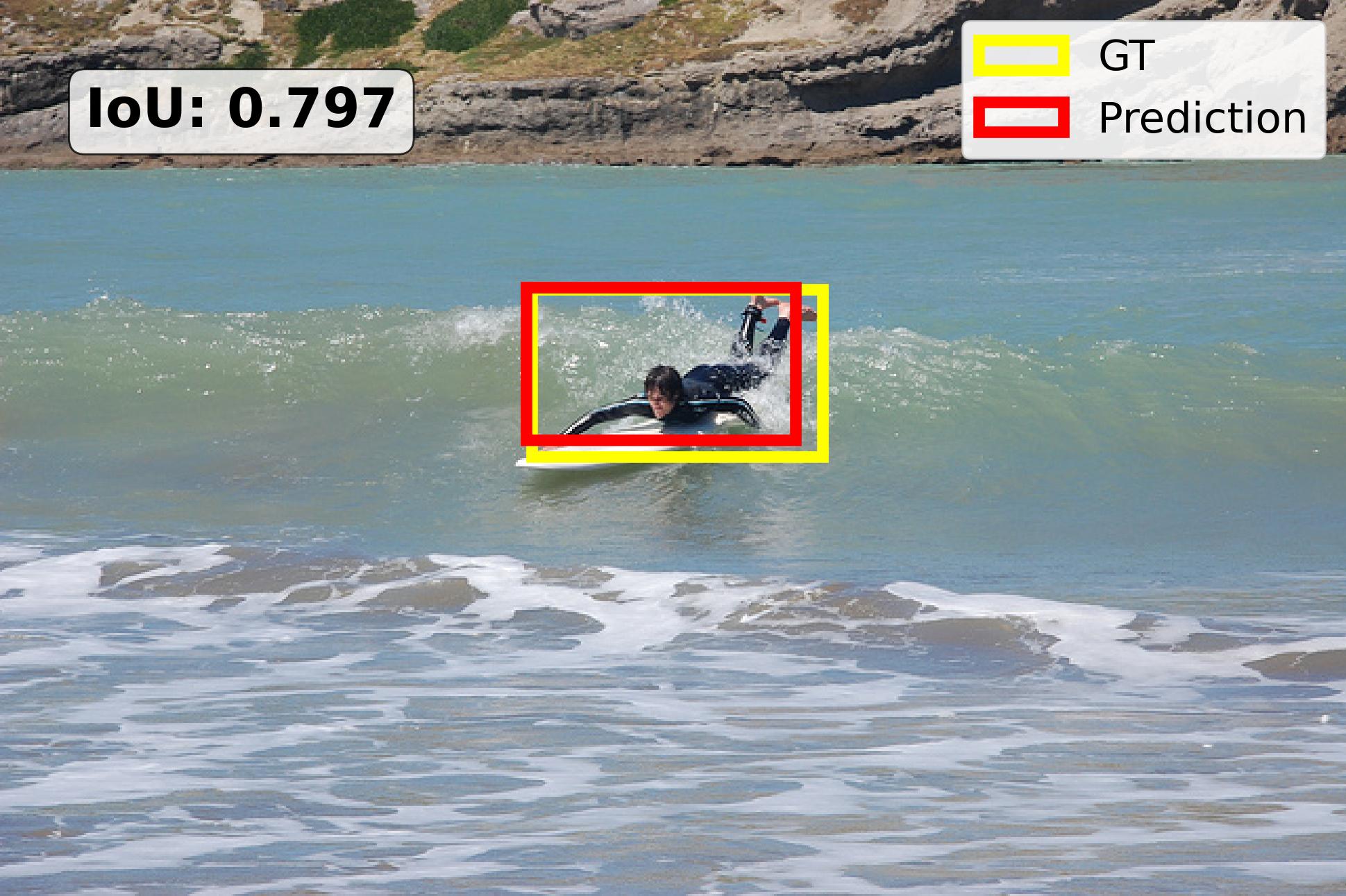}\\
\end{tabular}
\caption{
Additional qualitative comparisons on COCO using InternVL3.5.
Each case shows Greedy (top) and Ours (bottom).
}
\label{fig:qualitative_suppl_page1_coco_internvl}
\end{figure*}

\begin{figure*}[t]
\centering
\setlength{\tabcolsep}{4pt}
\begin{tabular}{ccc}
Scissors & Book & Knife\\
\includegraphics[height=1.35in]{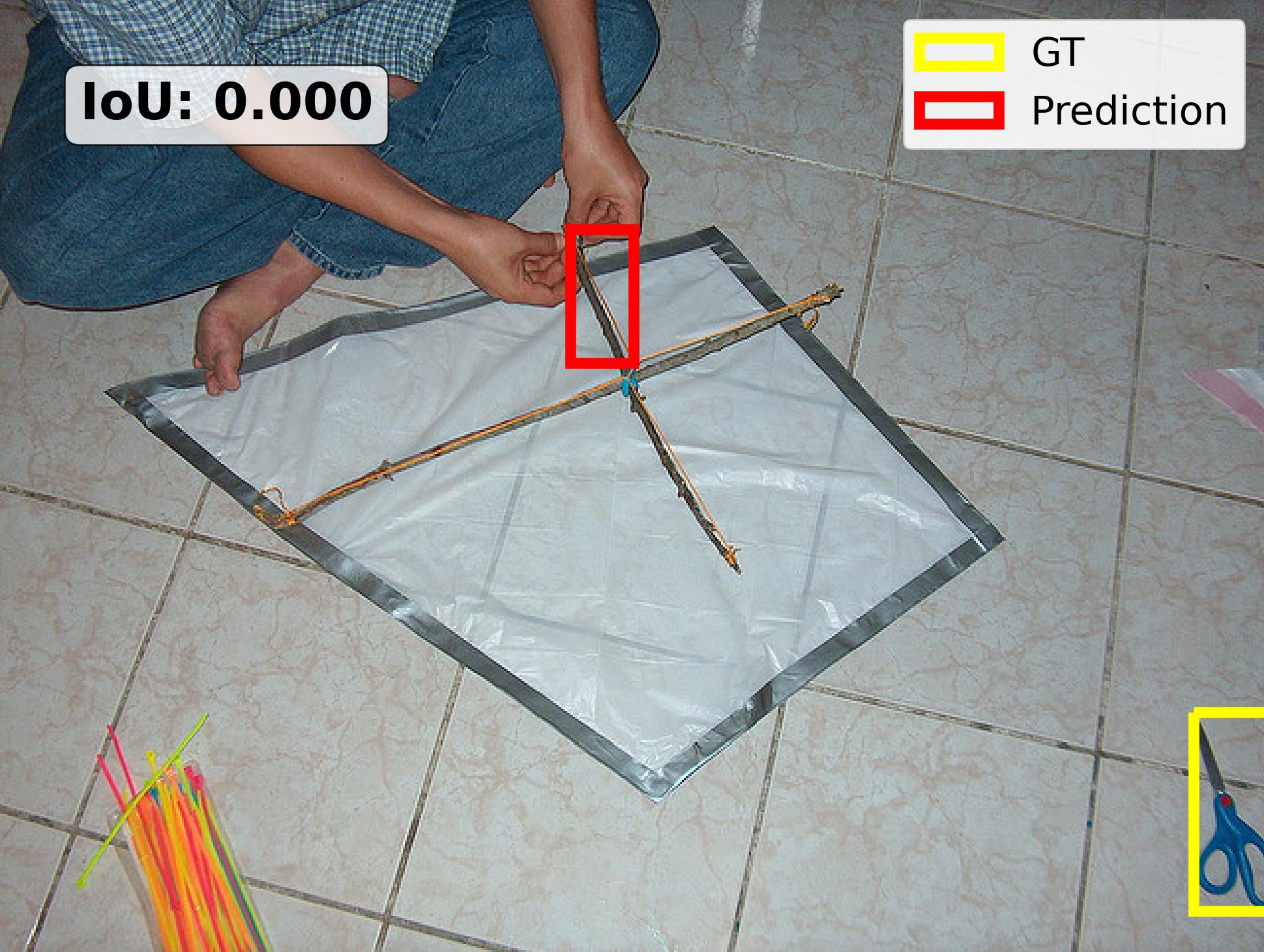} &
\includegraphics[height=1.35in]{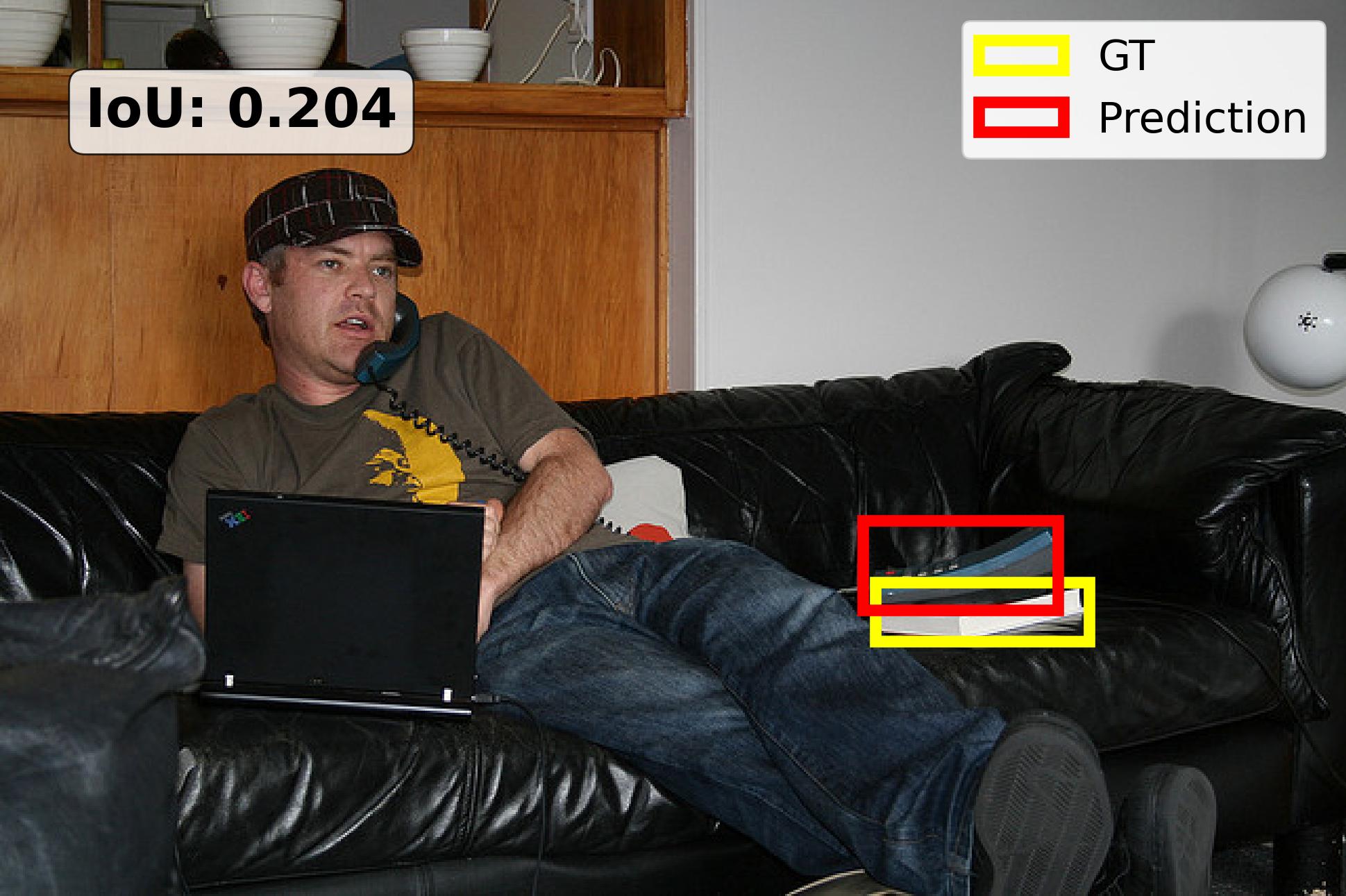} &
\includegraphics[height=1.35in]{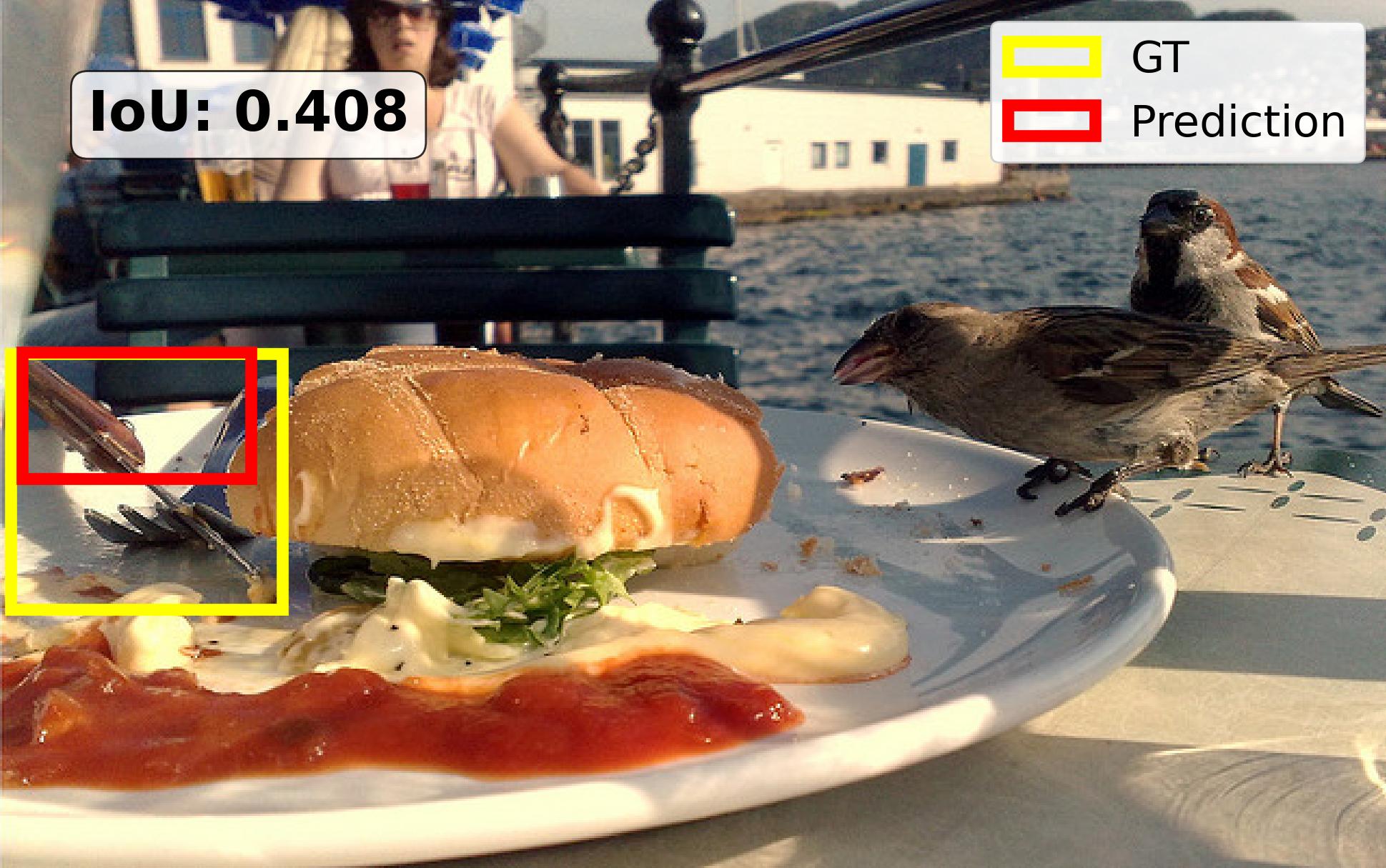}\\
\includegraphics[height=1.35in]{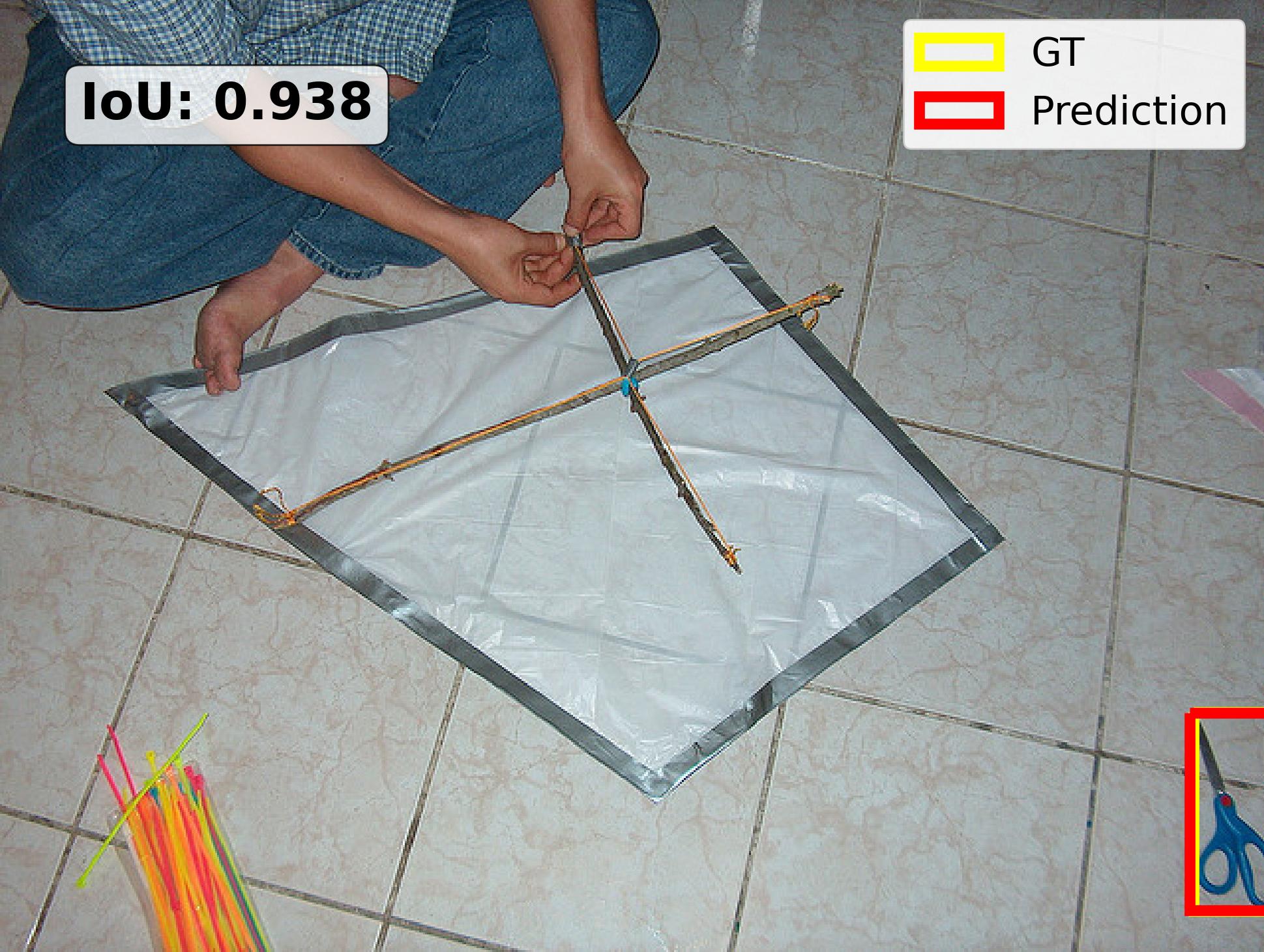} &
\includegraphics[height=1.35in]{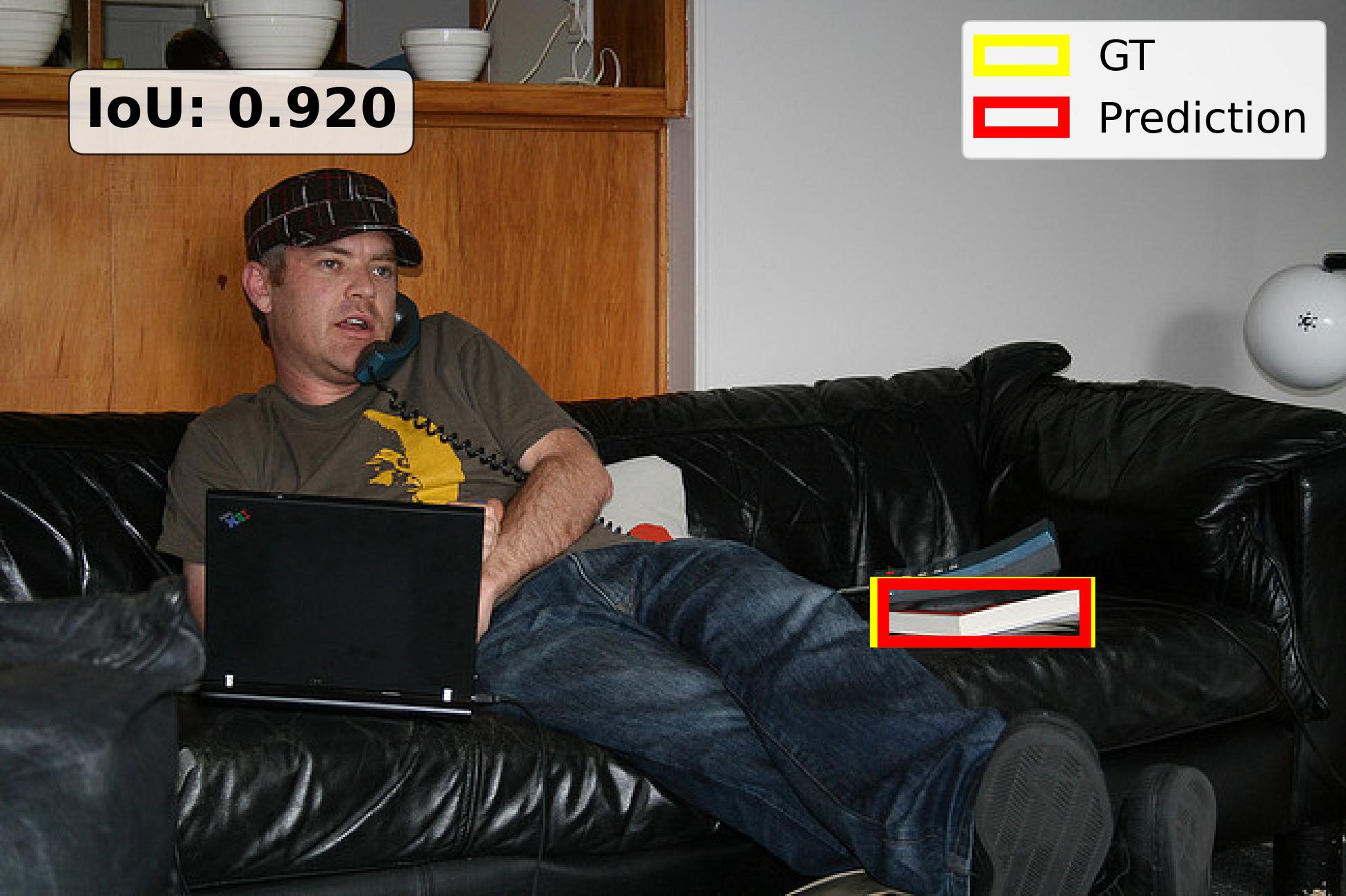} &
\includegraphics[height=1.35in]{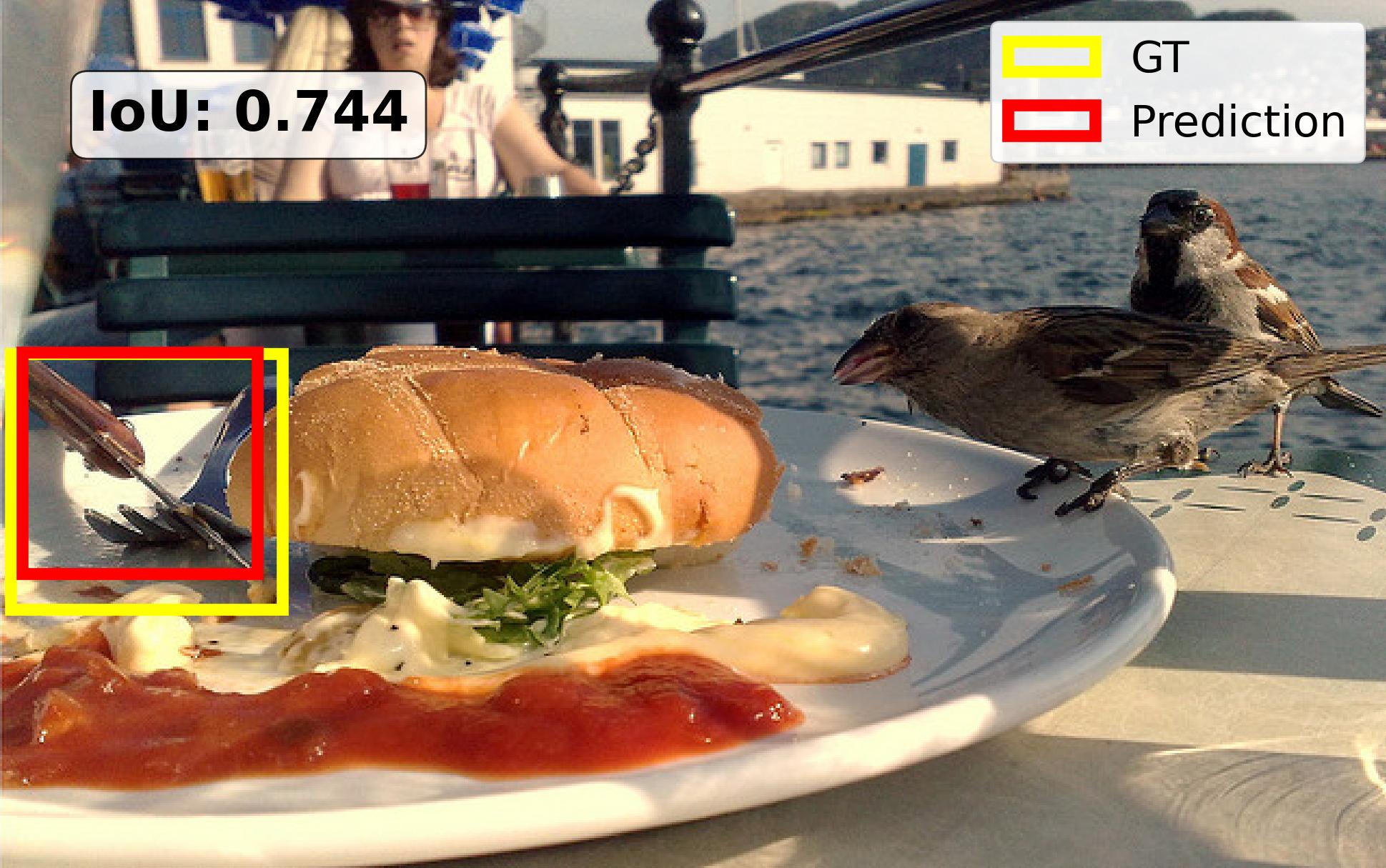}\\[4pt]
Mouse & Frisbee & Boat\\
\includegraphics[height=1.35in]{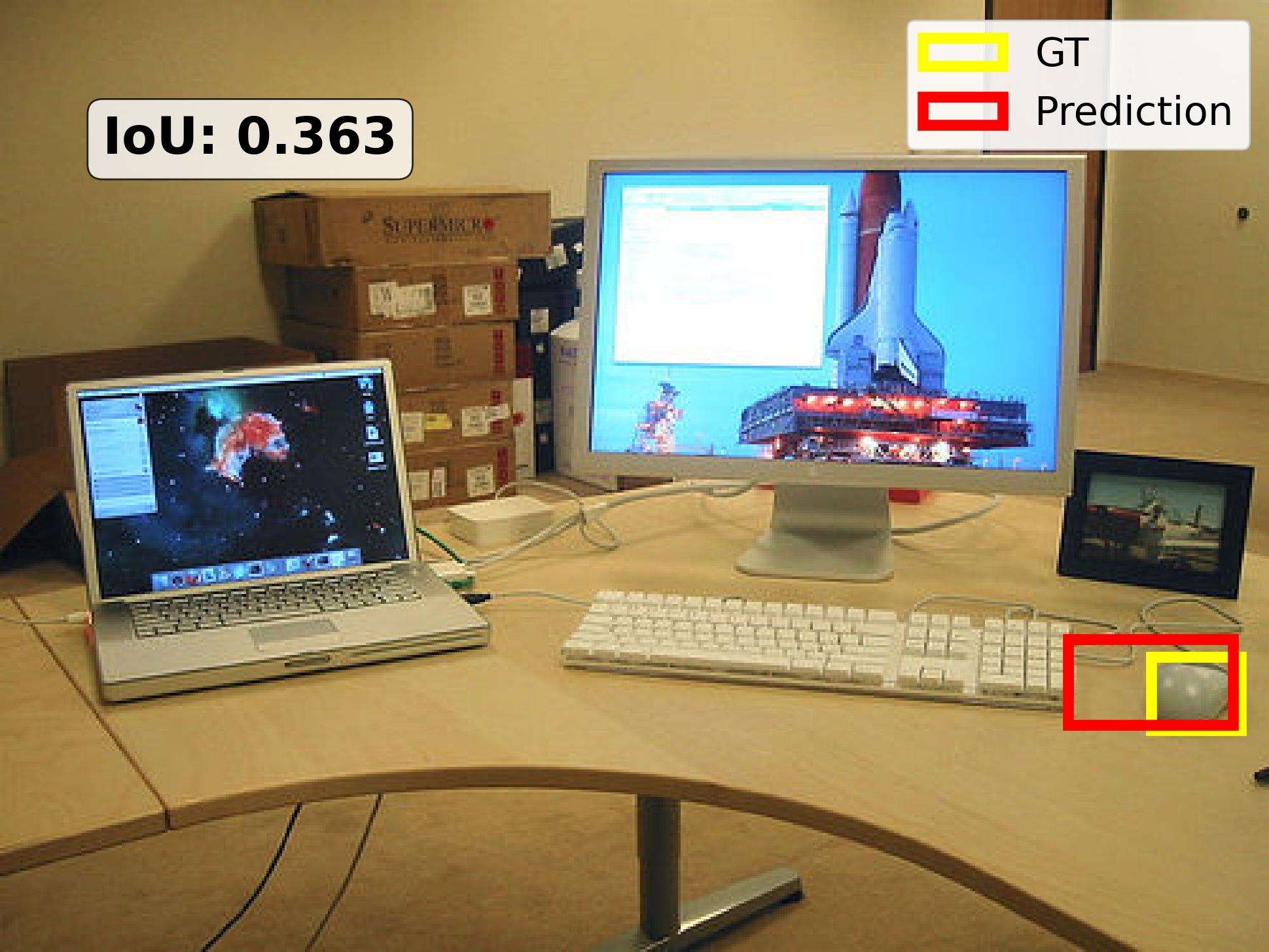}&
\includegraphics[height=1.35in]{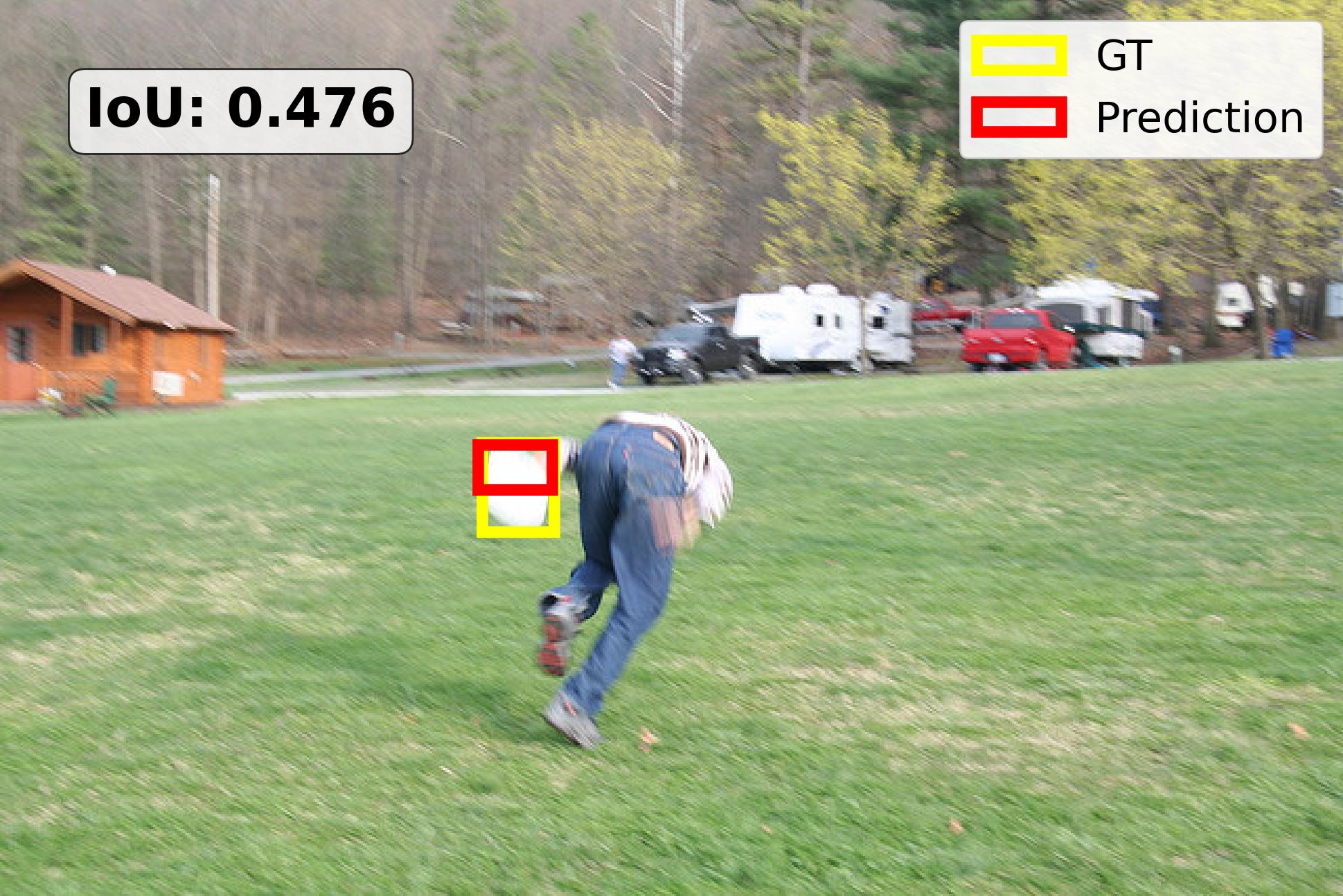}&
\includegraphics[height=1.35in]{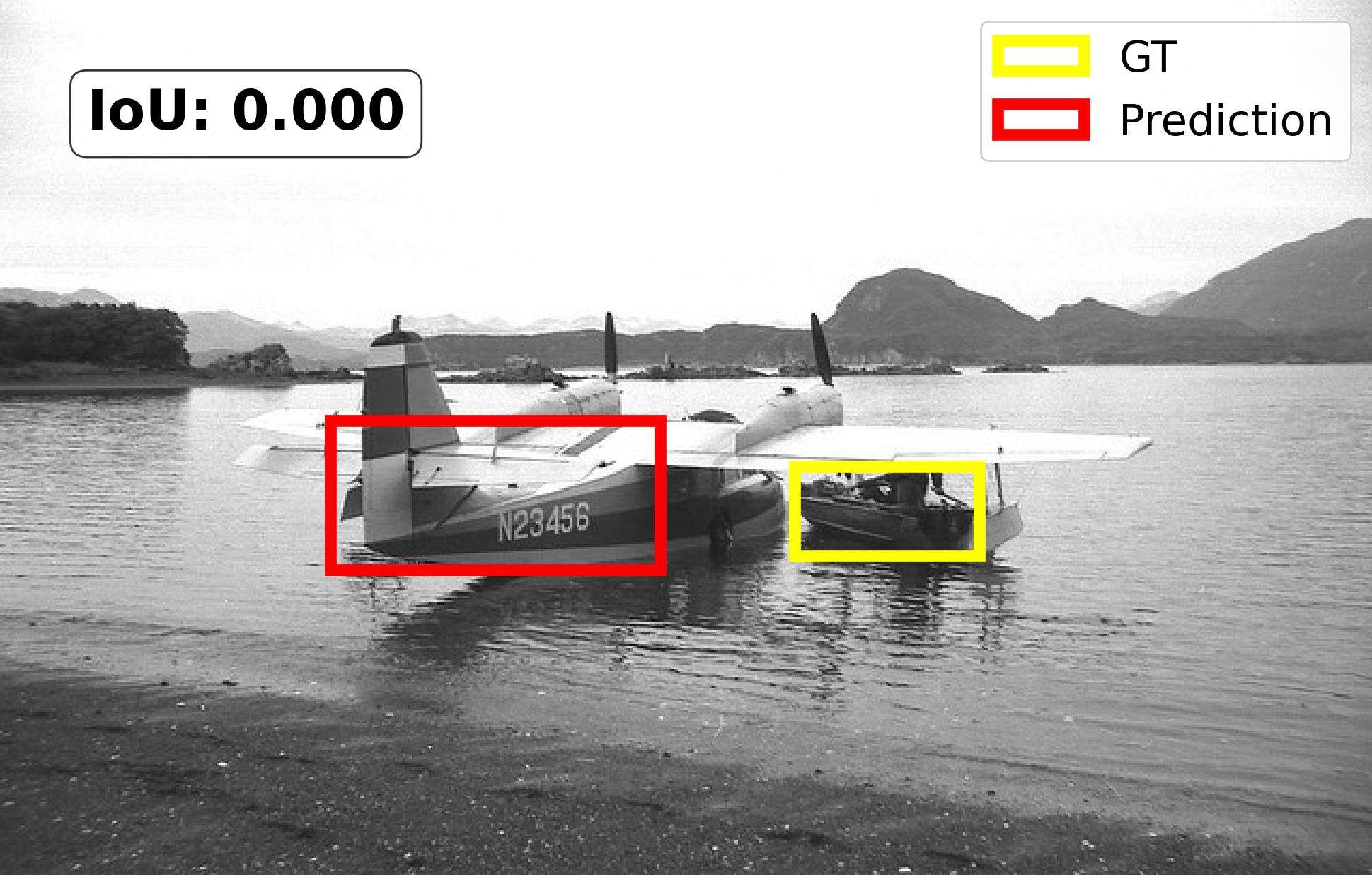}\\
\includegraphics[height=1.35in]{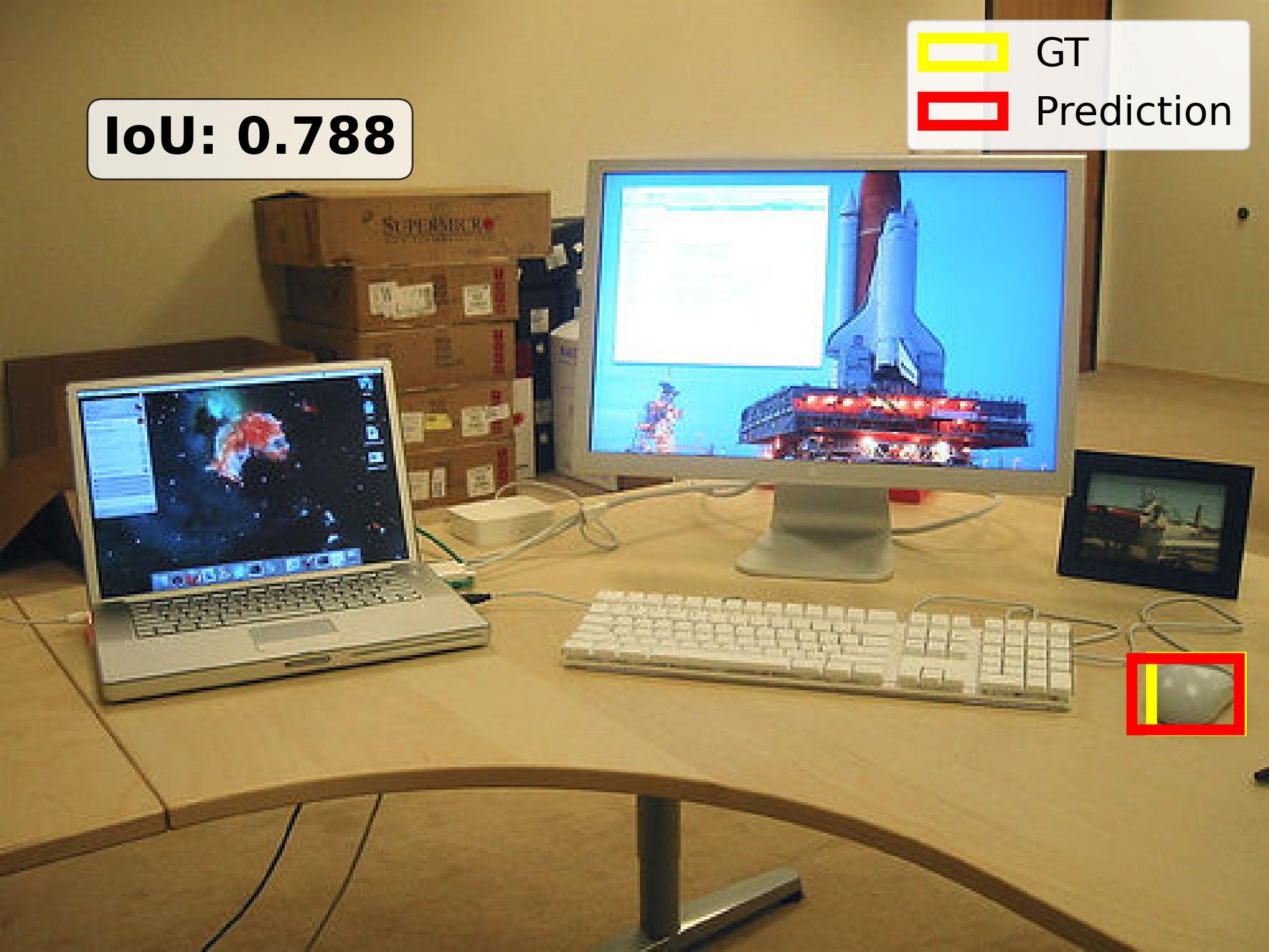}&
\includegraphics[height=1.35in]{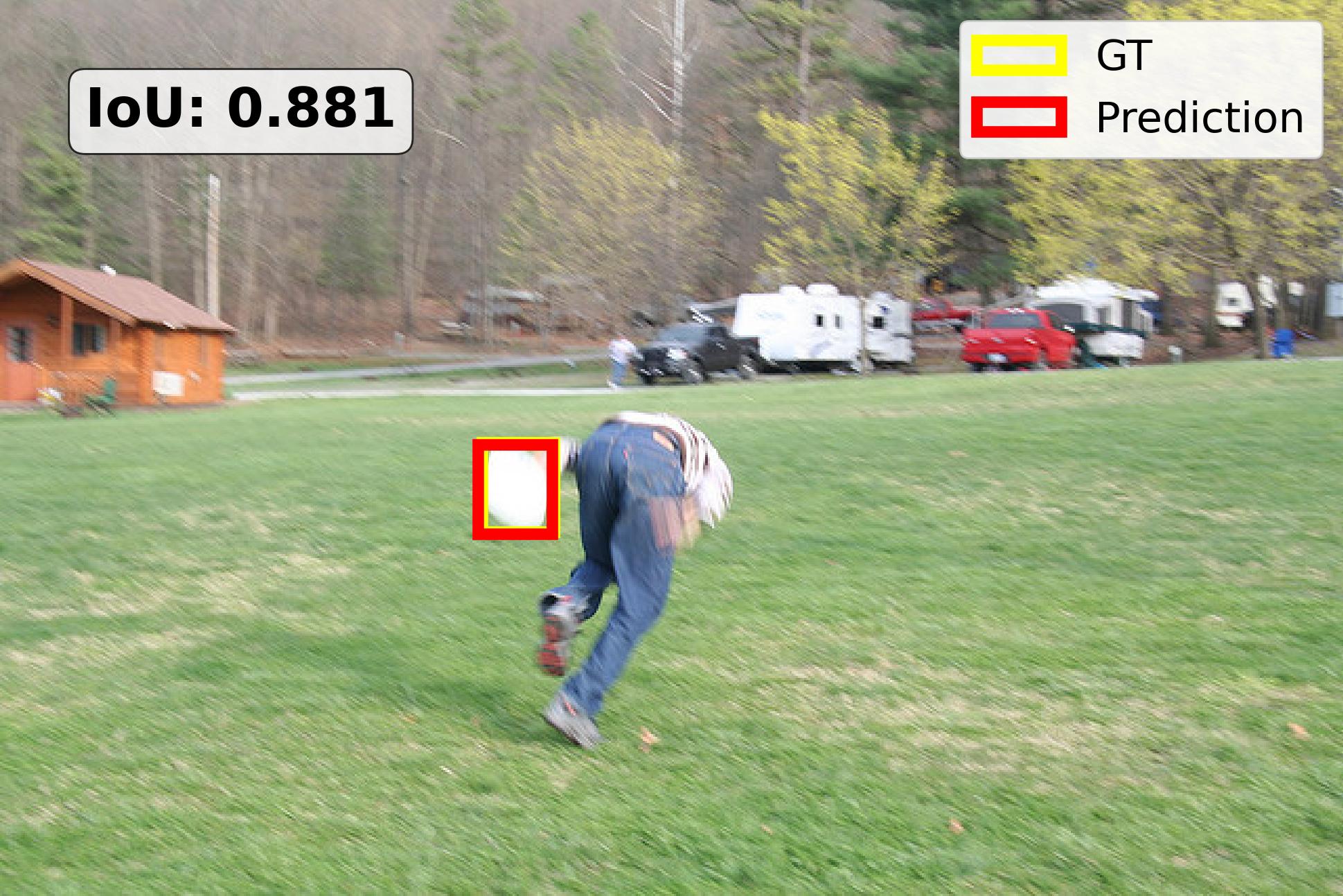}&
\includegraphics[height=1.35in]{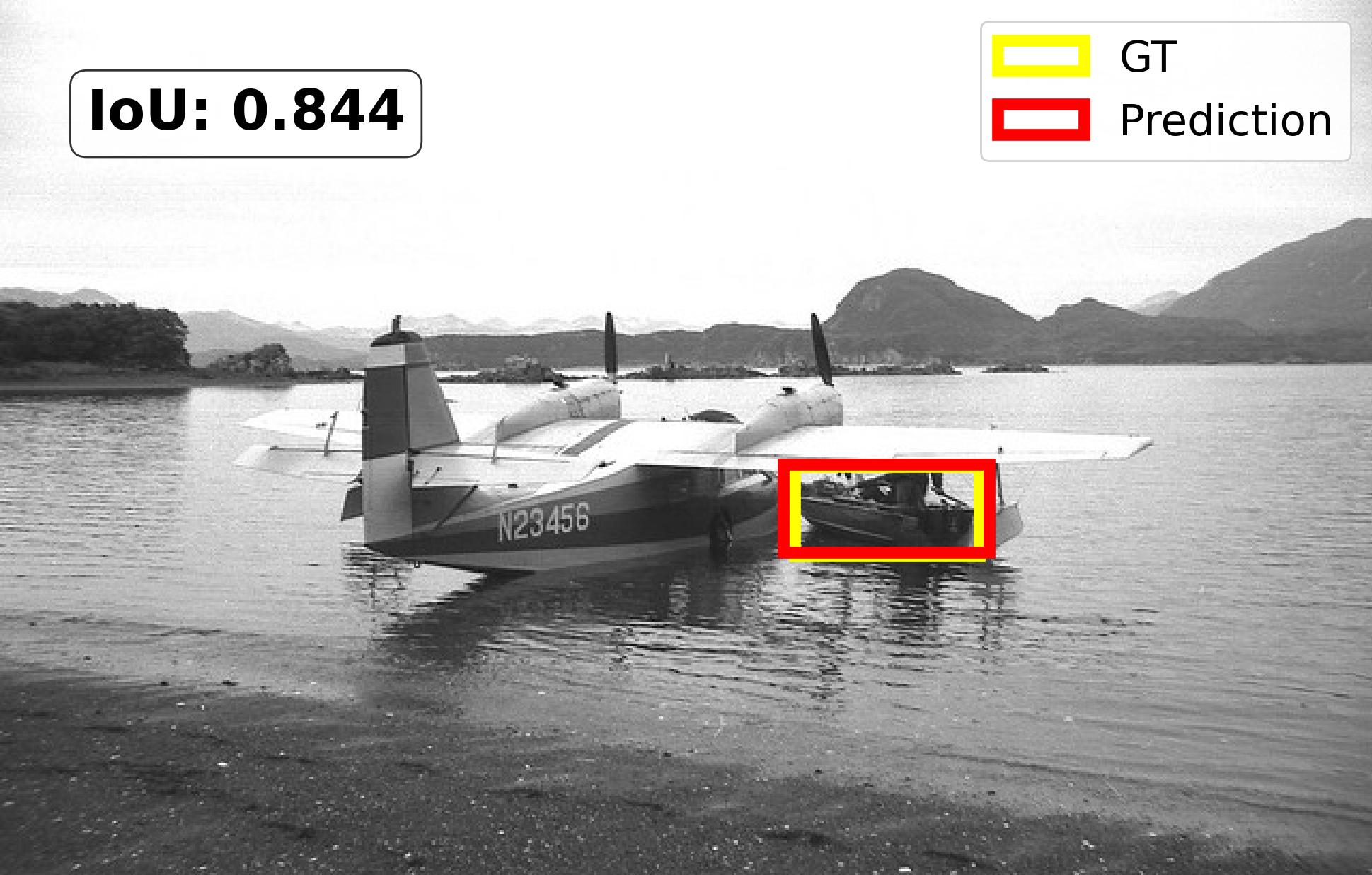}\\
\end{tabular}
\caption{
Additional qualitative comparisons on COCO using InternVL3.5.
Each case shows Greedy (top) and Ours (bottom).
}
\label{fig:qualitative_suppl_page2_coco_internvl}
\end{figure*}

\begin{figure*}[t]
\centering
\setlength{\tabcolsep}{4pt}
\begin{tabular}{ccc}
Laptop & Telephone & Mouse\\
\includegraphics[height=1.35in]{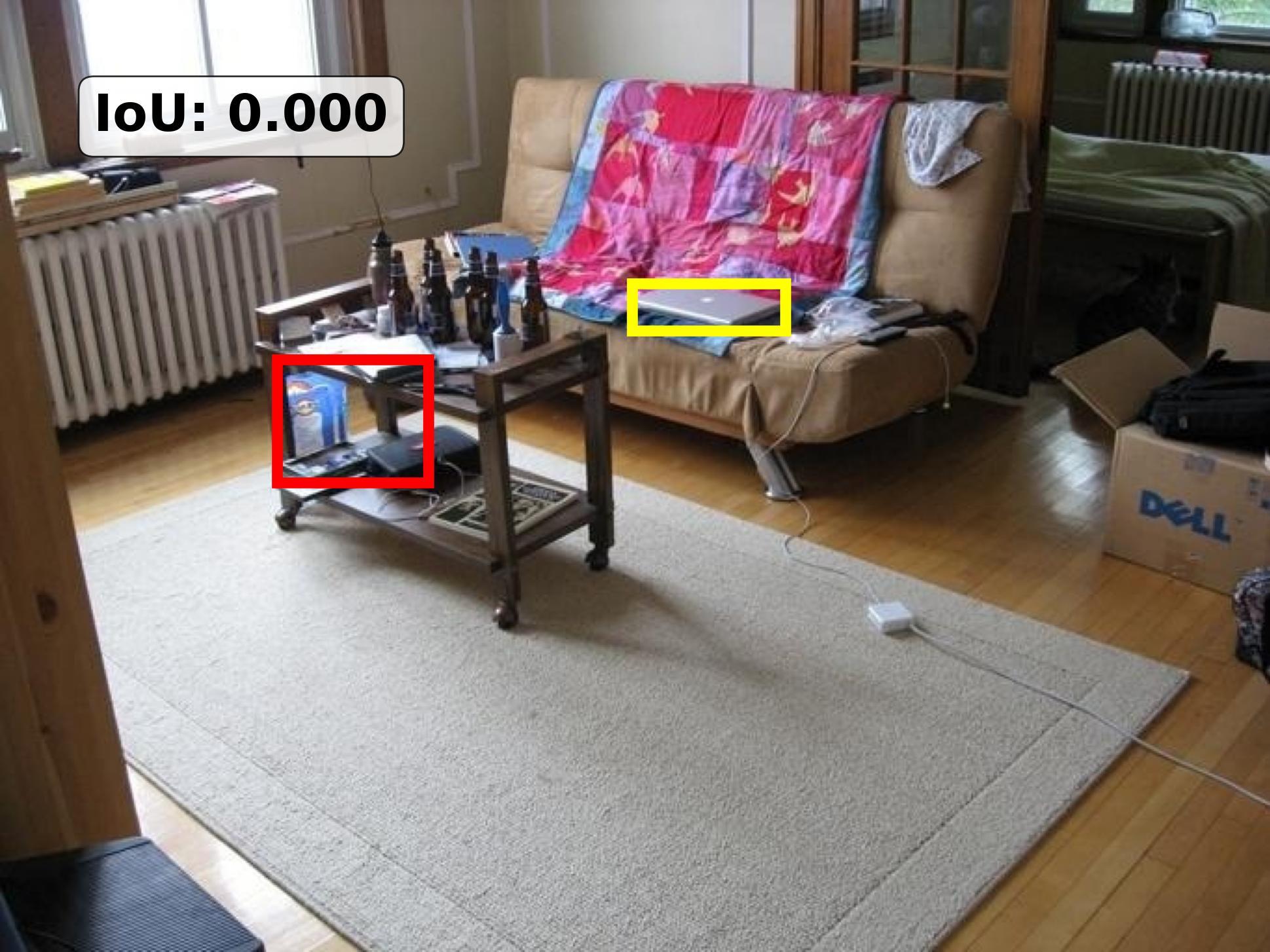} &
\includegraphics[height=1.35in]{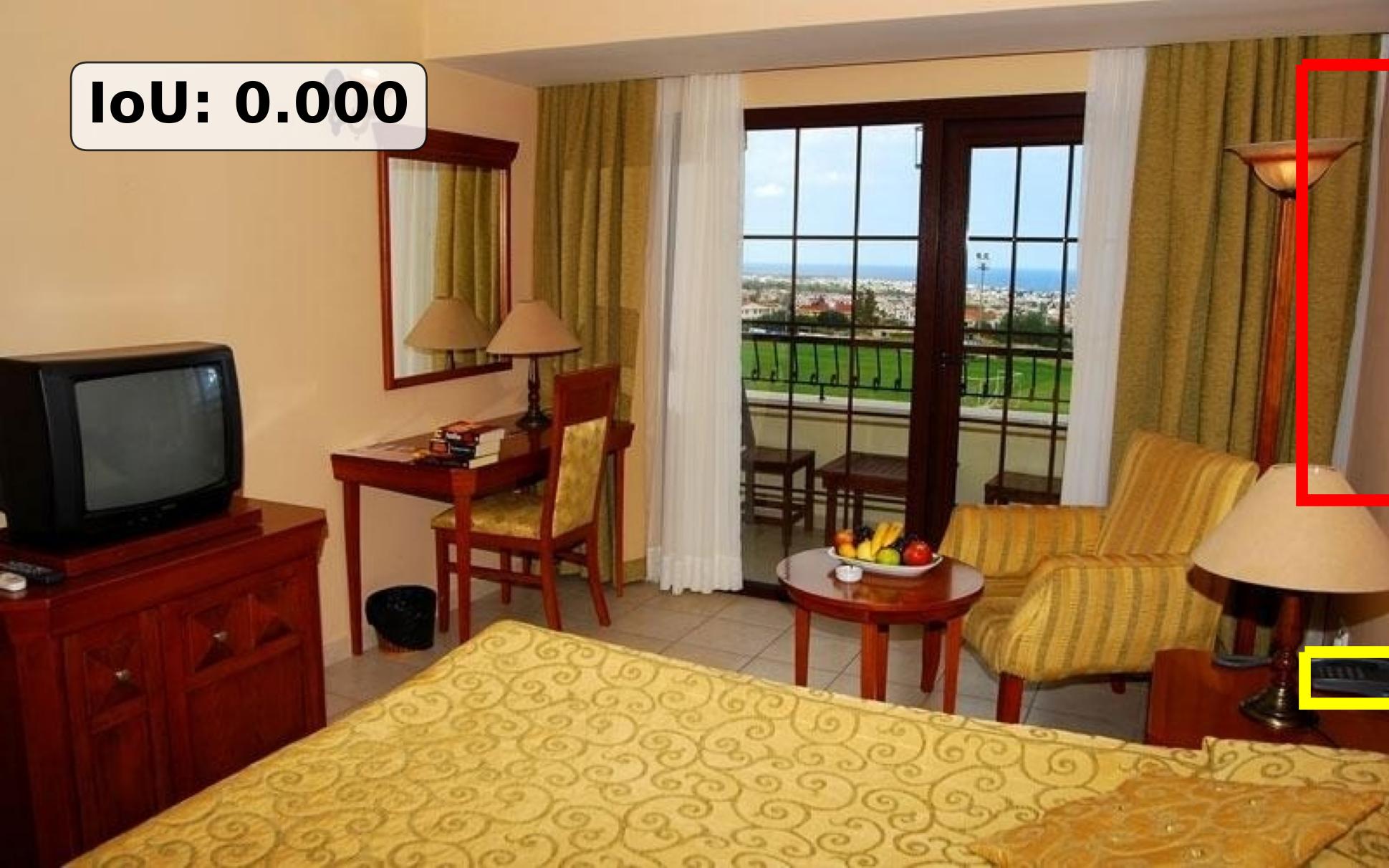} &
\includegraphics[height=1.35in]{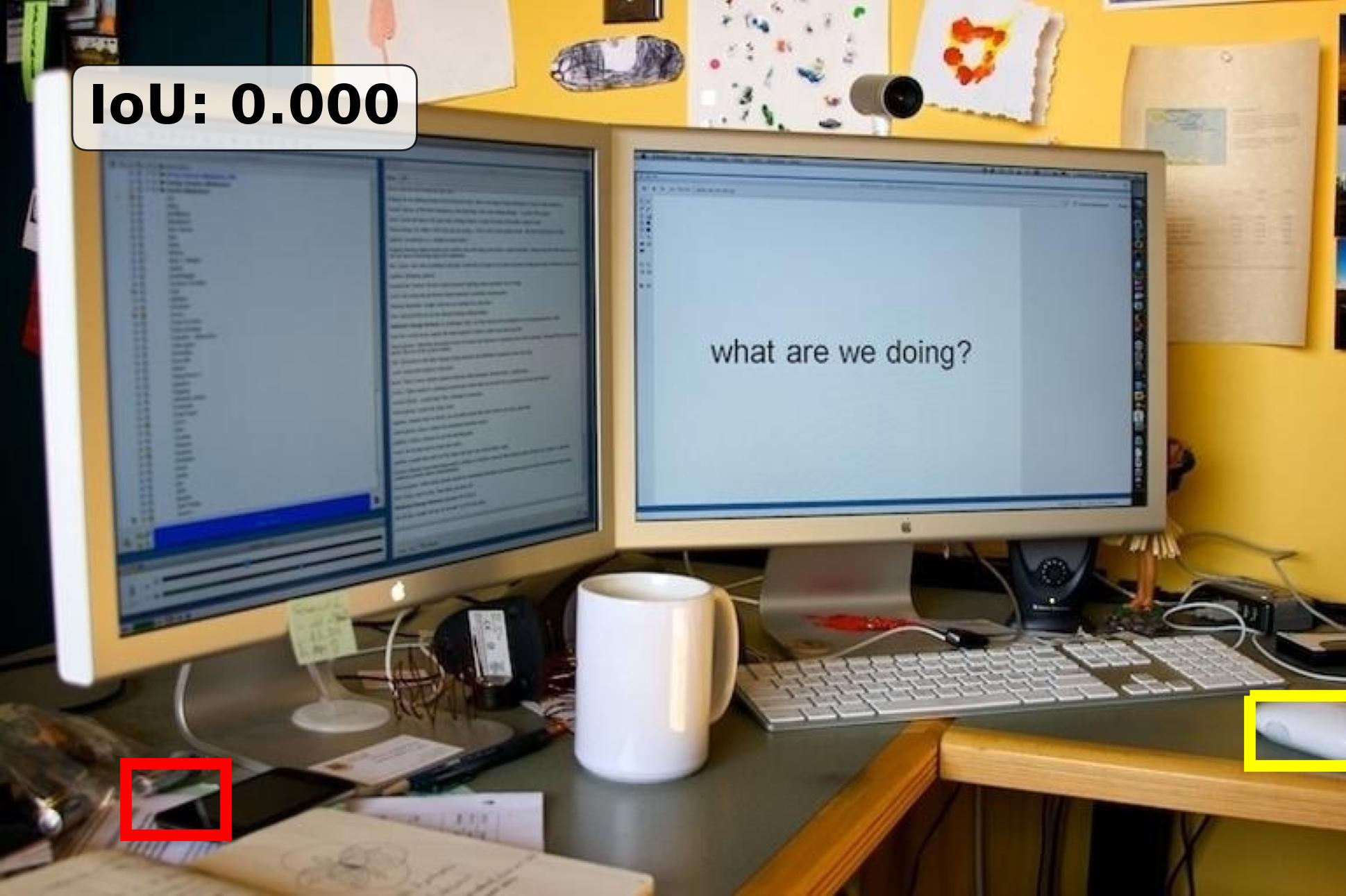}\\
\includegraphics[height=1.35in]{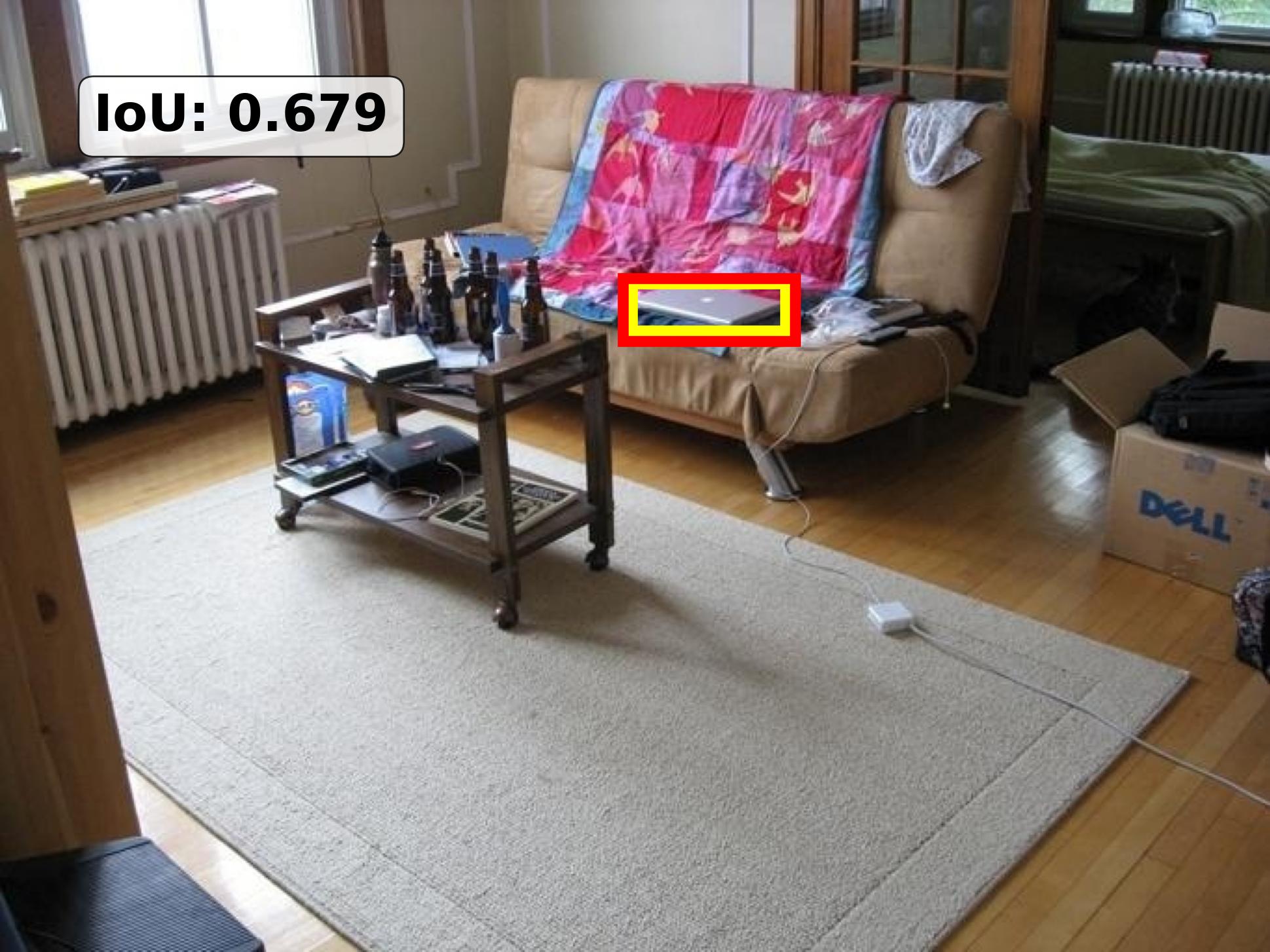} &
\includegraphics[height=1.35in]{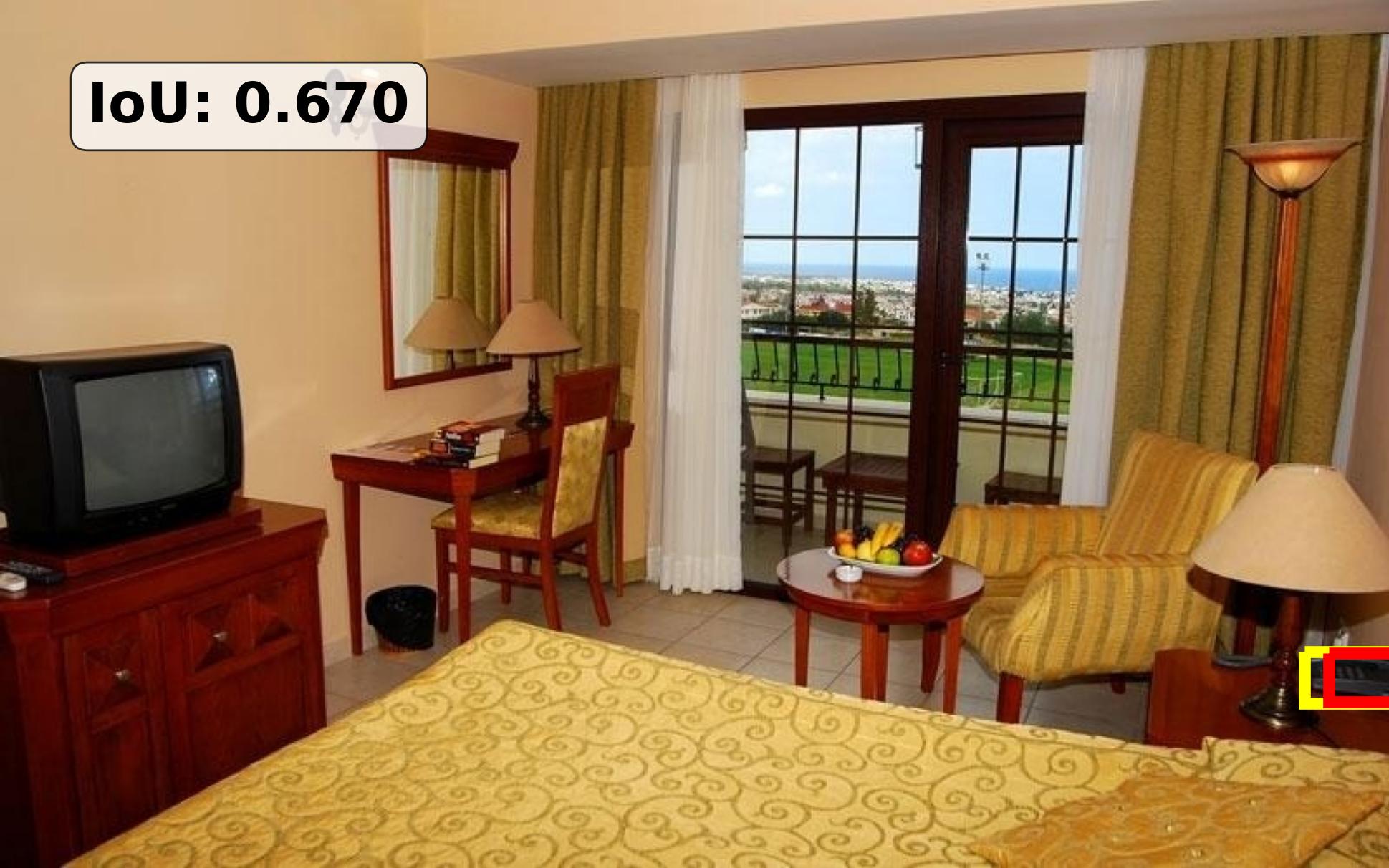} &
\includegraphics[height=1.35in]{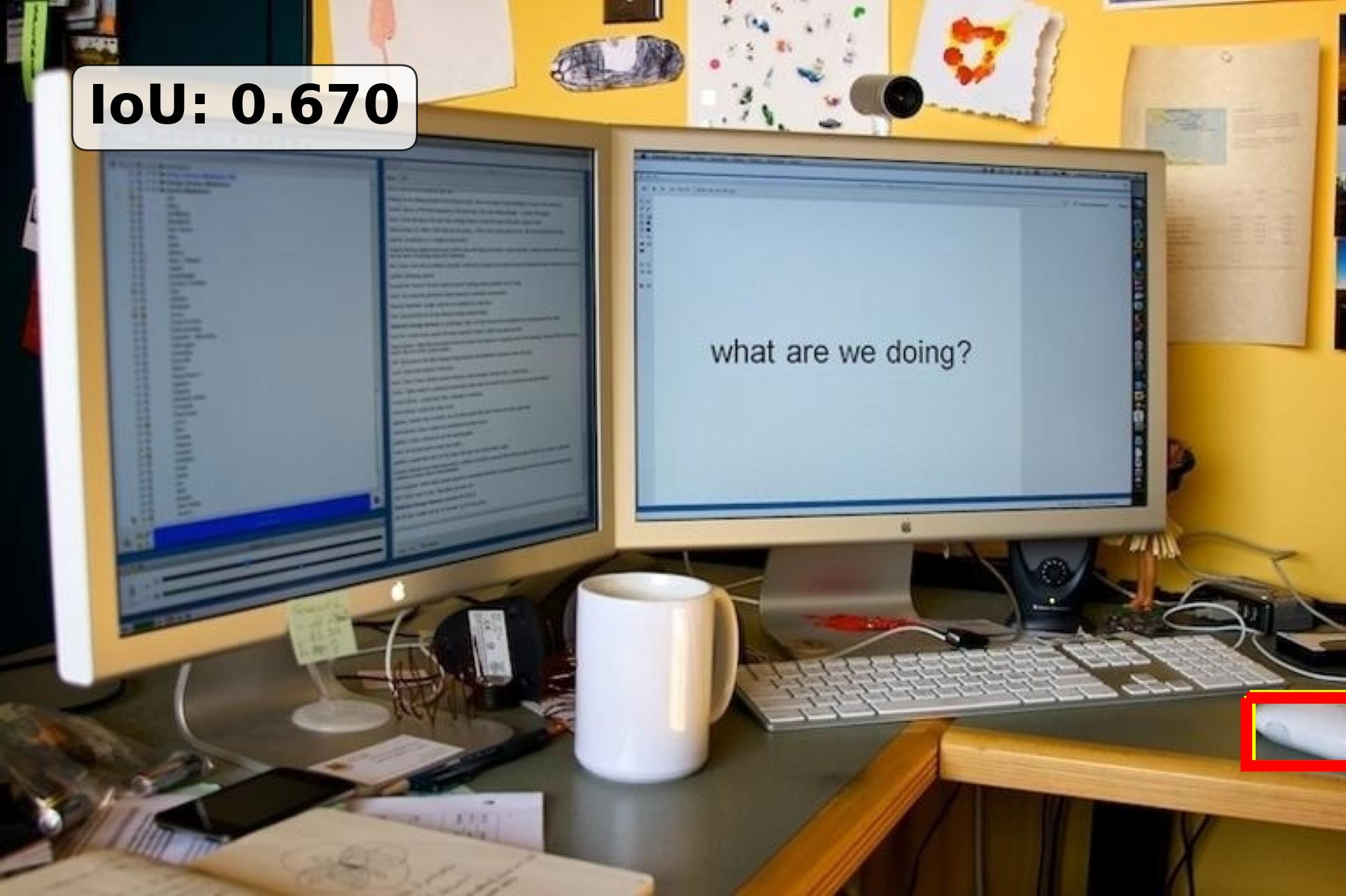}\\[4pt]
Extractor & Lamp & Coffee Machine\\
\includegraphics[height=1.35in]{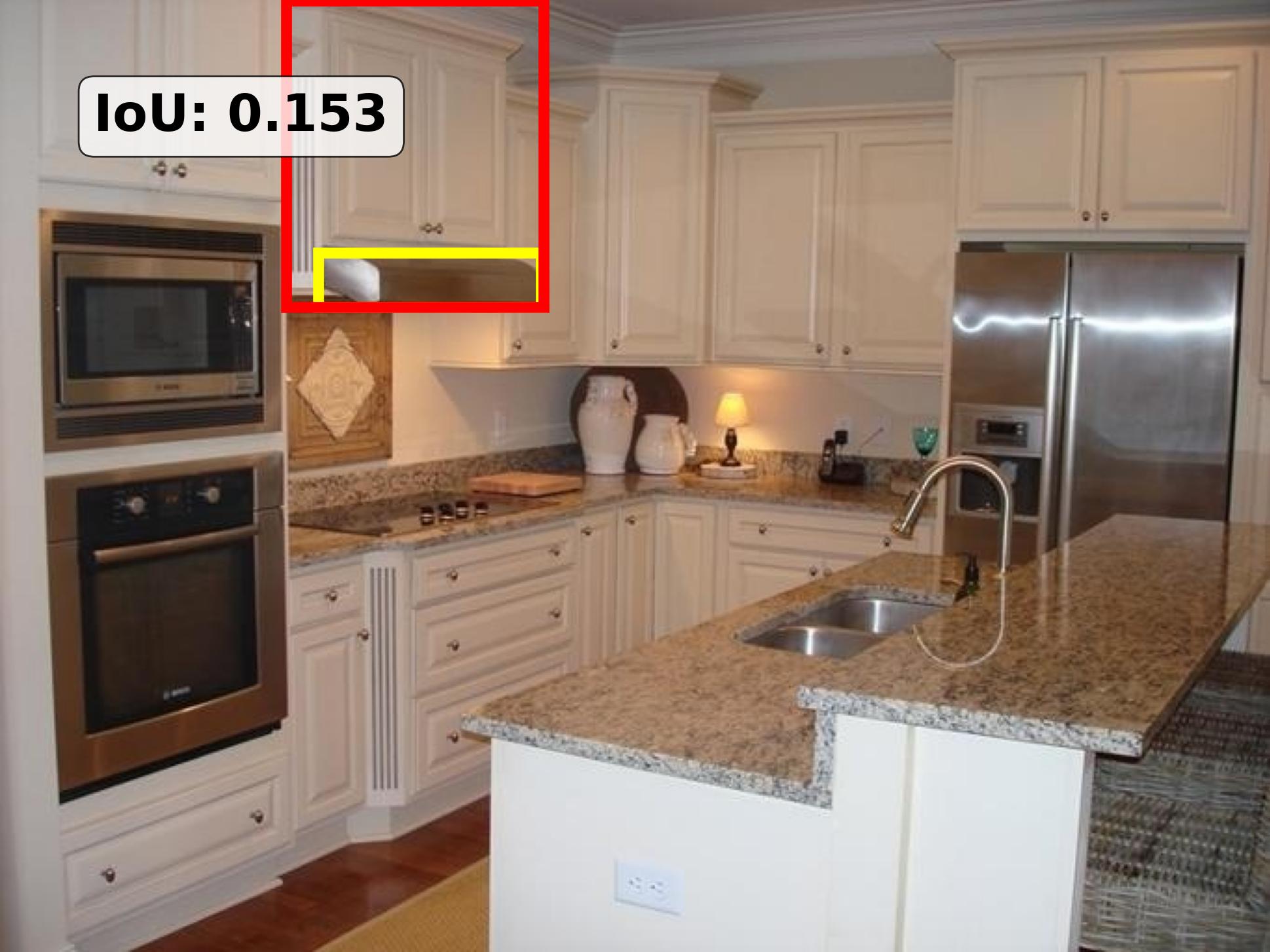}&
\includegraphics[height=1.35in]{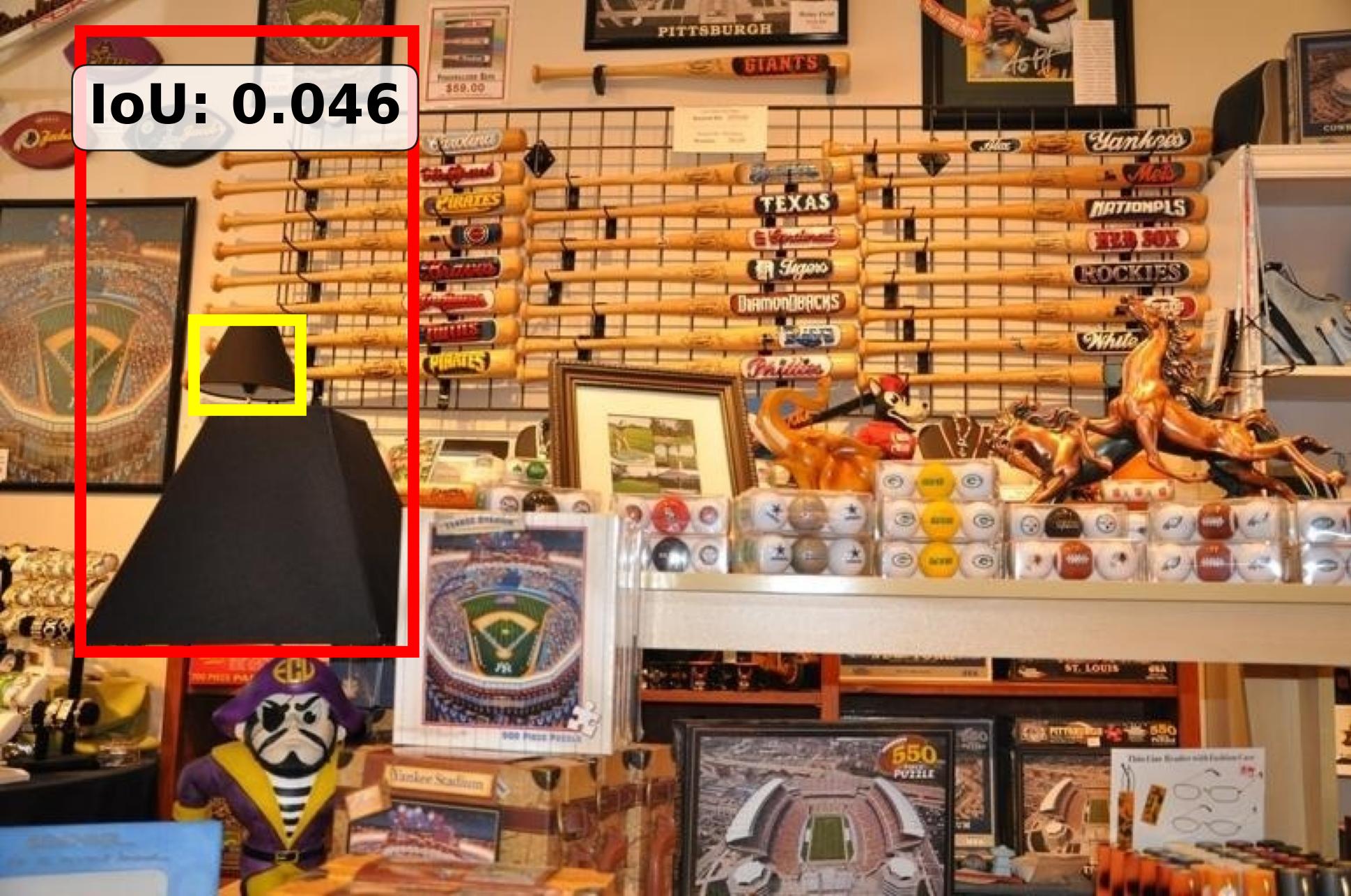}&
\includegraphics[height=1.35in]{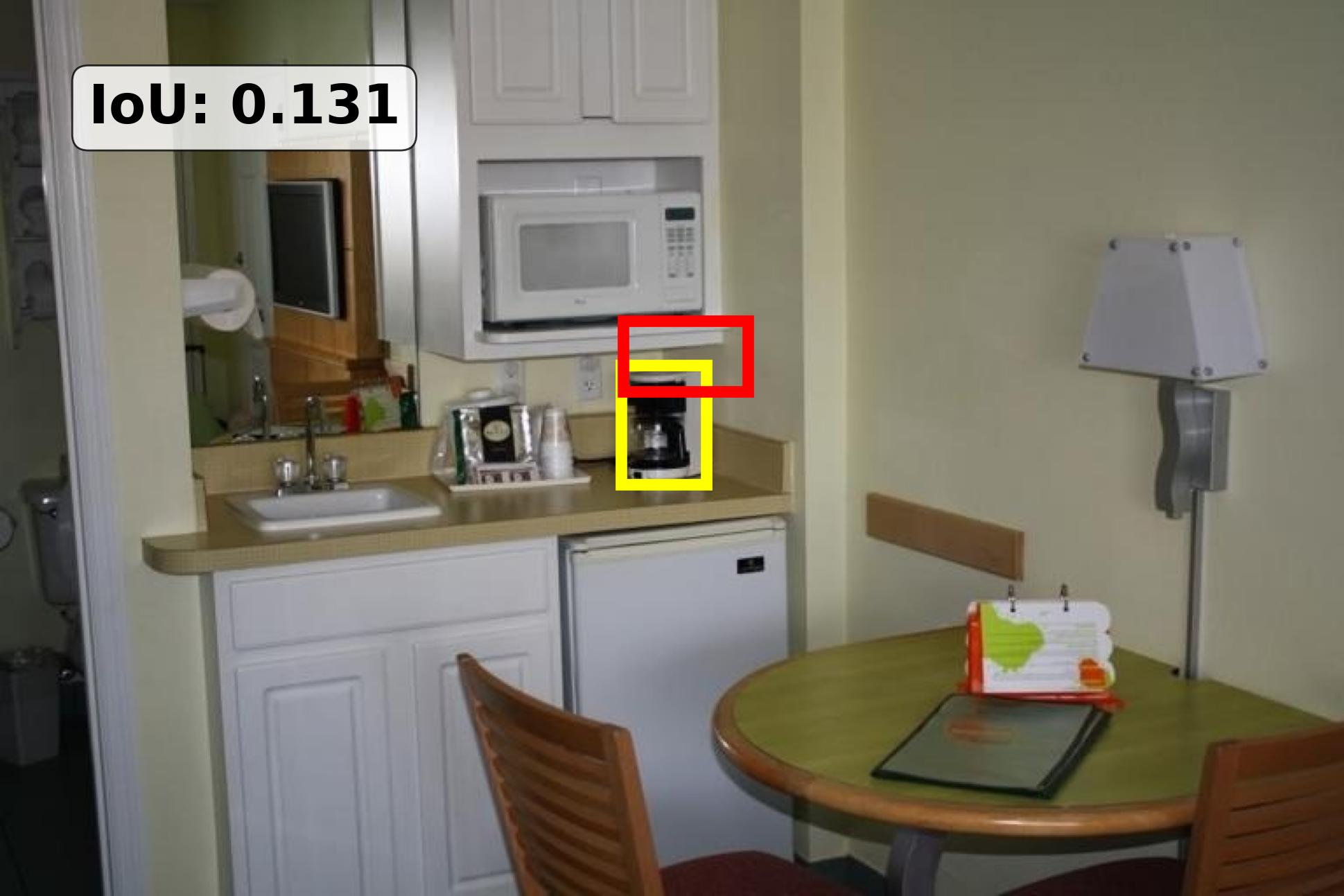}\\
\includegraphics[height=1.35in]{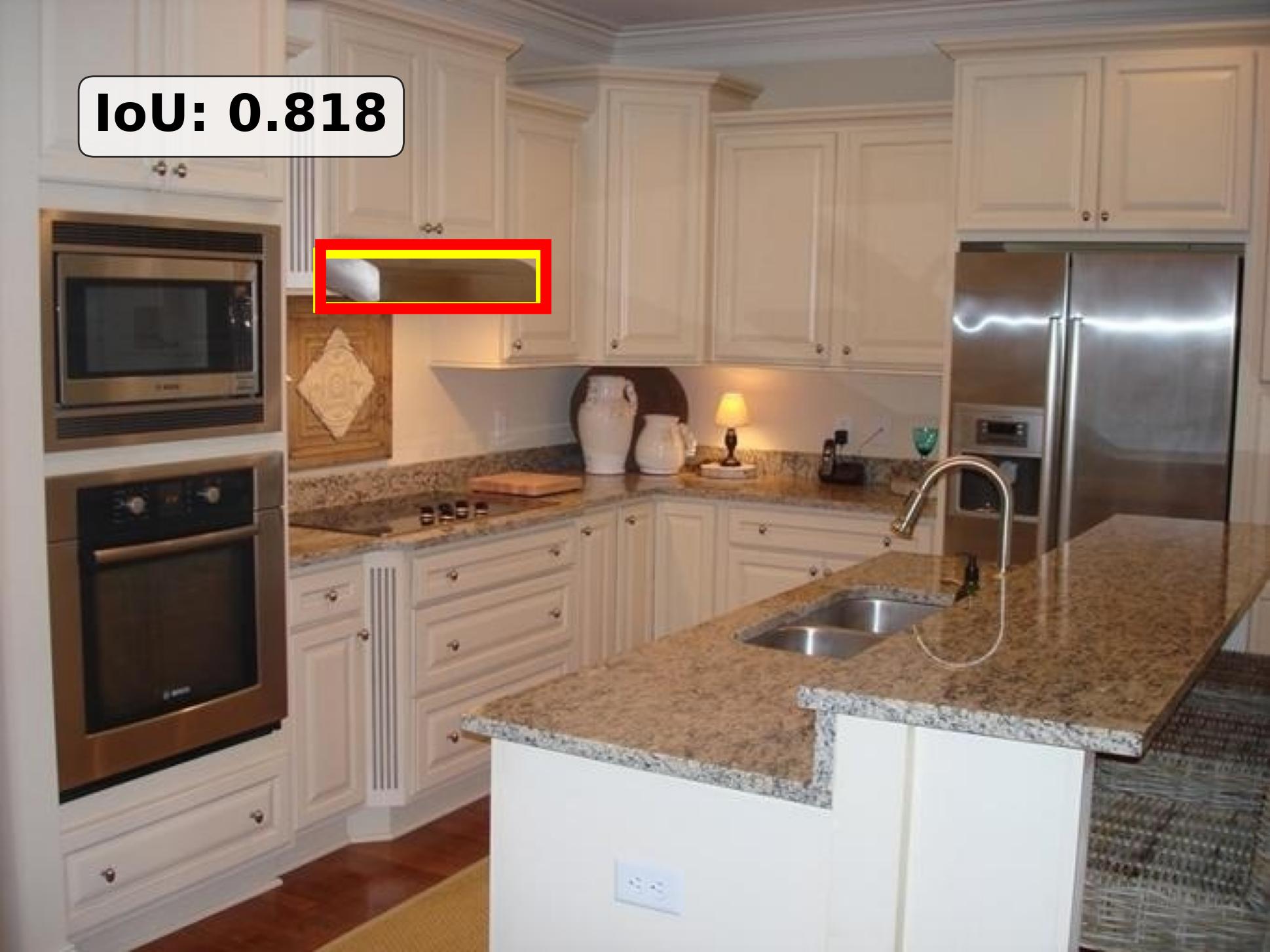}&
\includegraphics[height=1.35in]{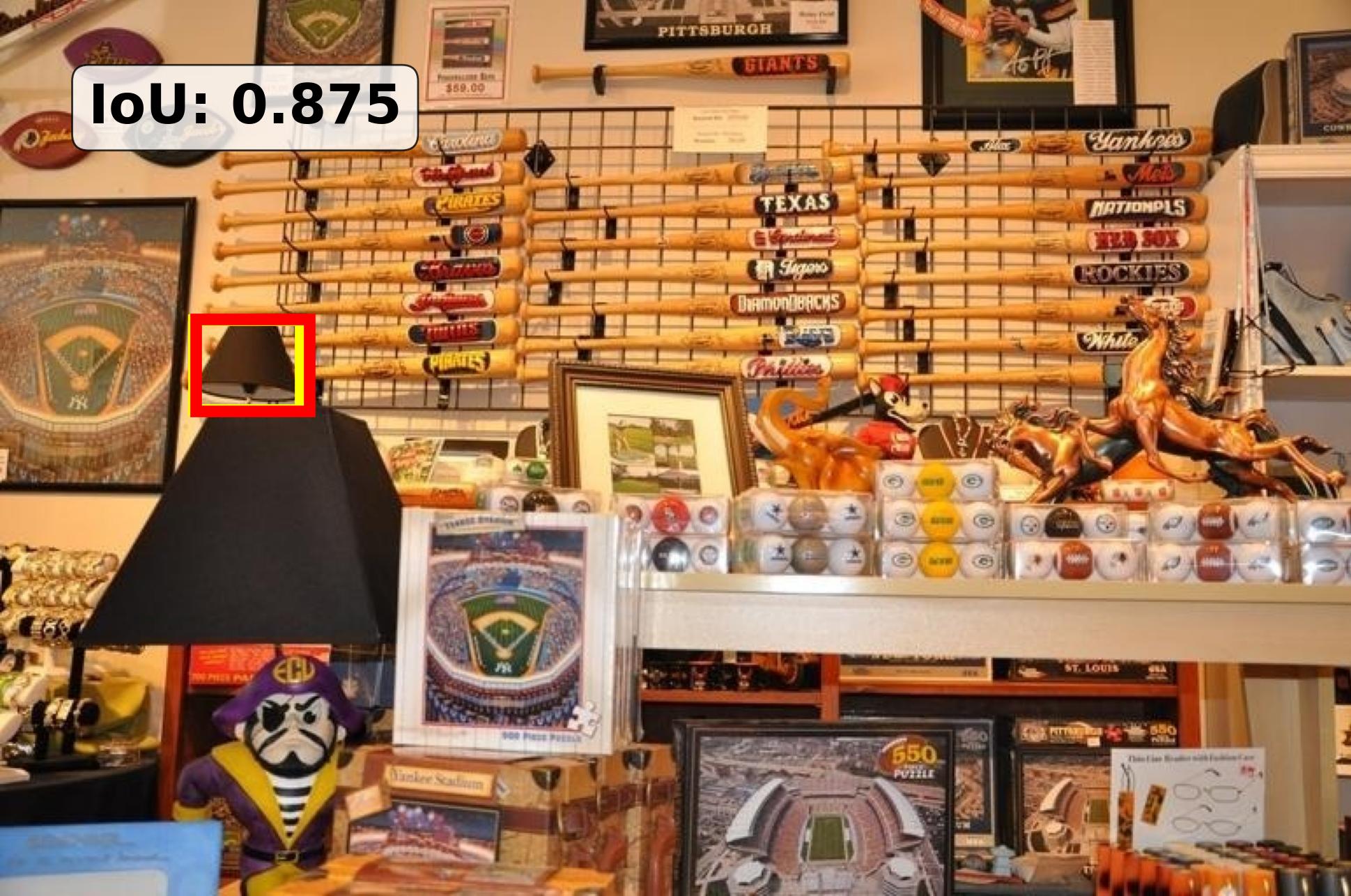}&
\includegraphics[height=1.35in]{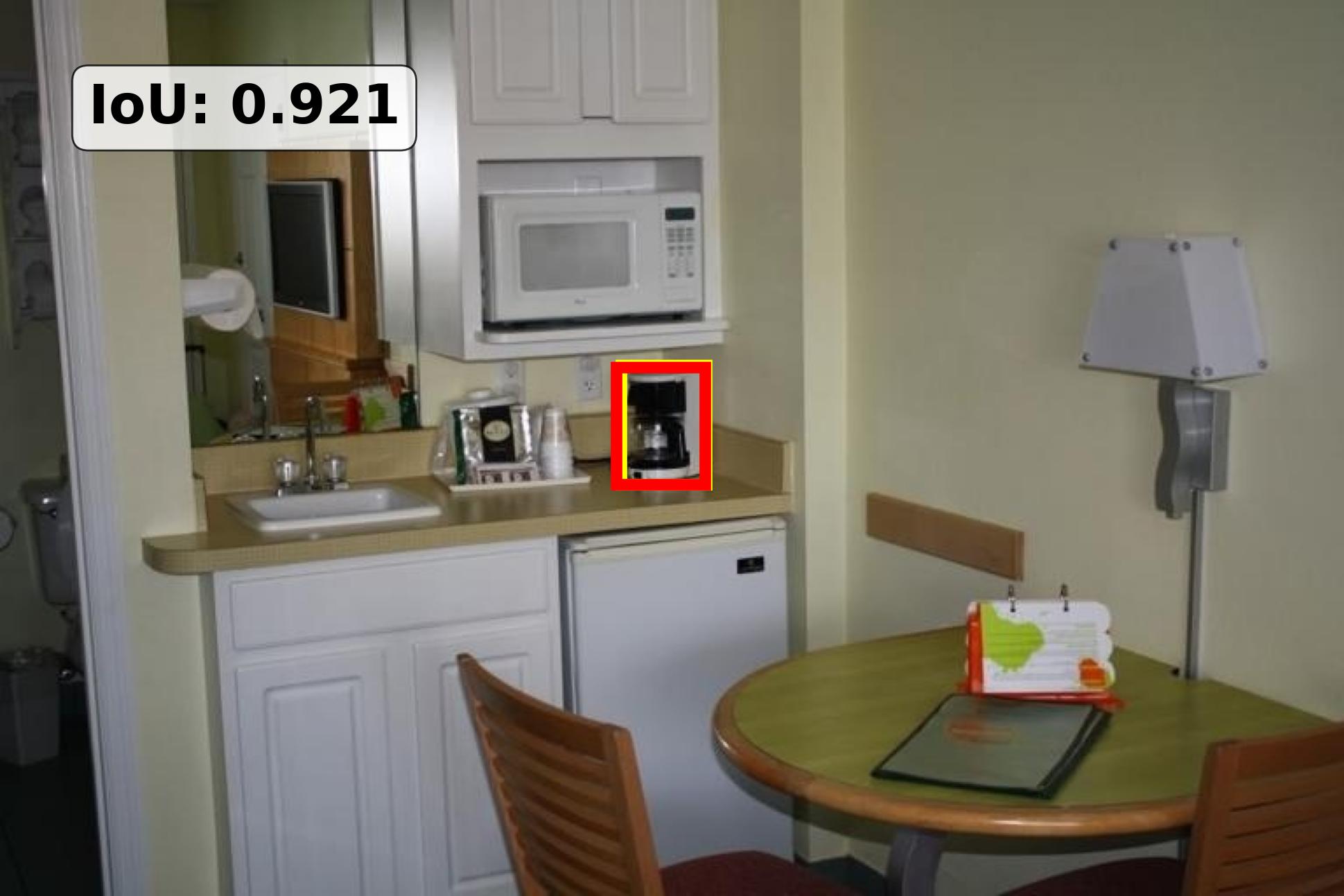}\\
\end{tabular}
\caption{
Additional qualitative comparisons on the Objects365 dataset using Qwen2.5-VL.
Each case shows Greedy (top) and Ours (bottom).
}

\label{fig:qualitative_suppl_objects365}
\end{figure*}

\cref{fig:qualitative_suppl_page1_coco_qwen,fig:qualitative_suppl_page2_coco_qwen,fig:qualitative_suppl_page1_coco_internvl,fig:qualitative_suppl_page2_coco_internvl,fig:qualitative_suppl_objects365}
provide additional qualitative localization results that complement Figure~5 in the main paper. Each panel corresponds to a different object category. For each case, the top image shows the prediction from Greedy decoding and the bottom image shows the result from our method, following the same visualization protocol as the main text: yellow boxes denote ground-truth bounding boxes and red boxes denote predicted bounding boxes, with the IoU value annotated in the corner.

\end{document}